\documentclass{bmvc2k}

\title{Analysing Training-Data Leakage from Gradients through Linear Systems and Gradient Matching}

\addauthor{Cangxiong Chen}{cc2458@bath.ac.uk}{1}
\addauthor{Neill D.\,F.~Campbell}{n.campbell@bath.ac.uk}{2}

\addinstitution{
 Institute for Mathematical Innovation,\\
 University of Bath, UK
}
\addinstitution{
 Department of Computer Science,\\
 University of Bath, UK
}

\runninghead{Chen, Campbell}{Analysing Training-Data Leakage from Gradients}

\usepackage{hyperref}
\usepackage{url}
\usepackage{ulem}
\usepackage{url}
\usepackage{enumitem}
\usepackage{algorithm}
\usepackage{xcolor}
\usepackage{amsmath}
\usepackage{amssymb}
\usepackage{mathtools}
\usepackage{amsthm}
\usepackage{booktabs} 

\usepackage{rotating}
\usepackage{caption}
\usepackage{subcaption}

\usepackage{etoolbox} 
\let\classAND\AND
\let\AND\relax
\usepackage{algorithmic}

\let\AND\classAND
\AtBeginEnvironment{algorithmic}{\let\AND\algoAND}

\usepackage{multicol}
\DeclareMathOperator*{\argmin}{argmin}
\newcommand{\CL}{\mathcal{L}}
\newcommand{\BR}{\mathbb{R}}
\newcommand{\CM}{\mathcal{M}}
\usepackage{bm}
\DeclareMathOperator{\rank}{rank}

\theoremstyle{plain}
\newtheorem{theorem}{Theorem}[section]

\newtheorem{lemma}[theorem]{Lemma}

\theoremstyle{definition}
\newtheorem{definition}[theorem]{Definition}

\theoremstyle{remark}
\newtheorem{remark}[theorem]{Remark}

\renewcommand{\paragraph}[1]{\noindent\textbf{#1}~}

\begin{document}

\maketitle

\begin{abstract}
Recent works have demonstrated that it is possible to reconstruct training images and their labels from gradients of an image-classification model when its architecture is known. Unfortunately, there is still an incomplete theoretical understanding of the efficacy and failure of these gradient-leakage attacks. In this paper, we propose a novel framework to analyse training-data leakage from gradients that draws insights from both analytic and optimisation-based gradient-leakage attacks. We formulate the reconstruction problem as solving a linear system from each layer iteratively, accompanied by corrections using gradient matching. Under this framework, we claim that the solubility of the reconstruction problem is primarily determined by that of the linear system at each layer. As a result, we are able to partially attribute the leakage of the training data in a deep network to its architecture. We also propose a metric to measure the level of security of a deep learning model against gradient-based attacks on the training data. 
\end{abstract}

\section{Introduction}
\label{introduction}
For a neural network performing image classification, can we reconstruct the training image given its gradients, its label and the model architecture? More precisely, suppose we know the model $f(\bm{x};\bm{w})$ with parameters $\bm{w}$ that are given by either random initialisations or pre-training. Let $\bm{x}^*$ be the training image we wish to reconstruct and assume its label $\bm{y}^*$ is known. Furthermore, let $\CL$ be the loss function for our classification problem. In mathematical terms, we are interested in inverting the mapping from $\bm{x}^*$ to its gradients $\nabla_{\bm{w}} \CL(f(\bm{x}^*;\bm{w}),\bm{y}^*)$. This mapping is far from being injective in general, which makes this problem challenging. Besides being an interesting inverse problem, this problem has important implications in understanding the privacy risk in Federated Learning. For example, in the scenario where the local participants have sensitive data that are not to be shared with other participants or the central server, how can we guarantee that their data cannot be reconstructed from the gradients that are shared during collaborative training?

A number of works have tried to solve this reconstruction problem, which we refer to as `gradient-leakage attacks'. Although each of these works can perform well in specific situations, it remains unclear why they performed well in those situations and why they failed in other ones. Our goal in this paper is to provide a novel framework to understand the efficacy and failure of existing gradient-leakage attacks, so that we can get more insight into the nature of gradient leakage in deep learning. 
The main insight from our work is that the analytic method in \cite{zhu2021rgap} and the optimisation method in \cite{dlg2019} can be unified under one framework (i.e. our hybrid framework) to understand gradient leakage attacks and shed light on the issue of gradient leakage itself. More precisely, we think the R-GAP method in \cite{zhu2021rgap} is essentially solving a least square problem by picking a particular representative in the family of solutions obtained from the Singular Value Decomposition (SVD). We claim that the idea of gradient matching proposed in \cite{dlg2019} can be used to choose a better representative from the family of solutions than R-GAP. In this way, these two seemingly distinct methods can be viewed as two consecutive steps in solving the same least square problem.
We hope that our framework to analyse gradient leakage can lead to further theoretical works on for example, proofs of the likelihood of a successful gradient-leakage attack for a given deep neural network.

\paragraph{Our contribution:}
Following the work in \cite{zhu2021rgap}, we formulate the reconstruction problem as solving one linear system at each layer, iterating backward over the entire network. The linear system at each layer is defined by the forward and backward propagation of the target image during training. The linear system can only be solved approximately in general due to the size and rank of the coefficients. To reduce the approximation error of the solution, we propose and solve an optimisation problem using the idea of gradient matching inspired by the work in \cite{dlg2019}: optimising the approximated solution so that the gradients of the loss function are close to those of the loss function evaluated at the target image according to the cosine distance function. By our formulation, the reconstruction is primarily determined by linear systems, which enables us to estimate the vulnerability of a deep network against gradient leakage attacks aiming at reconstructing the training image. We quantify this estimate by a novel metric, which is defined as a sum of rank-deficiency of the linear system at each layer, weighted by its position in the network. We apply our framework to convolutional networks with the number of layers ranging from two to four, which include either randomly initialised or pre-trained weights. The results have shown noticeable improvements over previous works.

\section{Related Work}
In \cite{dlg2019}, it is shown that one can reconstruct the training image fully for a single image in an untrained model performing classification tasks. In order to reconstruct a target image-label pair $\bm{x}^*, \bm{y}^*$ for an untrained model with randomly initialised weights $\bm{w}$, their method (which was referred to by the authors as `Deep Leakage from Gradients' or DLG) proceeds as follows. Start with a randomly initialised image-label pair $(\bm{x},\bm{y})$, minimise the $L2$ distance between the gradients from $(\bm{x},\bm{y})$ and those from the target pair $(\bm{x}^*, \bm{y}^*)$ by changing $(\bm{x},\bm{y})$. The solution to this optimisation will be the desired reconstruction of the target $(\bm{x}^*, \bm{y}^*)$. Although this method is able to reconstruct the target image and its label in high accuracy in some cases, it is prone to failure when we consider different target pairs or model architectures. Also, it is unable to handle pre-trained models. The work \cite{zhao2020idlg} provides an improvement to \cite{dlg2019} by showing that one can always reconstruct the true label of the target image first so that the optimisation only needs to run with respect to the dummy input $\bm{x}$. This has improved the robustness of DLG. A further improvement is provided by \cite{geiping2020inverting}, which replaces the $L2$ distance in DLG by a combination of cosine similarity and a Total Variation $L1$ regulariser. Recently, \cite{LiAuditingPrivacy} improves upon previous work by making use of GAN that has been trained on public datasets to provide image priors to guide the optimisation.

In a different line of thinking from the above formulation as optimisation problems, \cite{aono2017privacy} showed one can invert a fully connected layer if the bias term is nonzero. The input to the layer can be expressed in closed form using gradients of the weights and of the bias. We make use of this idea and show that it is possible to extend this solution to the case when the bias is zero. Unfortunately the same close-form solution cannot be extended directly to the case of a convolutional layer, because the convolutional operation involves weight sharing in general. Fortunately, we can represent the convolution operation in a circulant representation described in \cite{golub2013matrixchap482}. In this way, convolution can be expressed as a single matrix multiplication, like in a fully-connected layer. Note that the circulant representation does not change the nature of weight sharing in a convolutional layer, so we cannot use the same method from the case of a fully-connected layer. This is in contrast to the view in \cite{Fan2020}, who treated both cases in a uniform manner. 

Based on the circulant representation of convolution, \cite{zhu2021rgap} formulated a linear system of the input to each layer using the forward and backward propagation. By formulation, the training input must satisfy this linear system. Then by solving this linear system at each layer iteratively using pseudo inverse, they showed that we can ultimately solve the reconstruction problem. Furthermore, they formulated a so-called `virtual constraint' to capture additional constraints that are not captured by the linear system at each layer. Unfortunately, they were not able to incorporate the `virtual constraint' into their solution. We think that the `virtual constraint' becomes redundant if we solve the linear systems jointly across the entire network. Empirically, we also found that there has been very little improvement over the reconstructions by incorporating the `virtual constraint' into the objective function. In our work, we combine the linear system formulation from \cite{zhu2021rgap} with the idea of gradient matching from \cite{dlg2019} and its improvement from \cite{geiping2020inverting}. On the one hand, our approach can correct errors inherent in solving the least square problem defined by the linear system in \cite{zhu2021rgap}. On the other hand, our approach improves gradient matching by introducing strong constraints defined by the linear systems. Unlike \cite{LiAuditingPrivacy}, we do not assume additional prior knowledge of the image we want to reconstruct. Our formulation of the reconstruction problem provides a framework for us to analyse the vulnerability of a deep network against gradient-leakage attacks through its architecture, which is not available through purely optimisation-based methods. 
Based on our framework, we propose a quantifying metric \eqref{rankdefIndex} to measure the likelihood of a successful reconstruction. Notice that \cite{zhu2021rgap} proposed a similar metric `RA-i'. Our metric differs from `RA-i' in three ways: 1. We do not incorporate the `virtual constraint' based on our observation above; 2. Our metric is defined based on the rank of the linear system instead of the number of constraints; 3. We quantify the position of the layer by taking a weighted average of the rank-deficiency of the linear system at each layer.

\section{A hybrid framework}

\paragraph{Overall assumptions:}
We only consider the problem of reconstructing a single training image when the batch size is one. The target model for image-classification is assumed to consist of consecutive convolutional layers with the last layer being fully connected. We have considered 2, 3 and 4 layer CNNs with variations which represent different cases of changes in intermediate feature spaces. 
Our framework and analysis should apply to deeper networks composed of concatenated shallow building blocks of CNNs, as we have specified an exhaustive set of those building blocks for CNNs. The quality of reconstruction can be estimated by the average of the rank deficiency from each layer weighted by the position of the layer in the network.

We assume the ground-truth label of the target image is given. This will not lose generality, because otherwise we can reconstruct the true label via observing the signs of the gradients of the weights in the fully connected layer according to \cite{zhao2020idlg}. We assume that the activation function for each layer is piecewise invertible and piecewise differentiable. For simplicity we do not include pooling in the design of the network. Notice that average pooling can be regarded as convolution. Also notice that CNN without pooling was considered in \cite{springenberg2015striving}.

\paragraph{Notations}
We introduce notations used throughout the paper.
\begin{enumerate}[wide, labelindent=0pt, itemsep=-1mm]
    \item[$\bm{w}^{(i)}$:] weight from layer $i$ of size $m$ by $n$, where $0 \leq i \leq d$ and $d$ is the total number of layers. For simplicity of notation, we omit $i$ in $m$ and $n$ when possible, but readers should be aware that the size of the weight need not be the same for different layers. For a convolutional layer, it denotes the circulant representation of the kernel following \cite{golub2013matrixchap482}.
    \item[$\CL(\bm{x},\bm{y};\bm{w})$:] Cross entropy loss function of the network with input image $\bm{x}$, label $\bm{y}$ and weight $\bm{w}$. We use $\CL^{(i)}(\bm{x},\bm{y};\bm{w})$ for the shorthand notation denoting the loss with the truncated model starting from layer $i$ with corresponding intermediate input $\bm{x}^{(i)}$, weight from the $i$-th layer onward. We will omit the label $\bm{y}$ where possible. 
    \item[$\bm{z}^{(i)}$:] the linear output of the layer $i$ before activation given by $\bm{w}^{(i)} \bm{x}^{(i)} + \bm{b}^{(i)}$ with input $\bm{x}^{(i)}$ and bias $\bm{b}^{(i)}$. This also expresses the convolutional operation following the circulant form of $\bm{w}^{(i)}$.
    \item[$\bm{\alpha}^{(i)}(\cdot)$:] activation function after linearity in vector form. We use the unbold letter $\alpha^{(i)}$ to denote its component. 
    \item[$|.|$:] when applied to a matrix, the absolute value sign $|.|$ denotes the number of elements.
    \item[$w,x,z,b$:] we use unbold letters with subscripts to denote the component at specified indices of the corresponding matrix in bold letters.
\end{enumerate}
To reconstruct the input image, we adopt an iterative approach similar to \cite{zhu2021rgap}: starting from the label, we reconstruct the input to the last layer and repeat this procedure layer by layer, each time making use of the reconstructed input to the succeeding layer. We will treat the cases of a fully-connected layer and a convolutional layer separately. First we treat the forward and backward pass in training a neural network as imposing two linear constraints on the input to each layer.  

\paragraph{Weight and gradient constraints:}
At a given layer $i$, the forward and backward propagation gives rise to the following equations:

\begingroup
\vspace{-4mm}
\small
\begin{subequations}
\begin{align}
        & \bm{w}^{(i)} \bm{x}^{(i)} + \bm{b}^{(i)} = \bm{z}^{(i)} \label{weightconstraint},  \\
        & \nabla _{\bm{z}^{(i)}} \CL \cdot \bm{x}^{(i)}  = \nabla _{\bm{w}^{(i)}} \CL \label{gradconstraint}. 
\end{align}
\end{subequations}
\endgroup
We note that this represents both the fully-connected and the convolutional cases, using the circulant representation for the weight $\bm{w}^{(i)}$ and the gradient $\nabla _{\bm{z}^{(i)}} \CL$. Both $\bm{z}^{(i)}$ and $\nabla _{\bm{w}^{(i)}} \CL$ are written as vectors. From the reconstruction point of view, we treat $\bm{x}^{(i)}$ as the unknown and regard the above equations as weight and gradient constraints imposed on the unknown. The term $\bm{z}^{(i)}$ is computed from inverting the reconstruction from the subsequent layer $\bm{x}^{(i+1)}$ using inverse of the activation function:

\begingroup
\vspace{-2mm}
\small
\begin{equation}\label{getZbInvertingX}
    \bm{z}^{(i)} = (\bm{\alpha}^{(i)})^{-1}(\bm{x}^{(i+1)}).
\end{equation}
\endgroup
The term $\nabla _{\bm{z}^{(i)}} \CL$ can be computed by using the following relations deduced from backpropagation:

\begingroup
\vspace{-4mm}
\small
\begin{subequations}\label{gradrelations}
\begin{align}
    & \nabla _{\bm{z}^{(i)}}\CL = \nabla_{\bm{x}^{(i+1)}}\CL \cdot \nabla_{\bm{z}^{(i)}} \bm{\alpha}^{(i)} \label{gradrelationsZ}, \\
    & \nabla_{\bm{x}^{(i)}} \CL = \nabla_{\bm{x}^{(i+1)}} \CL \cdot \nabla_{\bm{z}^{(i)}} \bm{\alpha}^{(i)} \cdot \bm{w}^{(i)} \label{gradrelationsX}.
\end{align}
\end{subequations}
\endgroup
We notice that the circulant representation of the gradient $\nabla _{\bm{z}^{(i)}} \CL$ is determined by the circulant form of the weight $\bm{w}^{(i)}$ from backpropagating through the weight constraint \eqref{weightconstraint}.

\subsection{Fully connected layer}
For a fully connected layer, the input can be solved uniquely in closed form. We summarise the solution in the following lemma. The case of non-zero bias is due to \cite{aono2017privacy} and we show that it can be extended to the general case. Please see Supplementary~\ref{app:proof} for the proof.
\begin{lemma}\label{lemma}
If a fully connected layer has non-zero bias $\bm{b}^{(i)} = (b_1 ^{(i)},...,b_m ^{(i)}) \in \BR^m$, then the input $\bm{x}^{(i)} = (x_1 ^{(i)},...,x_n ^{(i)}) \in \BR^n$ is uniquely determined from the gradient constraint \eqref{gradconstraint}. Suppose $\exists k, 1 \leq k \leq m$, such that $b_k ^{(i)}\neq 0$ and $\frac{\partial \CL}{\partial b_k ^{(i)}} \neq 0$. Then $\bm{x}^{(i)}$ is given by:

\begingroup
\small
\begin{equation}
            x_l ^{(i)}= \frac{\partial \CL}{\partial w_{kl} ^{(i)} } \left (\frac{\partial \CL}{\partial b_k ^{(i)} } \right )^{-1}, \ \ 1 \leq l \leq n.
\end{equation}
\endgroup

More generally, assuming both $\frac{\partial \CL}{\partial x_k ^{(i+1)}}$ and $\frac{\partial \alpha_k ^{(i)}}{\partial z_k ^{(i)}}$ are nonzero, we have 

\begingroup
\small
\begin{equation}\label{fullyConnectedGeneral}
    x_l ^{(i)} = \frac{\partial \CL}{\partial w_{kl} ^{(i)}} \left (\frac{\partial \CL}{\partial x_k ^{(i+1)}} \right )^{-1} \left (\frac{\partial \alpha_k ^{(i)}}{\partial z_k ^{(i)}} \right )^{-1}.
\end{equation}
\endgroup
\end{lemma}
\begin{remark}
The reason that we can solve for $\bm{x}$ in closed form described above is essentially because there is no weight sharing in a fully-connected layer. This implies that the circulant form of $\nabla _{\bm{z}^{(i)}} \CL$ consists of blocks of diagonal matrices with the same element along the diagonal in each block, which allows \eqref{gradconstraint} to be solved exactly. 
\end{remark}

\subsection{Convolutional layer}
For a convolutional layer, we can no longer uniquely determine the input in general because of weight sharing. We first build on the work in \cite{zhu2021rgap} and formulate a linear system by combining the weight and gradient constraints from \eqref{weightconstraint} and \eqref{gradconstraint} into a single linear system of the input $\bm{x}$. Define $\bm{u}^{(i)}, \bm{v}^{(i)}$ to denote the following block matrices

\begingroup
\small
\begin{equation}\label{DefUandV}
    \bm{u}^{(i)} := \begin{bmatrix}
                 \bm{w}^{(i)} \\
                 \nabla _{\bm{z}^{(i)}} \CL
                \end{bmatrix},
    \bm{v}^{(i)} := \begin{bmatrix}
                (\bm{\alpha}^{(i)})^{-1}(\bm{x}^{(i+1)}) \\
                \nabla _{\bm{w}^{(i)}} \CL
                \end{bmatrix}.
\end{equation}
\endgroup
Here both the term $(\bm{\alpha}^{(i)})^{-1}(\bm{x}^{(i+1)})$ and the term $\nabla _{\bm{w}^{(i)}} \CL$ are written as vectors. The term $\nabla _{\bm{w}^{(i)}} \CL$ has the same dimension as $|\bm{w}^{(i)}|$, i.e.~the number of elements of the weight in its non-circulant form as a 4-dimensional array.

Notice that we can absorb the bias term into the product $\bm{w}^{(i)} \bm{x}^{(i)}$ by replacing the the weight matrix with its augmentation by $\bm{b}^{(i)}$ and $\bm{x}^{(i)}$ by its augmentation by one. Adopting this notation, \eqref{weightconstraint} and \eqref{gradconstraint} can be written as:

\begingroup
\vspace{-2mm}
\small
\begin{equation}\label{WeightAndGradConstraintCombined}
    \bm{u} ^{(i)} \bm{x}^{(i)} - \bm{v}^{(i)} = 0.
\end{equation}
\endgroup
Recall that under our assumptions and \eqref{getZbInvertingX}, we can get $\bm{z}^{(i)}$ that defines $\bm{v}^{(i)}$ by inverting the solution to the reconstruction problem for the following layer and other coefficients in $\bm{u} ^{(i)}$ and $\bm{v}^{(i)}$ are given from training. Based on \cite{zhu2021rgap}, we view \eqref{WeightAndGradConstraintCombined} as a linear system for the input $\bm{x}^{(i)}$ to the current layer $i$. In general, $\bm{u} ^{(i)}$ may not be square and it can be rank-deficient, which means we can only get an estimated solution using the pseudo inverse of $\bm{u} ^{(i)}$ . Proceeding layer-by-layer in this manner, we can obtain a reconstruction of the input to the network. This is the R-GAP approach introduced in \cite{zhu2021rgap}. We think there are several sources of error in the solution using this approach alone:
\begin{enumerate}[itemsep=-1mm]
	\item The linear system \eqref{WeightAndGradConstraintCombined} is defined using estimated value of $\bm{z}$ by inverting the solution from the succeeding layer. So the error from reconstructing the input to the succeeding layer will carry over to the current layer.
	\item Pseudo inverse to a linear system is not unique in general and a minimum norm solution might not be the best one in our reconstruction problem. 
	\item Bad conditioning of $\bm{u} ^{(i)}$ can contribute to the error of the estimated solution.
\end{enumerate}
To tackle these issues, we propose a correction procedure based on the idea of gradient matching.

\paragraph{Correcting the approximated solution:}
From the theory of linear systems, we know that we can get an approximated solution to a linear system such as  \eqref{WeightAndGradConstraintCombined} by solving a least square problem:
\begin{equation}\label{LeastSquareProbUV}
    \argmin_{\bm{x}} ||\bm{u}^{(i)} \bm{x} - \bm{v}^{(i)} ||^2.
\end{equation}
The solution is not unique without requiring the norm of the solution to be minimal: the sum of a given solution with another vector $\bm{x_0}$ such that $\bm{u}^{(i)} \bm{x_0} = 0$ is still a solution. On the other hand, the solution with the minimum norm may not be the most accurate one for our reconstruction problem. Assuming $\bm{x}_{LS}$ is the minimum norm solution to the least square problem \eqref{LeastSquareProbUV} obtained by using the Singular Value Decomposition (e.g. by Theorem 5.5.1 in \cite{golub2013matrix}), we propose to correct it using the idea of gradient matching from \cite{dlg2019}. More precisely, after we have obtained a solution $\bm{x}_{LS} ^{(i)}$ at a layer, we formulate and solve the following optimisation problem:

\begingroup
\vspace{-5mm}
\small
\begin{equation}\label{LayerwiseOptProb}
     \argmin_{\bm{x}} \mathcal{D} \left [ \nabla_{\bm{w}}\CL^{(i)}(\bm{x};\bm{w})\vert_{\bm{w} = \bm{w}^*},\nabla_{\bm{w}}\CL^{(i)}(\bm{x}_{true}; \bm{w})\vert_{\bm{w} = \bm{w}^*} \right ], \ \ 
     \text{subject to \ \ } \bm{u}^{(i)} \bm{x} - \bm{v}^{(i)} = 0.
\end{equation}
\endgroup
where $\bm{x}_{true}$ is the target image, $\bm{w}^*$ are given weights and $\mathcal{D} \left [\cdot,\cdot \right]$ is a chosen distance function. Instead of taking $\mathcal{D} \left [\cdot,\cdot \right]$ to be the $L2$-norm as in \cite{dlg2019}, we adopt the cosine distance function proposed in \cite{geiping2020inverting} because we believe it is less sensitive to the stage of training. More precisely, we define $\mathcal{D} \left [\cdot,\cdot \right]$ to be

\begingroup
\vspace{-2mm}
\small
\begin{equation}
    \mathcal{D} \left [\bm{x}_1,\bm{x}_2 \right] :=
    1 - \frac{\langle \bm{x}_1, \bm{x}_2 \rangle}{||\bm{x}_1||\cdot ||\bm{x}_2||},
\end{equation}
\endgroup
for $n$-dimensional vectors $\bm{x}_1$ and $\bm{x}_2$, where $\langle \cdot, \cdot \rangle$ and $||\cdot||$ are the Euclidean inner product and norm respectively.

In numerical experiments, we find it helpful in terms of reconstruction quality to add total variation to the objective function \eqref{LayerwiseOptProb}. The optimisation problem described by \eqref{LayerwiseOptProb} will now become:

\begingroup
\vspace{-2mm}
\footnotesize
\begin{equation}\label{LayerwiseOptProb2}
         \argmin_{\bm{x}} \Big \{ \mu_1 \mathcal{D} \left [ \nabla_{\bm{w}}\CL^{(i)}(\bm{x};\bm{w})\vert_{\bm{w} = \bm{w}^*},\nabla_{\bm{w}}\CL^{(i)}(\bm{x}_{true};         \bm{w})\vert_{\bm{w} = \bm{w}^*} \right ] 
         + \mu_2 \text{TV}(\bm{x}) \Big \}, \ \ 
         \text{subject to \ \ } \bm{u}^{(i)} \bm{x} - \bm{v}^{(i)} = 0. 
\end{equation}
\endgroup

where $\mu_1,\mu_2 \in \BR$ are some given weights. Performing the correction described in \eqref{LayerwiseOptProb2} at each convolutional layer, we have a hybrid method to reconstruct the input presented in Algorithm \ref{hybridv1}. We observe that if we turn the hard constraint in \eqref{LayerwiseOptProb2} into a soft one, the algorithm will converge more quickly. More details are provided in section \ref{app:convergence_algo}.
\begin{algorithm}[tb]
  \footnotesize
  \caption{Hybrid method.}
  \label{hybridv1}
\begin{algorithmic}
  \STATE {\bfseries Input:} Number of layers $d$ of the network; True label $\bm{y}$ of the target image $\bm{x}_{true}$; Initial weights $\bm{w}^*$; Gradients $\nabla_{\bm{w}}\CL^{(i)}(\bm{x};\bm{w})\vert_{\bm{w} = \bm{w}^*}$ at each layer $i, 0 \leq i \leq d-1$; Number of iterations $N^{(i)}$ at each layer $i$.
  \STATE Initialise $\overline{\bm{x}^{(d)}} = \bm{y}$.
  \FOR{$i=d-1$ {\bfseries to} $0$ \COMMENT{iterate backward from the last layer of the network}}   
      \STATE Compute the gradient $\nabla_{\bm{x}^{(i+1)}} \CL(\bm{x}_{true};\bm{w}^*) \big \vert_{\bm{x}^{(i+1)} = \overline{\bm{x}^{(i+1)}}}$ using \eqref{gradrelationsX} and $ \overline{\bm{x}^{(i+1)}}$. 
      \STATE Compute $\nabla_{\bm{z}^{(i)}}\CL(\bm{x}_{true};\bm{w}^*) \big \vert_{\bm{z}^{(i)} = (\bm{\alpha}^{(i)})^{-1}(\overline{x^{(i+1)}})}$ from $\nabla_{\bm{x}^{(i+1)}}\CL$ using \eqref{gradrelationsZ}.
	    \IF{the current layer is fully connected} \STATE solve for $\overline{\bm{x}^{(i)}}$ using \eqref{fullyConnectedGeneral}.
	    \ELSIF{the current layer is convolutional}
	         \STATE Define $\bm{u}^{(i)}, \bm{v}^{(i)}$ from \eqref{DefUandV} using $ (\bm{\alpha}^{(i)})^{-1}(\overline{\bm{x}^{(i+1)}})$ and gradients of $\CL$ computed above.
	         \STATE Get an estimate $\bm{x}^{(i)} _{LS}$ of the input to layer $i$ by solving the linear system $\bm{u}^{(i)}\bm{x} - \bm{v}^{(i)} =0$.
	         \STATE Get a corrected estimate $\overline{\bm{x}^{(i)}}$ based on $\bm{x}^{(i)} _{LS}$ by solving the optimisation problem \eqref{LayerwiseOptProb2} with initialisation $\bm{x}^{(i)} _{LS}$ for $N^{(i)}$ iterations. \COMMENT{{only compute and use gradients from the current layer to the last one}}
	   \ENDIF
  \ENDFOR
  \STATE {\bfseries Output:} Reconstruction $\overline{\bm{x}^{(0)}}$ of the target $\bm{x}$.
\end{algorithmic}
\end{algorithm}

\paragraph{A security measure:}
In light of the hybrid framework given for a convolutional layer, the problem of reconstructing a training image can be viewed as consisting of two parts : \textbf{i.} An iterative procedure starting from the output of the network ; \textbf{ii.} at each layer, solving a linear system with corrections using gradient matching when the layer is convolutional. 
Based on this insight, we define a metric that measures the efficacy of the hybrid method given by Algorithm \ref{hybridv1}, which depends partially on the architecture of the target model.  
\begin{definition}\label{def_secure_metric}
Suppose the model $\CM$ has $d$ convolutional layers indexed by $1, ... , d$, followed by a fully-connected layer. We define the following metric:

\begingroup
\small
\begin{equation}\label{rankdefIndex}
    c(\CM) : = \sum_{i = 1} ^d \frac{d-(i-1)}{d} \cdot \big( \rank(\bm{u}^{(i)}) - n_i \big),
\end{equation}
\endgroup
where $\bm{u}^{(i)}$ is defined in \eqref{DefUandV} and $n_i$ is the dimension of the input for the $i$-th layer as a vector.
\end{definition}
Because $\rank(\bm{u}^{(i)}) \leq n_i$ for each convolutional layer, $c(\CM)$ will be non-positive. The larger the value of the metric is, the less secure the model tends to be and the more likely it is to create better reconstructions. The metric is better interpreted as an estimate of the security of the model against the hybrid method. Our experiments have shown that it is possible to fully reconstruct the input to a model $\CM$ using our method when $c(\CM) = 0$. For more details on the thinking behind Definition \ref{def_secure_metric}, please refer to Section \ref{append:justification_for_metric} in the Supplementary.

\paragraph{Recommendations on architectural design:}
Based on the proposed metric and analysis, we can see that a network tends to be less secure against our method if it has wider convolutional layers that greatly increase the dimensions of the feature spaces at the beginning of the network and only shrinks the feature spaces towards the final layer, compared to other designs. 
Overall, we would recommend designing a model with a small value of $c(\CM)$ if defendability against gradient-leakage attacks is the main concern. 

\section{Experiments}
We demonstrate the performance of our hybrid method and how our proposed index $c(\CM)$ in \eqref{rankdefIndex} can be an indicator of model security in practice. We consider a series of shallow architectures performing classification on CIFAR-10 provided by \cite{krizhevsky2009learning} and \cite{cifar10license}. These architectures consist of two, three and four-layer convolutional networks. For simplicity, we assume the convolutional layer to be bias-free and the fully-connected layer to have non-zero bias. However, these assumptions on bias are not necessary for our method to work. In order to illustrate typical architecture designs for the network, we consider several variations of the network, each representing a different case of changes in dimensions of intermediate feature spaces. 
\begin{table*}[!ht]
		\centering
		\caption{\footnotesize Model architecture for all variants of the models, rank deficiency `rd' for each layer and values of metric $c(\CM)$. The numbers in columns `layer x' refer to kernel width, channels, strides and padding in order. The numbers in the column `fully connected' refer to the input dimension of that fully-connected layer, whereas the output dimension is always 10. The values of $c(\CM)$ are the same between two images used in the experiment.}\label{tbl:modelArch}
		
		\vspace{8pt}
		\scalebox{0.7}{%
		\begin{tabular}{lcccccccc}
		\toprule
		     & Layer 1 & Layer 2 & Layer 3 & Fully Connected & rd1 & rd2 & rd3 & $c(\CM)$  \\
		     \midrule
		     CNN2 Variant 1 & 3,6,1,0 & /       & / & 5400 & 0      & /       & / & \phantom{000}0 \\ 
		     CNN2 Variant 2 & 4,6,2,0 & /       & / & 1350 & -1470  & /        & / & -1470 \\
		     CNN3 Variant 1 & 3,6,1,0 & 4,3,2,0 & / & \phantom{0}588 & 0      & -4533   & / & -2266 \\
		     CNN3 Variant 2 & 4,6,2,0 & 3,3,2,0 & / & \phantom{0}147 & -1470  & -1050     & / & -1995 \\
		     CNN3 Variant 3 & 3,6,1,0 & 3,9,1,0 & / & 7056 & 0      & 0         & /  & \phantom{-000}0 \\
		     CNN3 Variant 4 & 3,1,1,0 & 3,6,1,0 & / & 4704 & -2146  & 0         & /  & -2146 \\
		     CNN4 Variant 1 & 3,6,1,0 & 4,5,2,0 & 4,3,1,0 & \phantom{0}363 & 0      & -3965         & -386  & -2772 \\
		     CNN4 Variant 2 & 5,16,1,0 & 5,6,2,0 & 5,32,1,2 & 4608 & 0      & -9316          & 0    & -6211 \\
		     \bottomrule
		\end{tabular}
		}
\end{table*}
\begin{figure*}[!ht]
		\centering
		%
		%
		\begin{subfigure}[h]{0.45\textwidth}
		\begin{minipage}[c]{0.05\linewidth}
		      \rotcaption{\parbox{1.75cm}{CNN2 V1}}\label{fig:cnn2c1}  
		    \end{minipage}
		    \begin{minipage}[c]{0.85\linewidth}
		    \includegraphics[width=\linewidth]{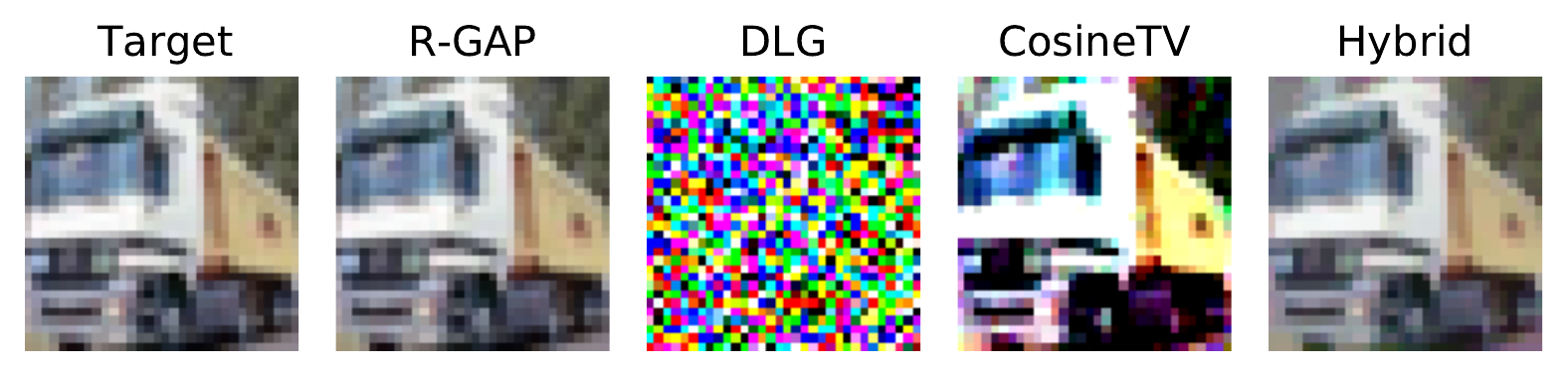}\\
			\includegraphics[width=\linewidth]{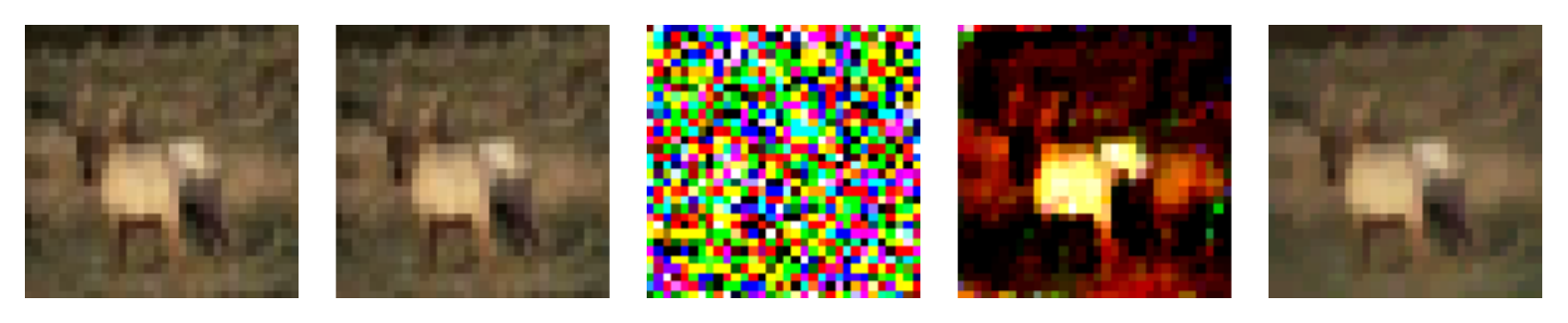}
		    \end{minipage}
		\end{subfigure}\hfill
		\begin{subfigure}[h]{0.45\textwidth}
		\begin{minipage}[c]{0.05\linewidth}
		      \rotcaption{\parbox{1.75cm}{CNN2 V2}}\label{fig:cnn2c2}  
		    \end{minipage}
		    \begin{minipage}[c]{0.85\linewidth}
		    \includegraphics[width=\linewidth]{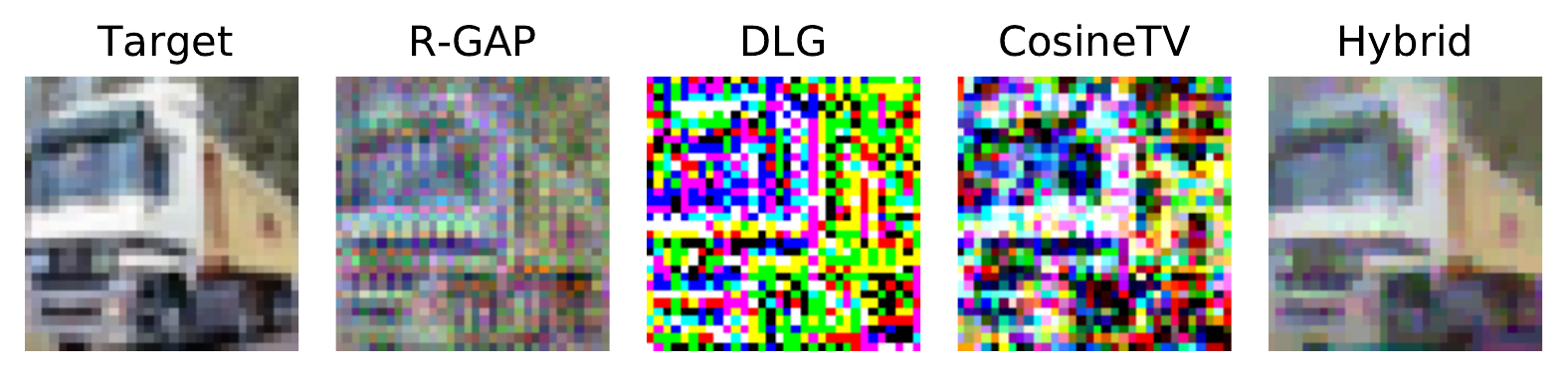}\\
			\includegraphics[width=\linewidth]{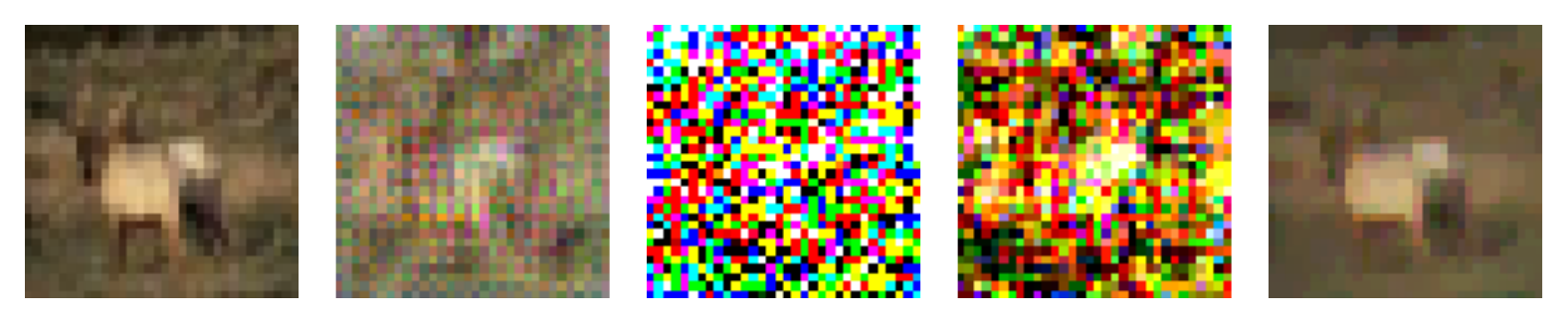}
		    \end{minipage}
		\end{subfigure}\hfill
		\begin{subfigure}[h]{0.45\textwidth}
		\begin{minipage}[c]{0.05\linewidth}
		      \rotcaption{\parbox{1.75cm}{CNN4 V1}}\label{fig:cnn4c1}  
		    \end{minipage}
		    \begin{minipage}[c]{0.85\linewidth}
		    \includegraphics[width=\linewidth]{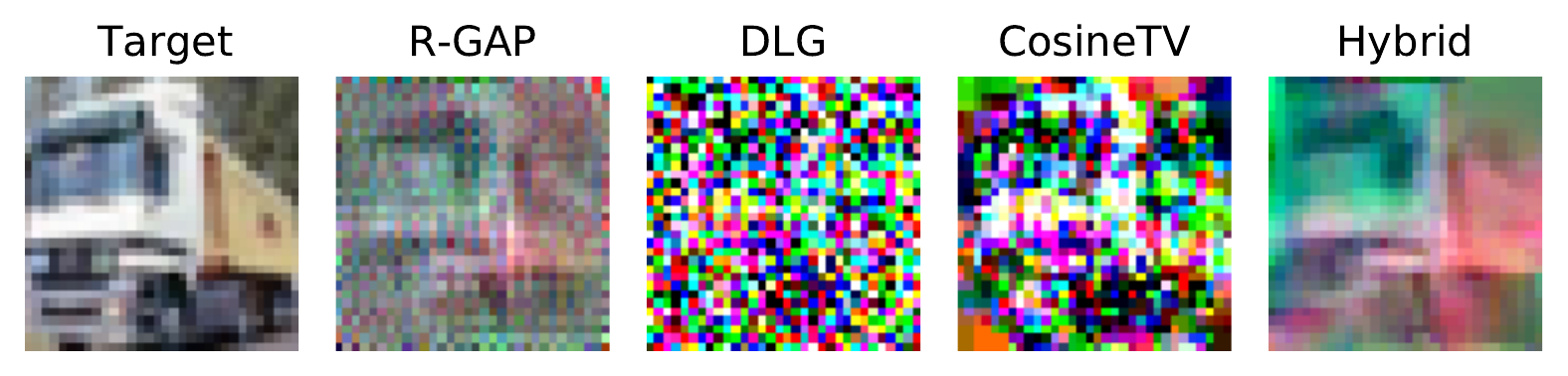}\\
			\includegraphics[width=\linewidth]{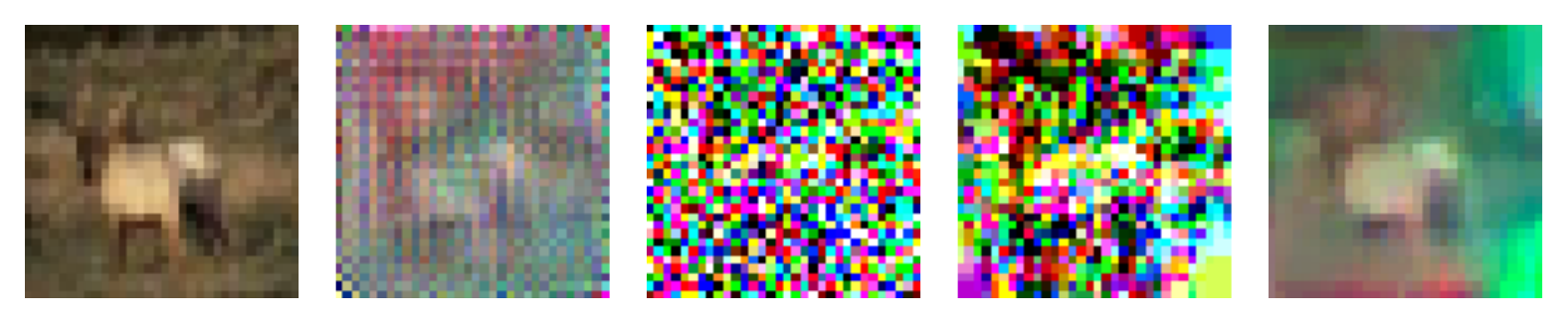}
		    \end{minipage}
		\end{subfigure}\hspace{1.2cm}
		\begin{subfigure}[h]{0.45\textwidth}
		\begin{minipage}[c]{0.05\linewidth}
		      \rotcaption{\parbox{1.75cm}{CNN4 V2}}\label{fig:cnn4c2}  
		    \end{minipage}
		    \begin{minipage}[c]{0.85\linewidth}
		    \includegraphics[width=\linewidth]{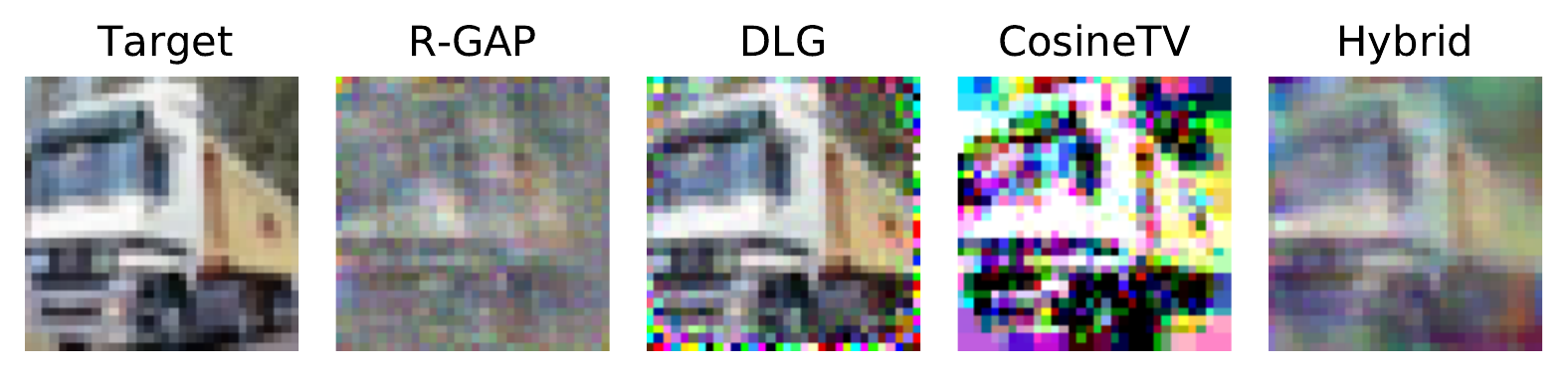}\\
			\includegraphics[width=\linewidth]{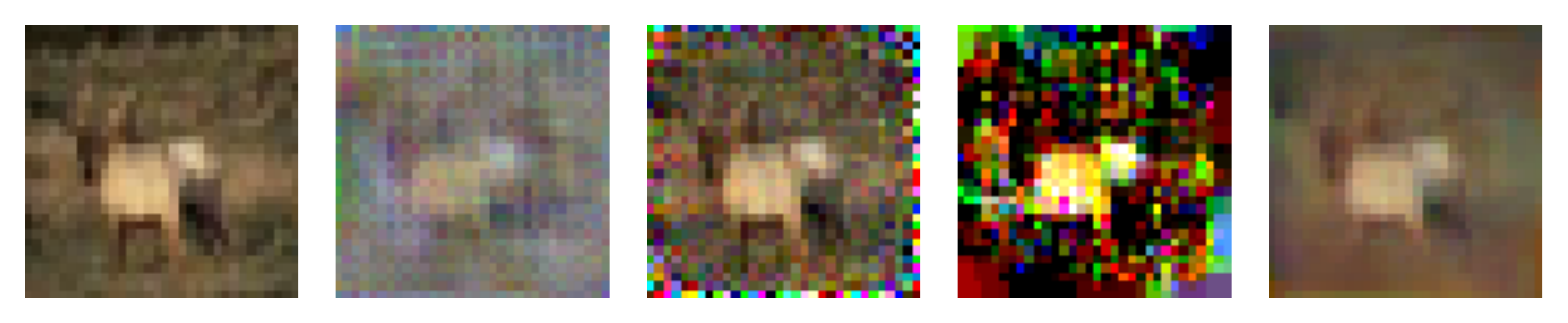}
		    \end{minipage}
		\end{subfigure}
		\vspace{1mm}
		\caption{Comparisons of reconstructions among all approaches for CNN2 and CNN4. Two examples are presented for each architecture. Observe that DLG is unable to reconstruct in all variants in CNN2 and in CNN4 Variant 1, but is able to produce good reconstruction with artefact for CNN4 Variant 2. CosineTV is more stable than DLG while R-GAP performs even more consistently. Our hybrid method improves the results from R-GAP and visually reduces its checkerboard effect and produces results with better overall quality.}
		\label{fig:cnn2andcnn4}
\vspace{-4mm}
\end{figure*}
For each network, weights are initialised randomly from a uniform distribution. The target image will go through one forward and backward pass to generate the gradients. For each convolutional layer, we assume Tanh activation and for each fully-connected layer we assume identity activation.
\begin{figure*}[!ht]
		\centering
		%
		%
		\begin{subfigure}[h]{0.45\textwidth}
		    \begin{minipage}[c]{0.05\linewidth}
		      \rotcaption{\parbox{1.70cm}{Variant 1}}\label{fig:cnn3c1}  
		    \end{minipage}
		    \begin{minipage}[c]{0.85\linewidth}
		    \includegraphics[width=\linewidth]{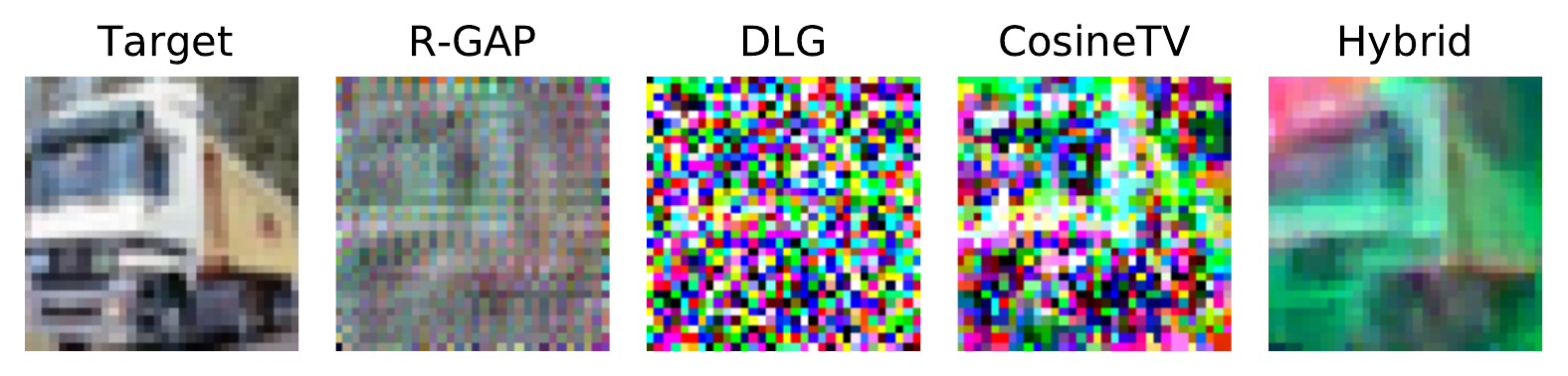}\\
			\includegraphics[width=\linewidth]{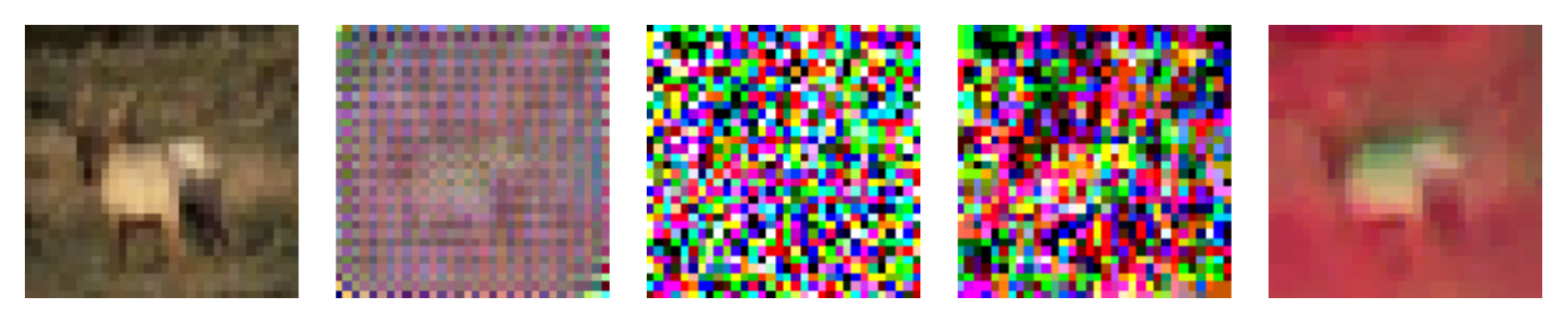}
		    \end{minipage}
		\end{subfigure}\hfill
		\begin{subfigure}[h]{0.45\textwidth}
			\begin{minipage}[c]{0.05\linewidth}
		      \rotcaption{\parbox{1.70cm}{Variant 2}}\label{fig:cnn3c2}  
		    \end{minipage}
		    \begin{minipage}[c]{0.85\linewidth}
		    \includegraphics[width=\linewidth]{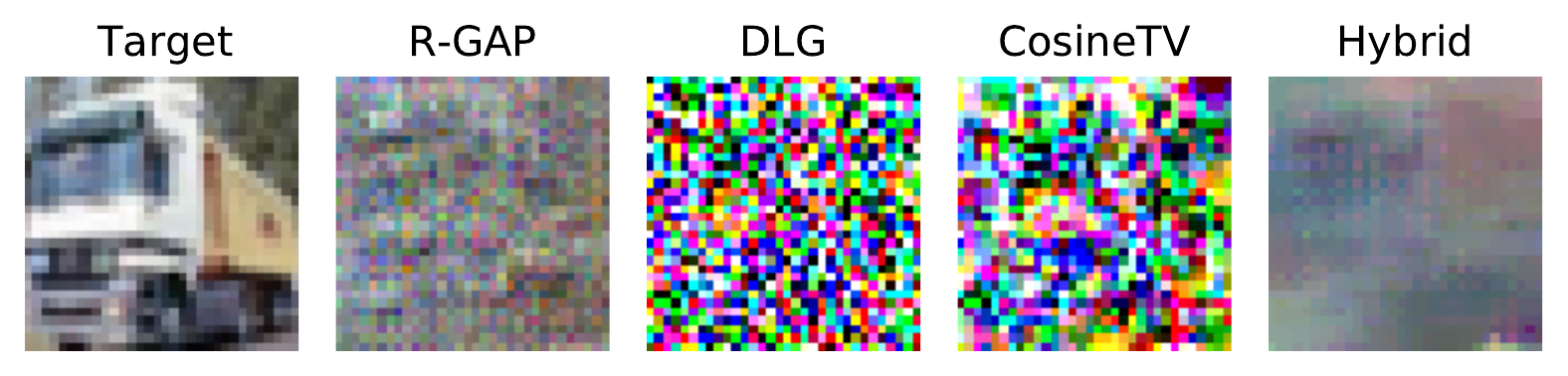}\\
			\includegraphics[width=\linewidth]{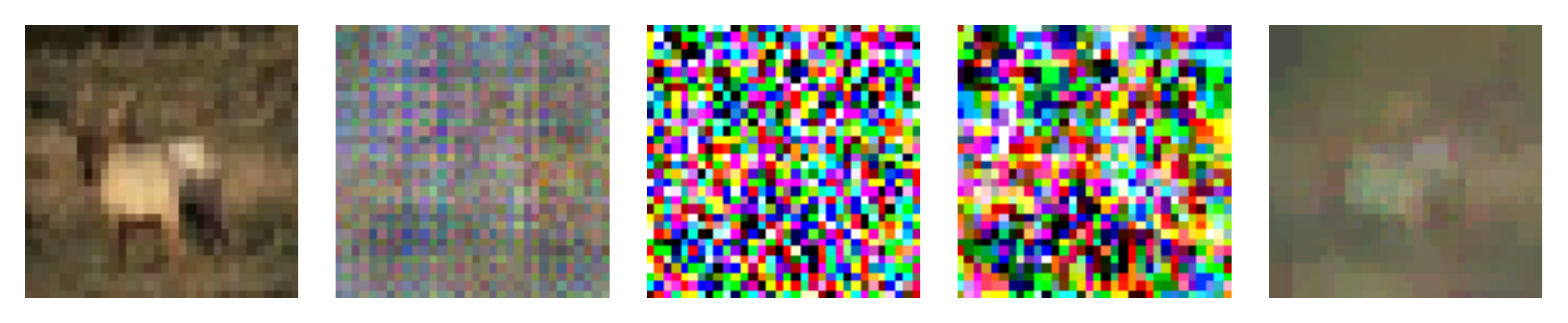}
		    \end{minipage}
		\end{subfigure} \\
		\begin{subfigure}[h]{0.45\textwidth}
		\begin{minipage}[c]{0.05\linewidth}
		      \rotcaption{\parbox{1.70cm}{Variant 3}}\label{fig:cnn3c3}  
		    \end{minipage}
		    \begin{minipage}[c]{0.85\linewidth}
		    \includegraphics[width=\linewidth]{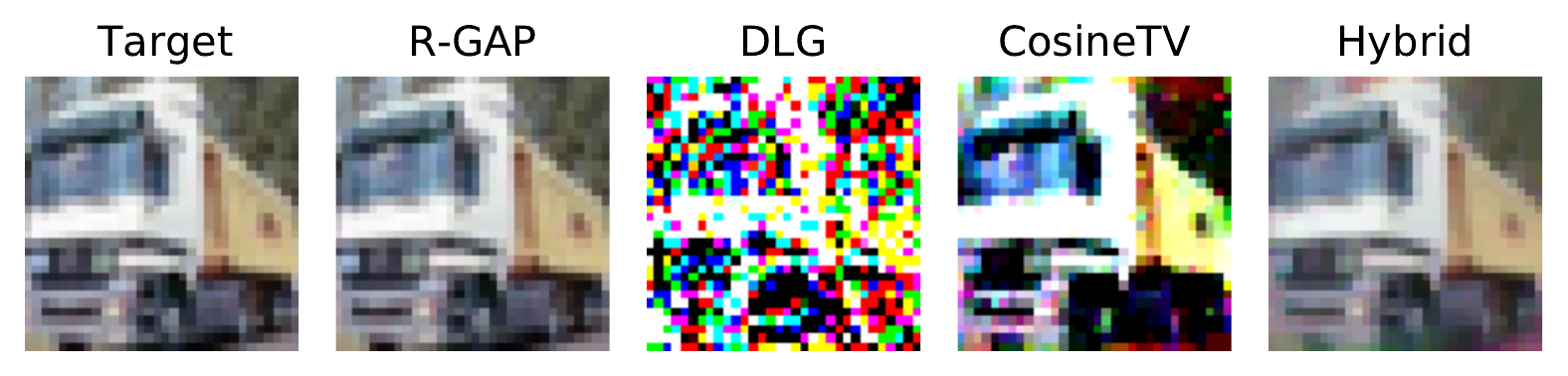}\\
			\includegraphics[width=\linewidth]{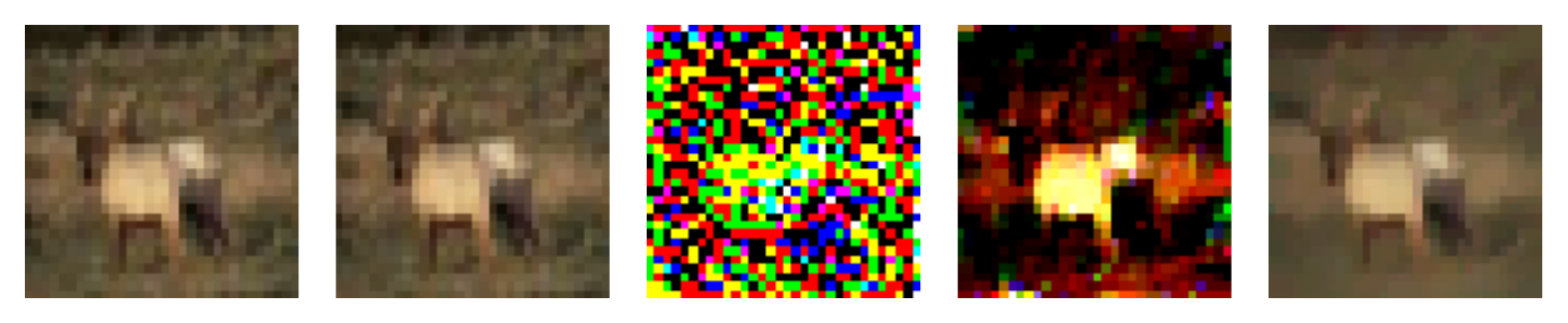}
		    \end{minipage}
		\end{subfigure} \hspace{1.1cm}
		\begin{subfigure}[h]{0.45\textwidth}
		\begin{minipage}[c]{0.05\linewidth}
		      \rotcaption{\parbox{1.70cm}{Variant 4}}\label{fig:cnn3c4}  
		    \end{minipage}
		    \begin{minipage}[c]{0.85\linewidth}
		    \includegraphics[width=\linewidth]{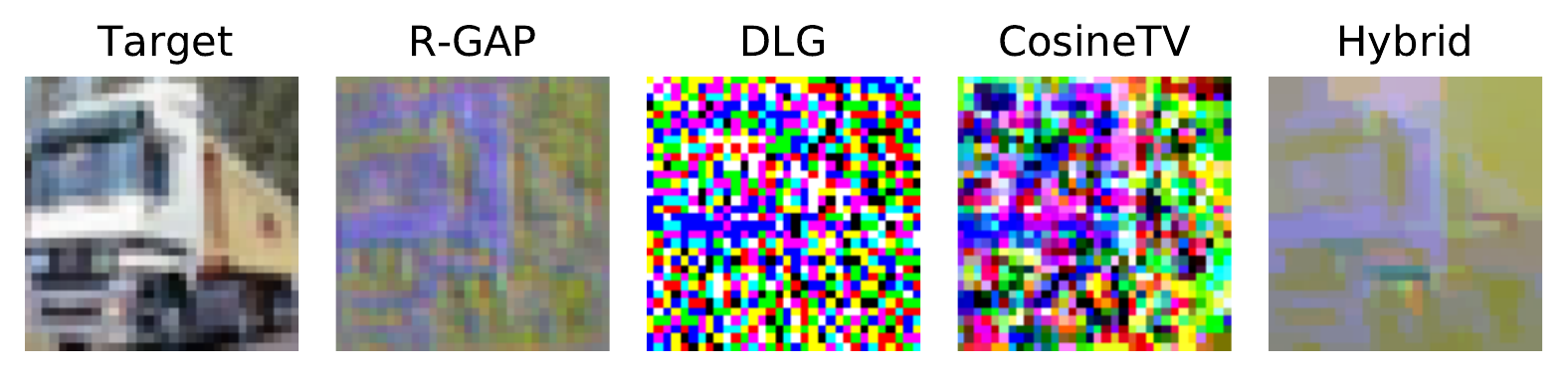}\\
			\includegraphics[width=\linewidth]{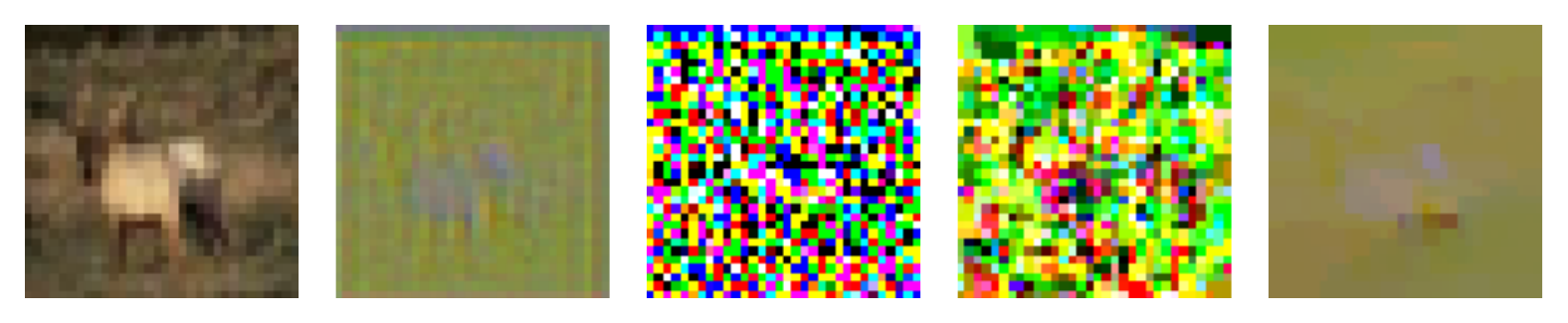}
		    \end{minipage}
		\end{subfigure}
		\vspace{1mm}
		\caption{Comparisons of reconstructions among all approaches for CNN3. Two examples are presented for each architecture. Similar to Figure \ref{fig:cnn2andcnn4}, we notice that DLG and CosineTV show similar performance although CosineTV provides improvement overall. Our hybrid method is consistent with R-GAP but produces smoother results. Also notice that all methods generally produce best results in Figure \ref{fig:cnn3c3} and worst in Figure \ref{fig:cnn3c4}, which is mostly consistent with the value of the metric $c(\CM)$ given in Table \ref{tbl:modelArch}.}
		\label{fig:cnn3}
\end{figure*}
We adopt the unconstrained strategy \eqref{layerwiseOptSoftVersion} from Section \ref{app:convergence_algo} in Algorithm \ref{hybridv1}, using ADAM optimiser from \cite{DBLP:journals/corr/KingmaB14} with default learning rate of $0.001$. The architecture of the models in the experiments are shown in Table \ref{tbl:modelArch}, together with the values of the metric $c(\CM)$. For each variant of the model, we compare our method with that of DLG from \cite{dlg2019}, R-GAP from \cite{zhu2021rgap}, and \cite{geiping2020inverting} (which we name `CosineTV' for short). For details of the number of iterations for all the methods, and other hyperparameter settings in the experiment, please refer to Section \ref{FurtherDetailsExperiment} in the Supplementary.

We present sample outputs in Figure \ref{fig:cnn2andcnn4}, Figure \ref{fig:cnn3} with their MSE and PSNR scores in Table \ref{app:tbl:scores} in the Supplementary. 
Comparing the reconstruction qualities of results in Figure \ref{fig:cnn2andcnn4}, Figure \ref{fig:cnn3} with the values of $c(\CM)$ in Table \ref{tbl:modelArch}, we notice that it is more likely for all methods to have reconstructions with better visual quality on architectures with a bigger value of $c(\CM)$, despite that the relation is not strictly monotonic. 

\begin{figure*}[!ht]
		\centering
		%
		%
		\begin{subfigure}[h]{0.49\textwidth}
		\begin{minipage}[c]{0.15\linewidth}
			\rotcaption{\parbox{1.70cm}{pre-trained Variant 1}}\label{fig:cnn4c1Pretrained}
		\end{minipage}\hfill%
		\begin{minipage}[c]{0.85\linewidth}
			\includegraphics[width=\linewidth]{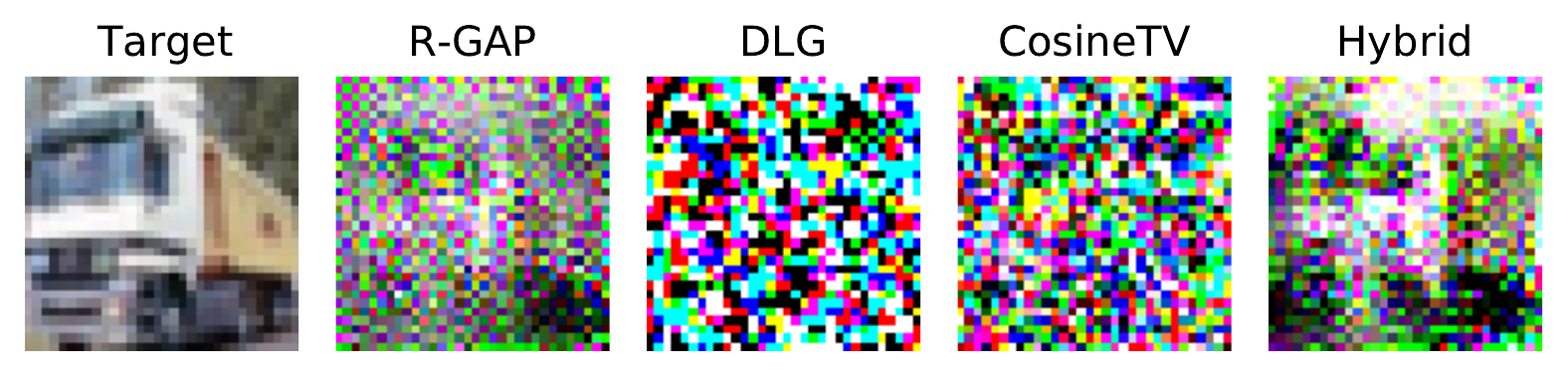}\\
			\includegraphics[width=\linewidth]{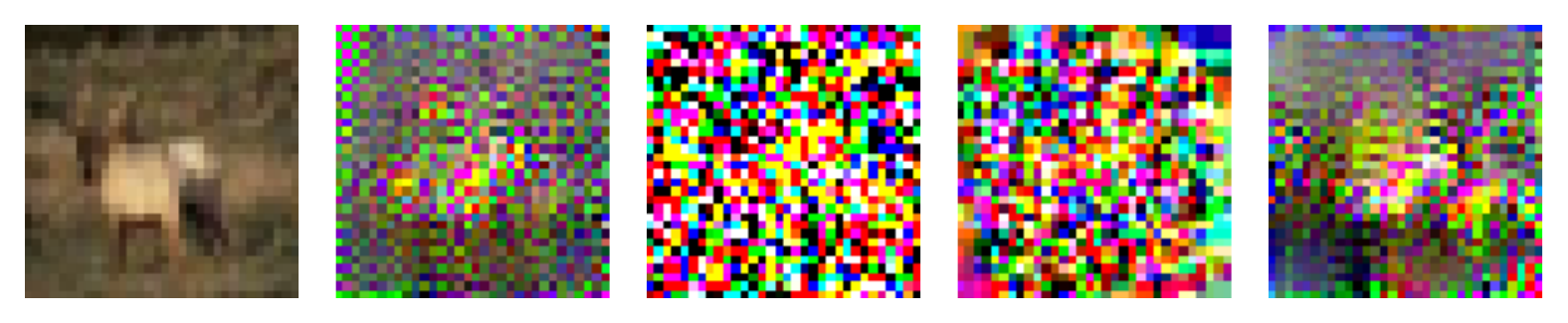}%
		\end{minipage}
		\end{subfigure}\hfill
		\begin{subfigure}[h]{0.49\textwidth}
		\begin{minipage}[c]{0.15\linewidth}
			\rotcaption{\parbox{1.70cm}{pre-trained Variant 2}}\label{fig:cnn4c2Pretrained}
		\end{minipage}\hfill%
		\begin{minipage}[c]{0.85\linewidth}
			\includegraphics[width=\linewidth]{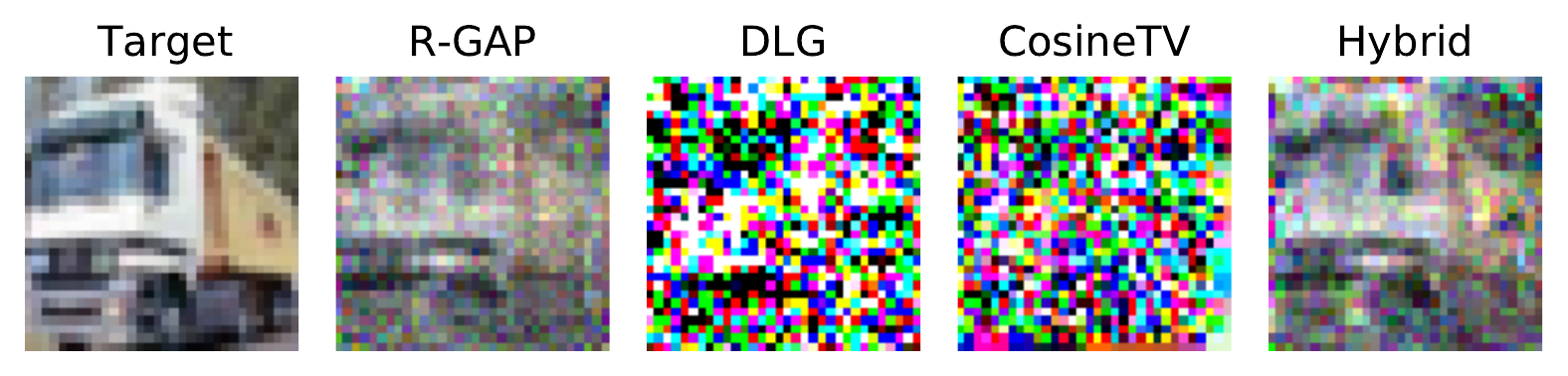}\\
			\includegraphics[width=\linewidth]{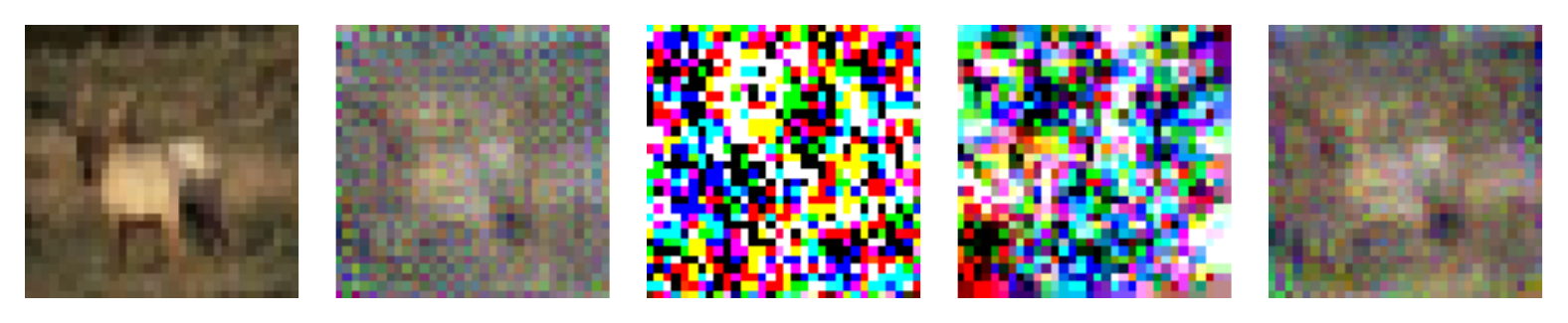}
		\end{minipage}
		\end{subfigure} \\
		\vspace{1mm}
		\caption{Comparisons of reconstructions among all approaches for pre-trained CNN4. Two examples are presented for each architecture. Compared to Figure \ref{fig:cnn4c1} and \ref{fig:cnn4c2}, we notice that reconstructions from all methods have worsened, and DLG and CosineTV are no longer producing visually recognisable results. On the other hand, R-GAP and our hybrid method are still showing more recognisable results.}
		\label{fig:cnn4Pretrained}
\end{figure*}

It is worth noticing that in CNN2 Variant 1 and CNN3 Variant 3, all of their convolutional layers have positive index given by the summand $\rank(\bm{u}^{(i)}) - n_i$ in \eqref{rankdefIndex}, which explains the most information leakage about the training image compared to other variants. On the other hand, we noticed that the value of $\rank(\bm{u}^{(i)}) - n_i$ is negative in the first layer in CNN3 Variant 4. Although it becomes positive in layer 2, it seems that this cannot make up for the loss of information occurred in layer 1. We provide the value of $\rank(\bm{u}^{(i)}) - n_i$ at each layer in all the models in Table \ref{tbl:modelArch}.

We also apply our algorithm against CNN4 that are pre-trained, see Figure \ref{fig:cnn4Pretrained}. The quality of reconstructions has degraded for all methods, although our method is still showing more recognisable results. Details of these experiments are provided in Section \ref{app:detailsPretraining}. We provide more reconstructed examples from CIFAR-10 in Section \ref{app:moreExamplesCIFAR10}. All results are consistent with our analysis.

In summary, we have shown that reconstructions from our hybrid method improve over those from R-GAP while minimising the instability that is inherent in DLG and CosineTV. 

\section{Conclusion}

In this paper, we try to further our understanding of existing gradient-leakage attacks by developing a hybrid framework which combines solving a linear system at each layer accompanied by gradient matching for corrections. Our framework provides a connected viewpoint between the existing analytic and optimisation-based methods. It also partially attributes the vulnerability of a deep network against gradient-leakage attacks to its architecture. The metric we propose can provide us with a guideline in designing a deep network more securely. 

\paragraph{Limitations and Future Work} There are a few important questions that we haven't addressed in this work, e.g. how does one apply our framework when the batch size is greater than one? We provide detailed discussions on limitations and future work in Section \ref{app:limit_and_future_work}. 
The code for this work will be posted on \url{https://github.com/CangxiongChen/training-data-leakage}.

\paragraph{Acknowledgements} We would like to acknowledge funding support from the EPSRC CAMERA Research Centre (EP/M023281/1 and EP/T022523/1), the Innovate UK SmartRoto project (04642027) and the Royal Society.

\bibliography{main_reference}

\newpage
\appendix
\onecolumn
\begin{center}
\section*{Analysing Training-Data Leakage from Gradients through Linear Systems and Gradient Matching: Supplementary Material}
\end{center}
\section{Proof of Lemma~\ref{lemma}}
\label{app:proof}
\begin{proof}
We will prove the general case first. By the construction of a fully-connected layer, we can take the $j$-th column of the gradient constraint \eqref{gradconstraint} which gives:
\begin{equation}
    (\frac{\partial \CL}{\partial w_{1j} ^{(i)}},...,\frac{\partial \CL}{\partial w_{nj} ^{(i)}})^T = x_j ^{(i)} (\frac{\partial \CL}{\partial z_1 ^{(i)}}, ... , \frac{\partial \CL}{\partial z_n ^{(i)}})^T.
\end{equation}
This implies that if $\frac{\partial \CL}{\partial z_k ^{(i)}} \neq 0$ for some $k, 1 \leq k \leq n$, then $\bm{x}^{(i)}$ can be uniquely determined:
\begin{equation}
        x_j ^{(i)}= \frac{\partial \CL}{\partial w_{kj} ^{(i)}} (\frac{\partial \CL}{\partial z_k ^{(i)}})^{-1}.
\end{equation}
In the special case when $b_k ^{(i)}\neq 0$, we can see from the weight constraint \eqref{weightconstraint} that:
\begin{equation}
    \frac{\partial \CL}{\partial z_k ^{(i)}} = \frac{\partial \CL}{\partial b_k ^{(i)}},
\end{equation}
which was observed in \cite{aono2017privacy} and subsequently also in \cite{Fan2020}. In general, since the activation functions are assumed to be piecewise invertible and piecewise differentiable, and since $\frac{\partial \CL}{\partial x_k ^{(i+1)}}$ is assumed to be nonzero, we can compute $x_j ^{(i)}$ using \eqref{gradrelations}, which gives \eqref{fullyConnectedGeneral}.
\end{proof}

\section{Convergence of the algorithm}\label{app:convergence_algo}
At a convolutional layer, we have formulated the optimisation problem where the linear system \eqref{WeightAndGradConstraintCombined} defines a hard constraint, so that it is satisfied throughout the optimisation. Although this formulation makes it clear that the difference $\overline{\bm{x}^{(i)}} - \bm{x}^{(i)} _{LS}$ is inside the null space of $\bm{u}^{(i)}$, we find that in practice it does not always lead to the convergence of the optimisation within a reasonable amount of run time when we use trust-region methods such as \cite{nocedal2006numerical} and \cite{lalee1998implementation} to solve the constrained optimisation problem given in \eqref{LayerwiseOptProb2}. On the other hand, allowing constraint violation by defining \eqref{WeightAndGradConstraintCombined} as a soft constraint can often speed up the convergence. More precisely, instead of the problem \eqref{LayerwiseOptProb2}, we consider an unconstrained optimisation problem:

\begingroup
\small
\begin{equation}\label{layerwiseOptSoftVersion}
        \argmin_{\bm{x}} \Big \{ \mu_1 \mathcal{D} \left [ \nabla_{\bm{w}}\CL^{(i)}(\bm{x};\bm{w})\vert_{\bm{w} = \bm{w}^*},\nabla_{\bm{w}}\CL^{(i)}(\bm{x}_{true}; \bm{w})\vert_{\bm{w} = \bm{w}^*} \right ] 
         + \mu_2 \text{TV}(\bm{x}) 
         + \mu_3 ||\bm{u}^{(i)} \bm{x} - \bm{v}^{(i)}||^2 \Big \},
\end{equation}
\endgroup

where $\mu_1, \mu_2, \mu_3 \in \BR$ are some given weights. We have observed in our experiments that with using the unconstrained optimisation, Algorithm \ref{hybridv1} will be able to converge much faster whereas it can take much longer for the original version to converge to the same result.

\section{Justification for the security measure}\label{append:justification_for_metric}
In light of the hybrid framework given for a convolutional layer, the problem of reconstructing a training image can be viewed as consisting of two parts:
\begin{enumerate}
    \item An iterative procedure starting from the output of the network.
    \item At each layer, we first solve a linear system defined by the forward and backward pass of the target image-label pair, then we correct the solution by gradient matching with the target image if the layer is convolutional. If the layer is fully connected, the correction is not necessary. 
\end{enumerate}

For a fully connected layer, we have shown in Lemma \ref{lemma} that we can always reconstruct the input in full. This can be regarded as no level of security and we omit it from our definition of the metric. For a convolutional layer, since the basic criterion for measuring the solubility of a linear system is given by comparing the rank of the coefficient matrix with the number of unknowns, and that the corrected solution still satisfies the linear system, we consider $\rank(\bm{u}) - |\bm{x}|$ as an index to measure the efficacy of the hybrid method. The larger this number is, the less rank-deficient the linear system \eqref{WeightAndGradConstraintCombined} is and so more likely to have a full reconstruction for this layer. We also notice that the position where the rank-deficiency happens also matters. The closer it is to the first layer, the bigger impact it has on the reconstruction. This is consistent with our intuition that if the representation of the input data loses information at the first layer, it will be unlikely to substitute that loss in latter layers. To accommodate for this effect, we discount the index $\rank(\bm{u}) - |\bm{x}|$ by the position of the layer in the network.

\section{Details of the implementation}\label{FurtherDetailsExperiment}
For the re-implementations of DLG and CosineTV, we follow the number of iterations used by the authors of the corresponding work, i.e. 300 for DLG and 4800 for CosineTV. In the implementation for our hybrid method, we adopt the following setting of hyperparameters:
\begin{table}[h]
		\centering
		\caption{Hyperparameters for our implementation of the hybrid method. The weights are placed according to the objective function given in  \eqref{layerwiseOptSoftVersion}.}\label{tbl:hyperparaHybrid}
		\vspace{4pt}
		\begin{tabular}{lccc}
		\toprule
		     & layer 1 & layer 2 & other layers  \\
		     \midrule
		     Iterations & 10000 & 8000 & 1000  \\ 
		     $\mu_1$ & 1.0 &  1.0 & 10.0 \\
		     $\mu_2$  & 1.0 & 1.0 & 0.1  \\
		     $\mu_3$   & 0.05 & 0.1 & 1.0 \\
		     \bottomrule
		\end{tabular}%
\end{table}
\newpage

\section{Evaluations of the experiments}
\begin{table*}[!ht]
		\centering
		\caption{MSE and PSNR (as in the first and second component in the pair) of the reconstructions in each variant, averaged over the two images; they are consistent with the visual qualities in Figures \ref{fig:cnn2andcnn4}, \ref{fig:cnn3} and \ref{fig:cnn4Pretrained}.} \label{app:tbl:scores}
		
		\vspace{4pt}
		\scalebox{0.7}{%
		\begin{tabular}{lcccc}
		\toprule
		     & R-GAP & DLG & CosineTV & Hybrid  \\
		     \midrule
		     CNN2 Variant 1 & 0.0000, 197.00     & 2.0181, 45.08  & 0.2290, 54.54 & 0.0008, 79.82 \\ 
		     CNN2 Variant 2 & 0.0346, 62.75 & 2.15E+08, -9.32       & 0.3257, 53.01  & 0.0051, 71.69 \\
		     CNN3 Variant 1 & 0.0531, 60.90 & 0.9279, 48.46 & 0.4086, 52.03 & 0.0478, 61.48 \\
		     CNN3 Variant 2 & 0.0518, 60.99 & 0.9900, 48.17 & 0.4739, 51.37 & 0.0322, 63.78 \\
		     CNN3 Variant 3 & 0.0000, 181.26 & 3.75E+17, -59.49 & 0.2302, 54.51 & 0.0020, 75.79  \\
		     CNN3 Variant 4 & 0.0429, 61.83 & 5.11E+13, -29.17 & 0.5082, 51.07 & 0.0417, 61.96 \\
		     CNN4 Variant 1 & 0.0547, 60.81 & 0.8585, 48.79 & 0.4255, 51.86 & 0.0610, 60.28 \\
		     CNN4 Variant 2 & 0.0406, 62.05 & 0.0951, 58.35 & 0.2177, 54.82 & 0.0139, 67.72 \\
		     CNN4 Variant 1 (pre-trained) & 0.2174, 54.90 & 7.58E+08, -3.07 & 0.8048, 49.25 & 0.3449, 53.33 \\
		     CNN4 Variant 2 (pre-trained) & 0.0341, 62.81 & 406.2, 26.98 & 0.7353, 49.71 & 0.0288, 63.63 \\
		     \bottomrule
		\end{tabular}%
		}
\end{table*}

\section{Details on the pre-training of CNN4}\label{app:detailsPretraining}
\begin{figure}[h]
		\centering
		%
		%
		\begin{subfigure}[h]{0.30\textwidth}
			\includegraphics[width=\linewidth]{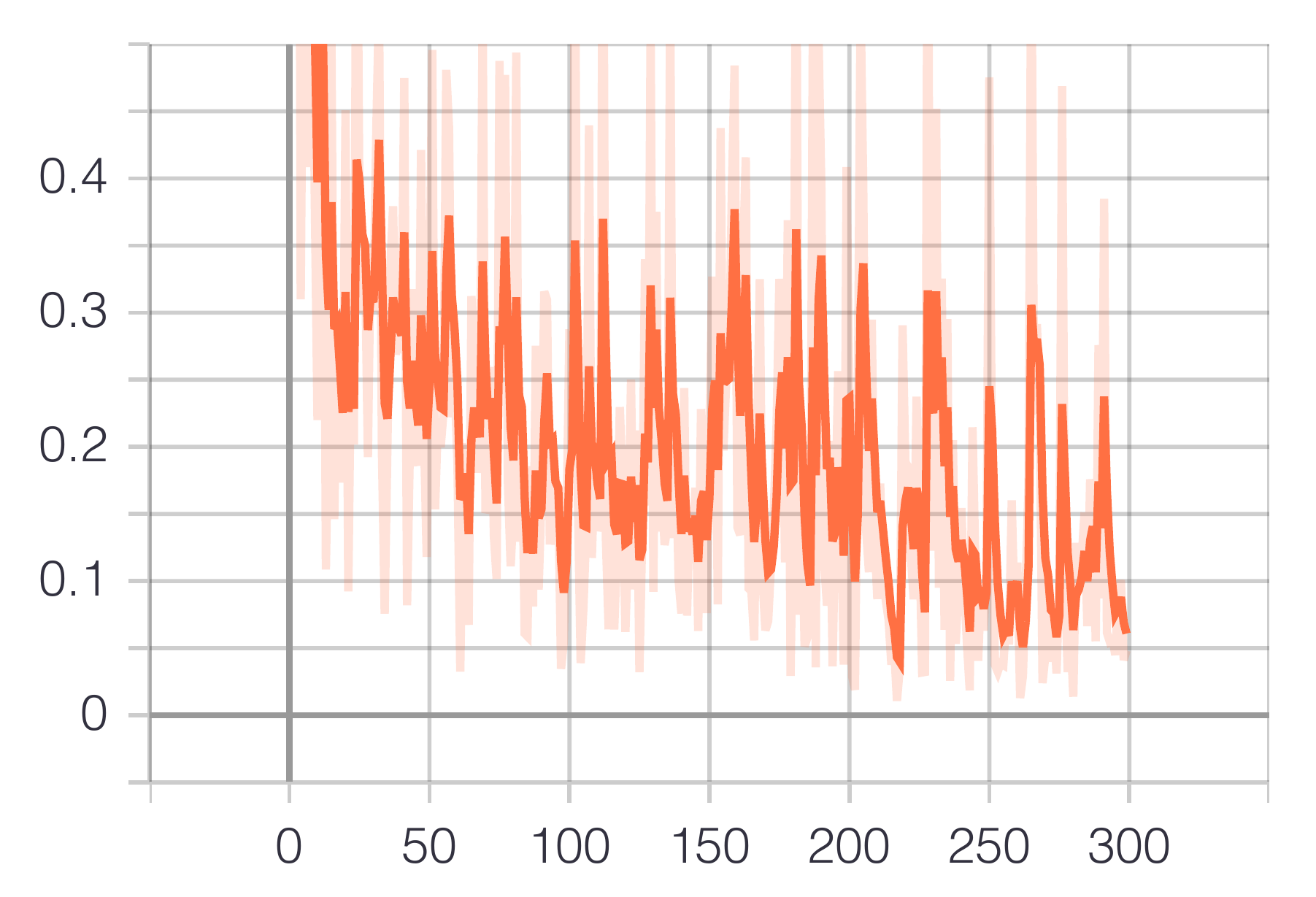}\\
			\includegraphics[width=\linewidth]{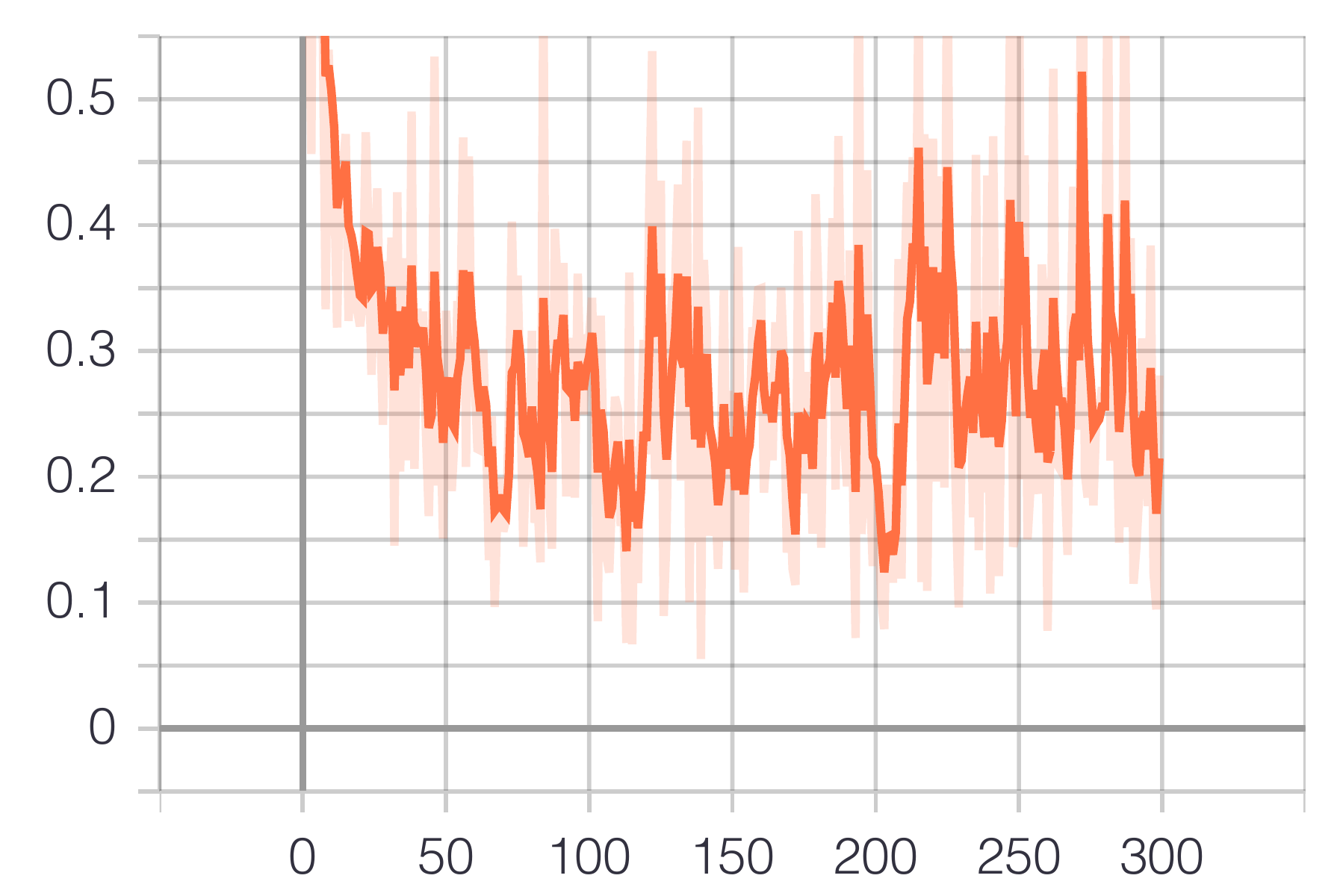}\caption{CNN4 Variant 1. Top: training loss; Bottom: testing loss}
		\end{subfigure}\hspace{1cm}
		\begin{subfigure}[h]{0.30\textwidth}
			\includegraphics[width=\linewidth]{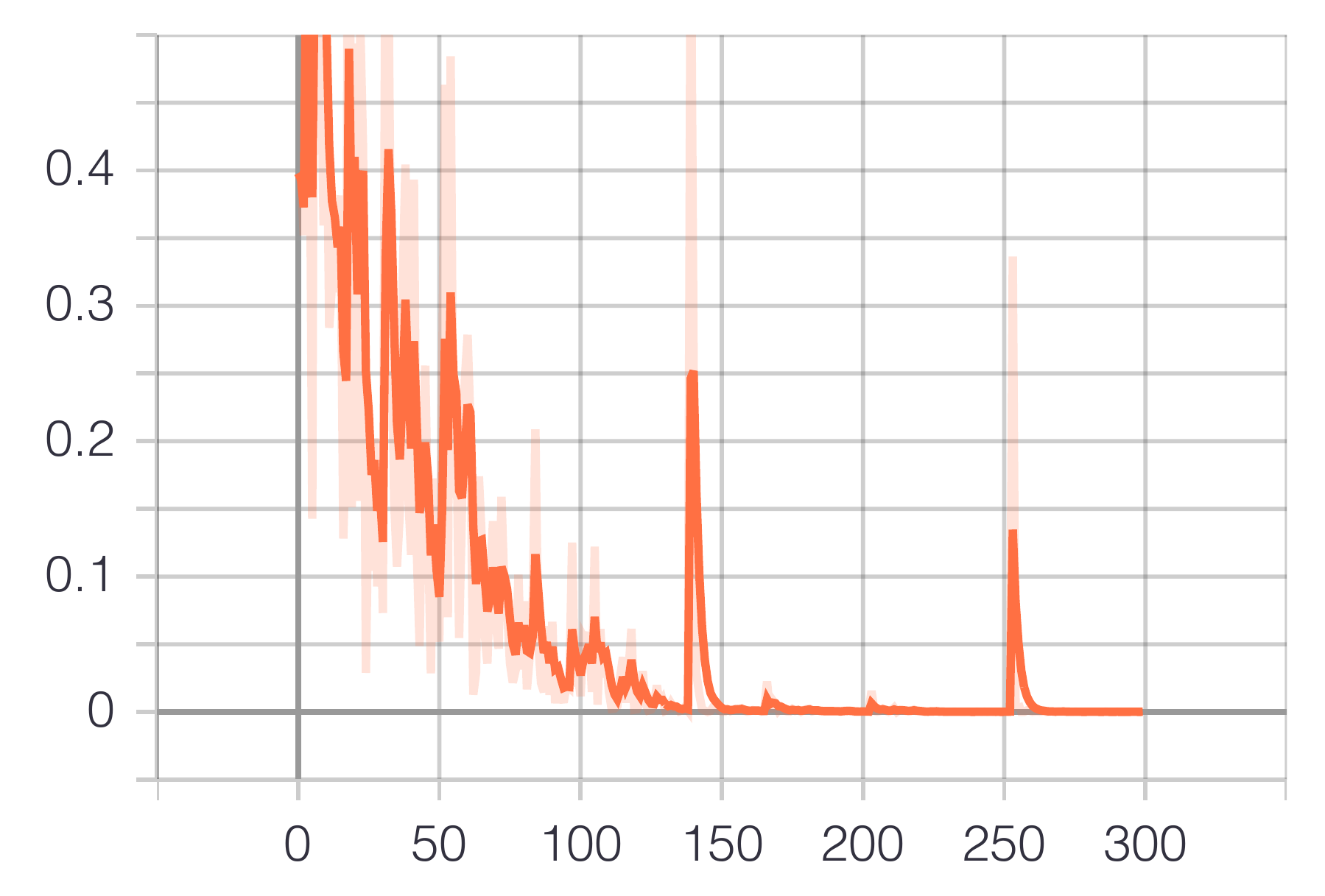}\\
			\includegraphics[width=\linewidth]{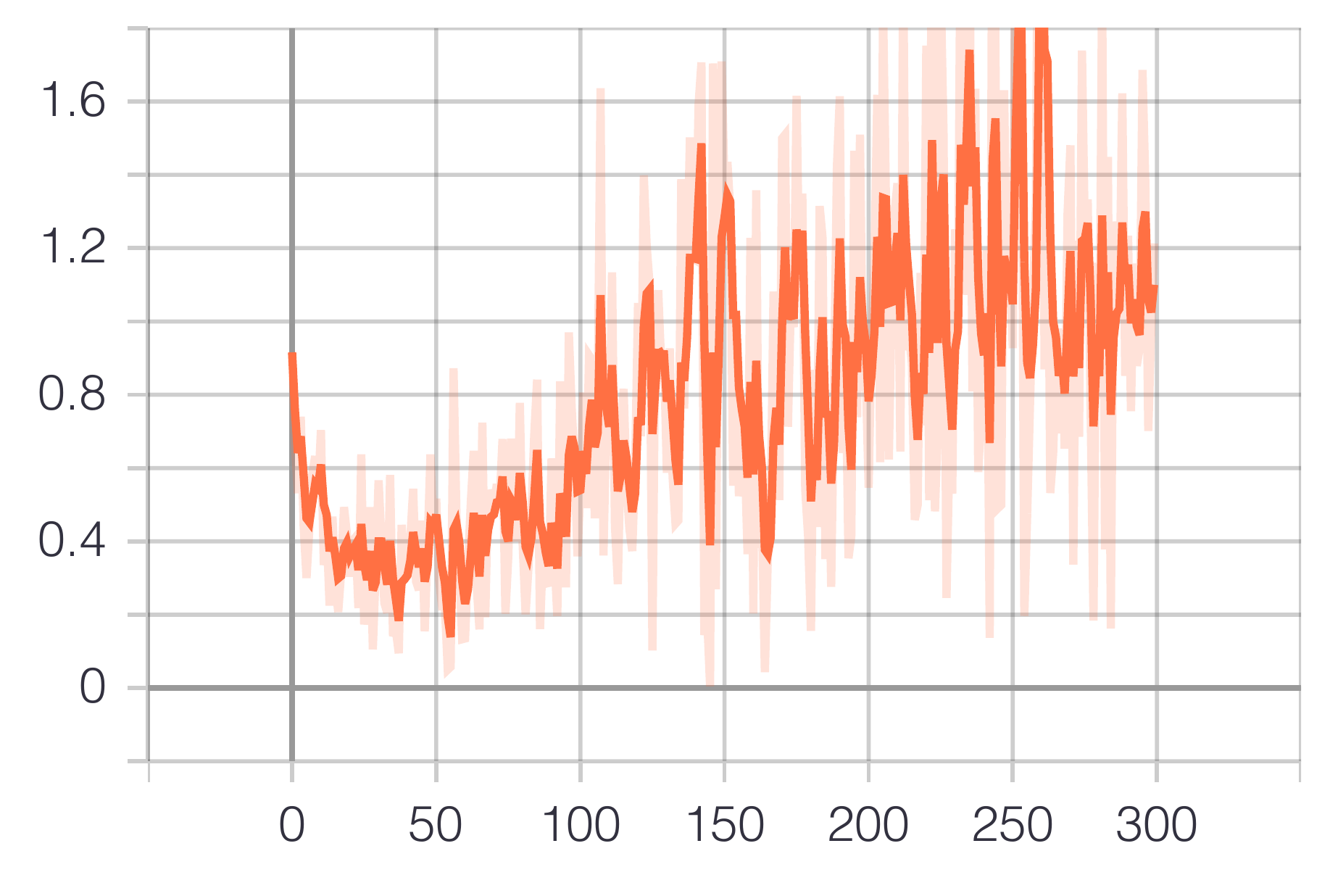}
			\caption{CNN4 Variant 2. Top: training loss; Bottom: testing loss}
		\end{subfigure} \\
		\vspace{1mm}
		\caption{Plots of the losses during training and testing. Notice that variant 2 has noticeable overfitting.}
		\label{fig:trainingTestingLossCNN4}
\end{figure}

We pre-train CNN4 on images from only two classes from CIFAR-10, namely `automobiles' and `birds'. The target images used for reconstructions have not been seen by the models during pre-training. Both variants of CNN4 have been trained on 10000 images and tested on 1000 images, with batch size 64. We have used ADAM optimiser with initial learning rate of 0.001 and the models were trained for 300 epochs. Figure \ref{fig:trainingTestingLossCNN4} shows losses during training and testing.
 We expect that reconstructions from all methods to deteriorate, because a pre-trained model is likely to produce gradients with smaller magnitude and variance when it is retrained on an unseen image compared to an untrained model. Results shown in Figure \ref{fig:cnn4Pretrained} seems to have confirmed this guess. However, there might be exceptions when the model is overfitted and then retrained on an unseen image. We notice that across untrained and trained cases and for both images,  the metric $c(\CM)$ and the layerwise rank deficiency $\rank(\bm{u}^{(i)}) - n_i$ have the same value in each variant respectively.

\section{Limitations and future work}\label{app:limit_and_future_work}

\paragraph{Batch size}
In this work, we are only considering the problem to reconstruct one single training image. It is natural to wonder how to apply our framework to the case when we are given the gradients from a batch of training images. When a batch of images are used, the gradients used to define the gradient constraint \eqref{gradconstraint} will be an average of those from each image in the batch. It is unclear how to decompose the gradient constraint without introducing further assumptions. Without those assumptions, a straightforward application of our hybrid framework in this case will give a reconstruction that is difficult to interpret. Although there exist works that try to tackle this problem (for example \cite{YinSeeThroughGradients} and  \cite{geiping2020inverting}), we have not found any approach that offers theoretical insight nor guarantees to this problem. For the interest of obtaining theoretical guarantees of the reconstructions and the security of the architecture, we would leave this as future work.

\paragraph{Scope of the security measure}
We think that the security measure \eqref{rankdefIndex} captures the solubility of the linear system \eqref{WeightAndGradConstraintCombined} by computing its rank deficiency which depends on the input and output dimensions of the layer and the values of weights and gradients. The security measure does not take into account other factors such as the condition number of the the system \eqref{WeightAndGradConstraintCombined}, which although affects the stability of the solution rather than solubility, can also affect the quality of the reconstructions. It would be interesting to extend the security measure to include the condition number of the system to give a more accurate measure of the security of the architecture under our hybrid framework.

\paragraph{Activation functions}
In all convolutional networks used in the experiment, we have assumed the activation function in a convolutional layer to be Tanh. We believe that our framework will also apply to other activations as long as they are smooth and invertible. We noticed that if we use LeakyReLU and solve the optimisation problem in \eqref{LayerwiseOptProb2} using constrained optimisation such as trust-region methods from \cite{nocedal2006numerical} and \cite{lalee1998implementation}, it will be difficult for it to converge to the correct optimum within a reasonable amount of running time. Although this is more of a limitation with the optimisation than our framework, we will be looking for strategies for optimisation that can better deal with non-smooth functions in the future.

\paragraph{Architectures of the target model}
 Although we have not considered other popular architectures in image classifications such as Residual networks \cite{he2015deep}, we believe our framework can be adapted to these networks if we can define the corresponding linear system for a residual block. For example, we can define a similar linear system as \eqref{weightconstraint} by approximating the ResNet block defined in \cite{he2015deep} using Taylor expansion for the non-linear terms given by activations inside the block. More details will be given in future work. 

\paragraph{Reducing gradient leakage}
Another future avenue of work is investigating strategies to reduce gradient leakage with theoretical guarantee. A promising direction is considering training methods that provides Differential Privacy. One insight from our analysis leading to the metric $c(\CM)$ is that to reduce gradient leakage, we can add noise to the gradients so that the value of $c(\CM)$ can be reduced sufficiently. 
\newpage

\section{More examples of reconstructions}\label{app:moreExamplesCIFAR10}
We provide more examples from CIFAR-10 for comparing all the methods discussed in the Experiment section. One image from each of the 10 classes is chosen. 
\begin{figure}[h]
		\centering
		%
		%
		\begin{subfigure}[h]{0.45\textwidth}
			\includegraphics[width=\linewidth]{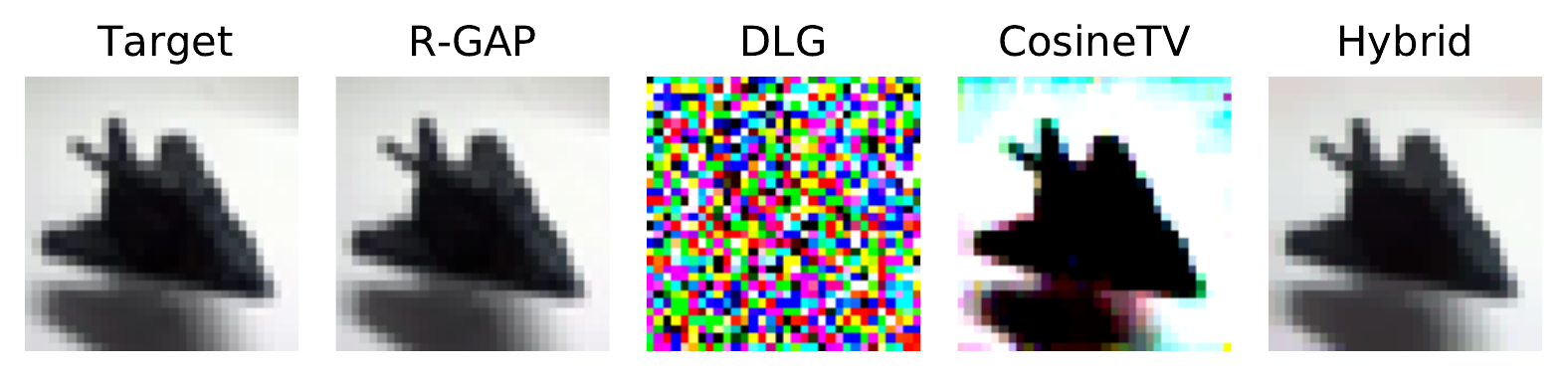}\\
			\includegraphics[width=\linewidth]{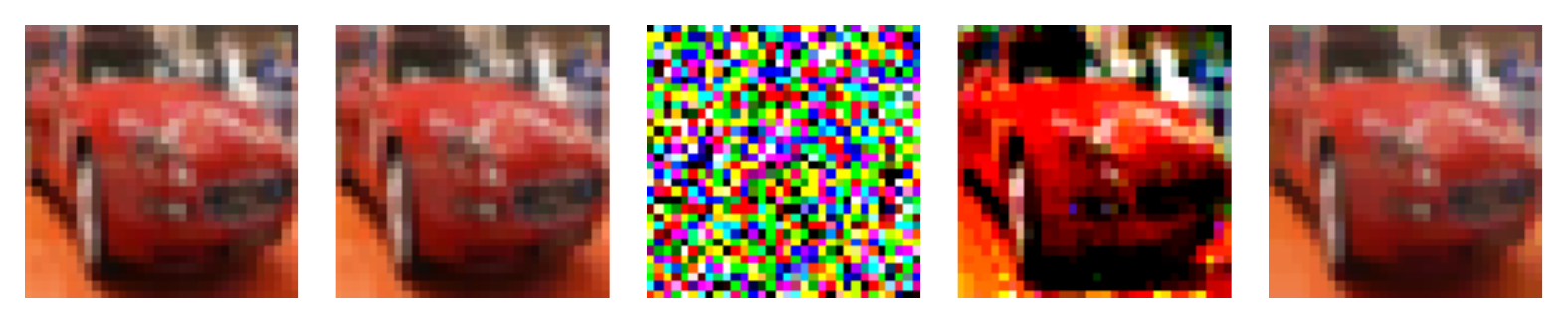}\\
			\includegraphics[width=\linewidth]{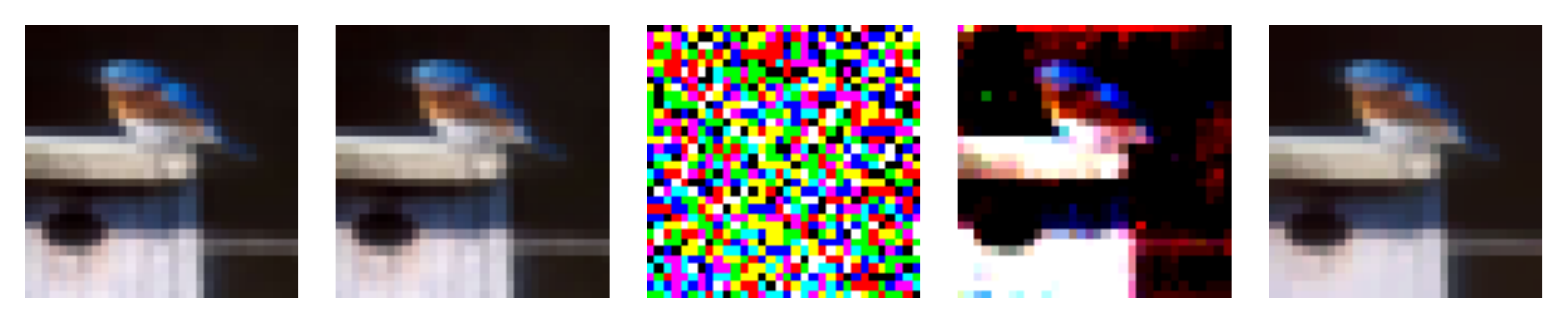}\\
			\includegraphics[width=\linewidth]{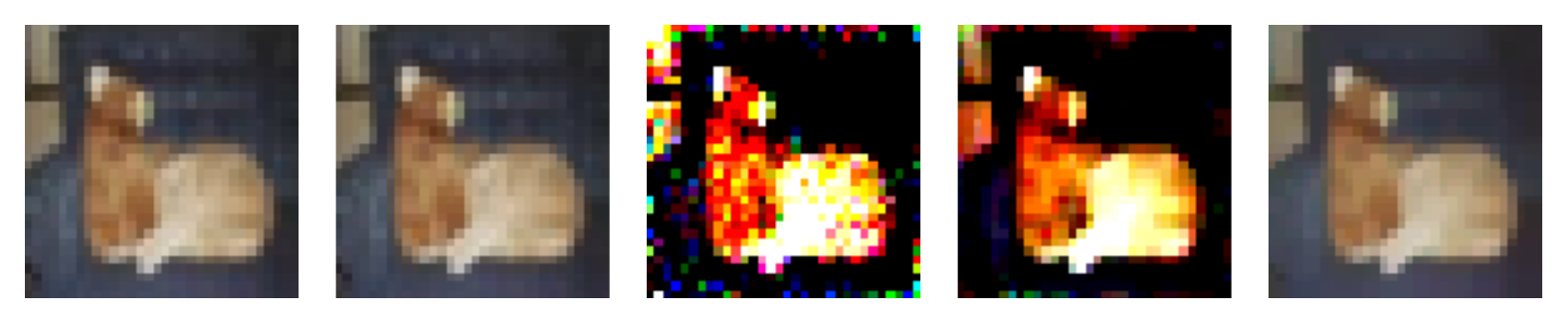}\\
			\includegraphics[width=\linewidth]{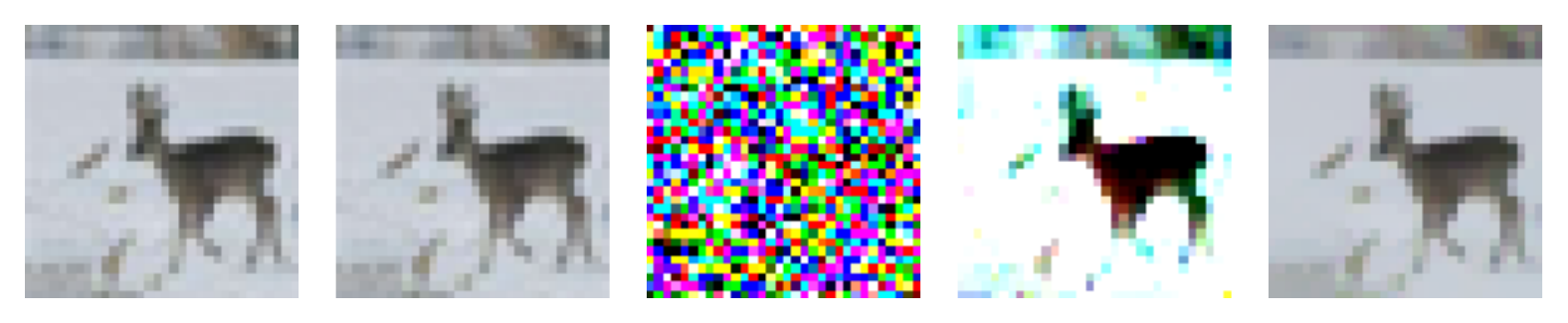}\\
			\includegraphics[width=\linewidth]{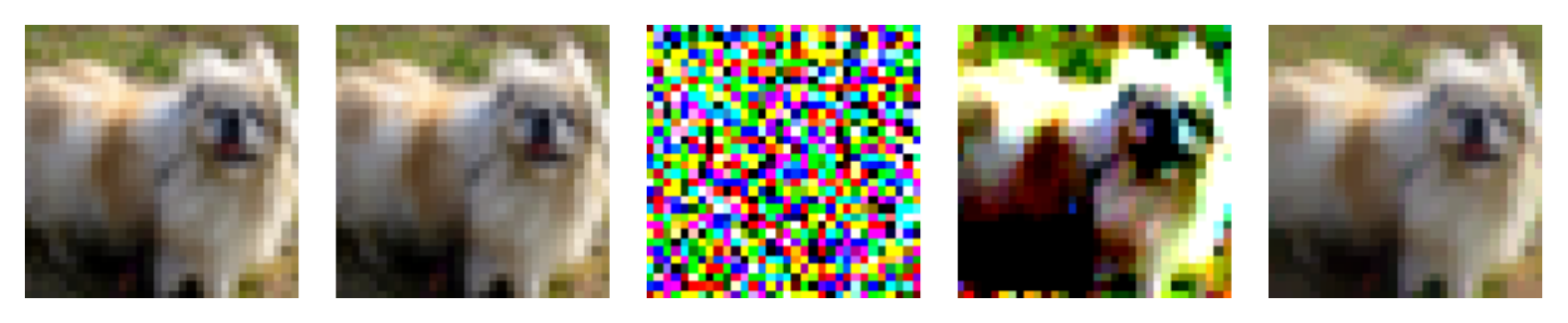}\\
			\includegraphics[width=\linewidth]{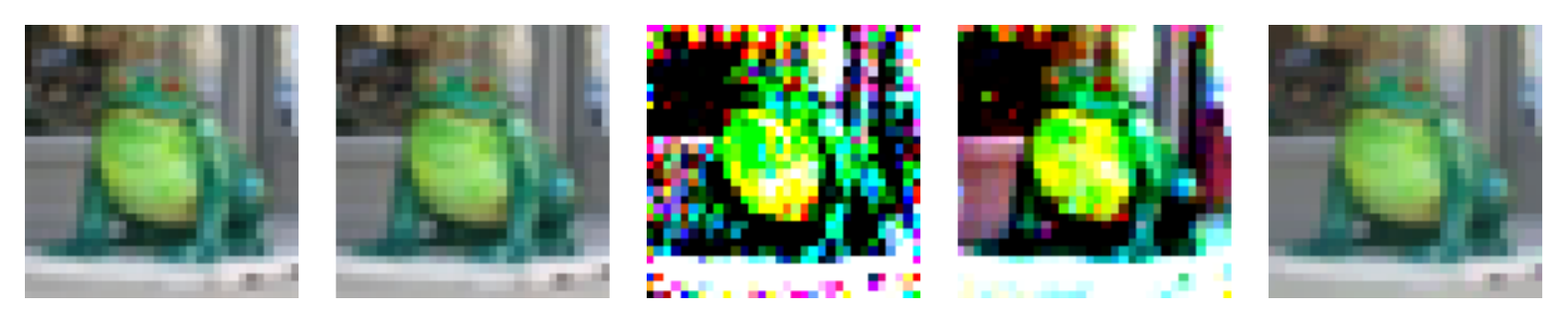}\\
			\includegraphics[width=\linewidth]{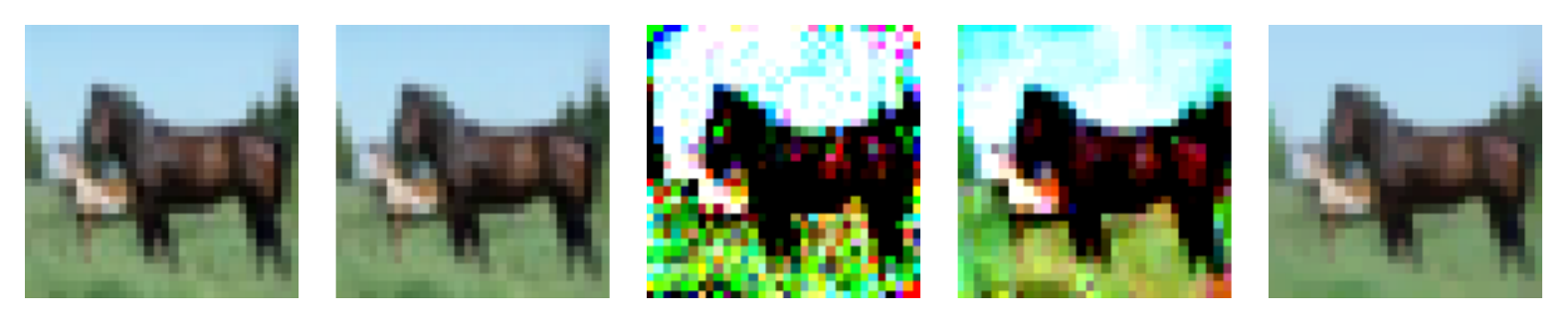}\\
			\includegraphics[width=\linewidth]{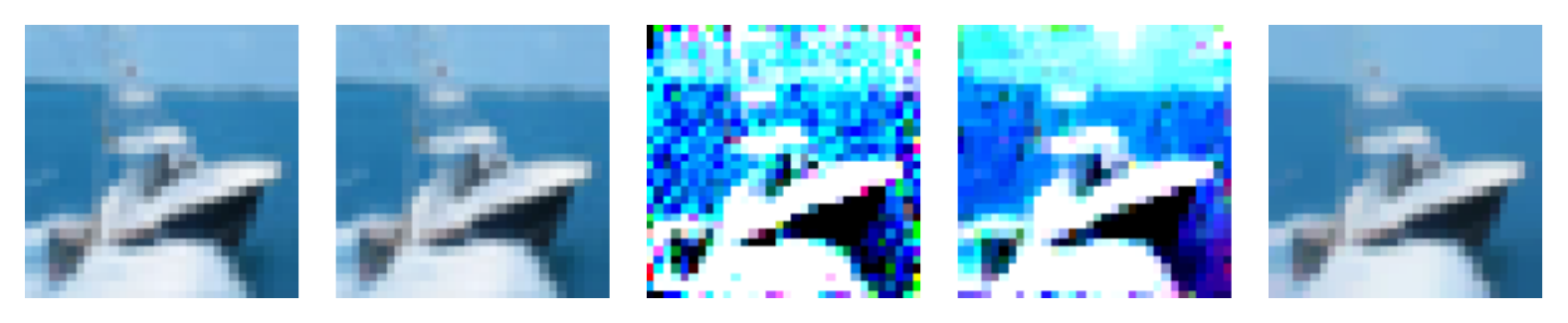}\\
			\includegraphics[width=\linewidth]{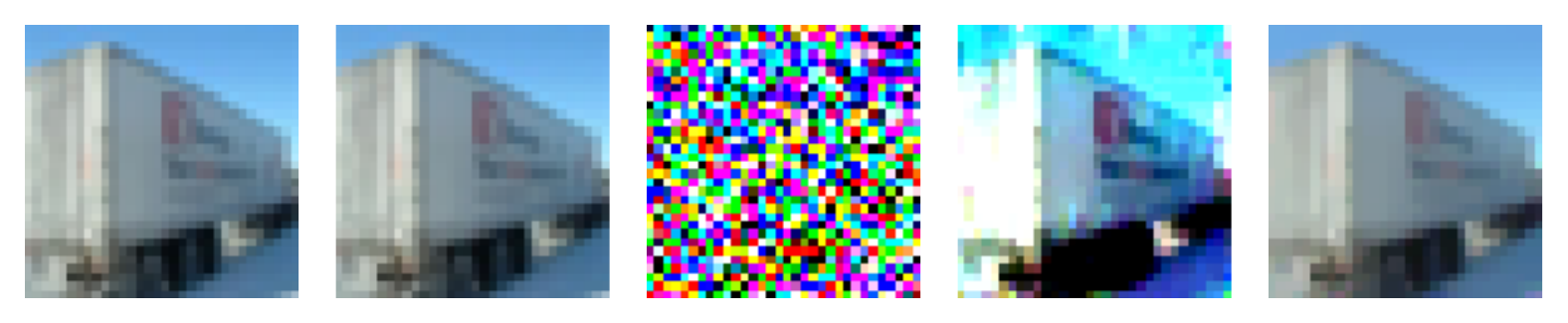}\\
			\caption{CNN2 Variant 1}
		\end{subfigure}\hspace{1cm}
		\begin{subfigure}[h]{0.45\textwidth}
			\includegraphics[width=\linewidth]{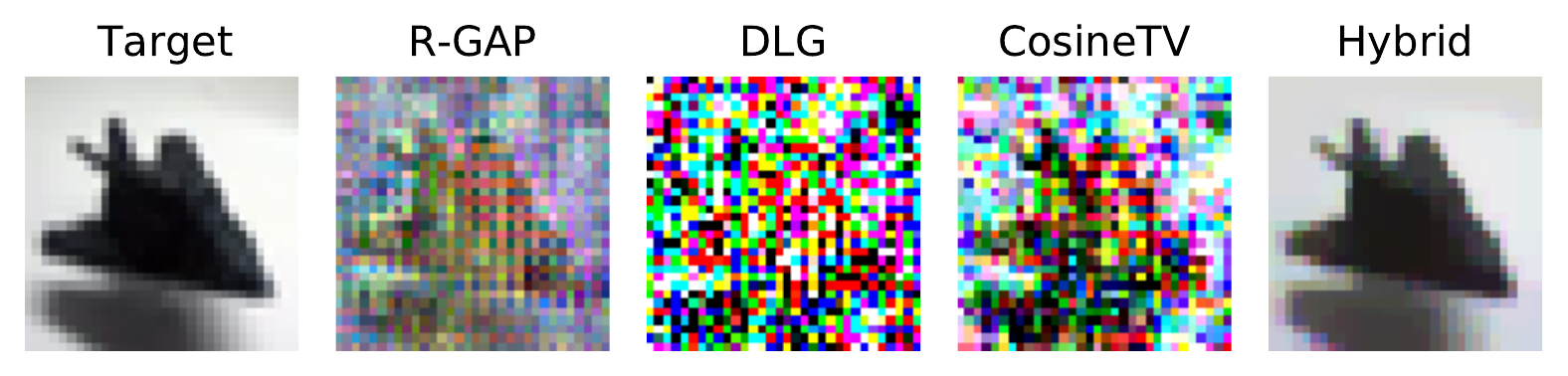}\\
			\includegraphics[width=\linewidth]{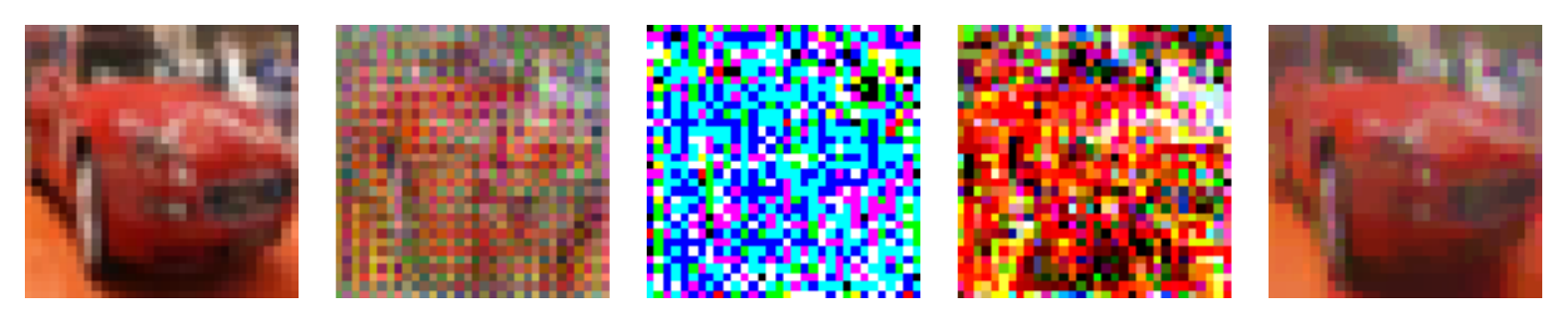}\\
			\includegraphics[width=\linewidth]{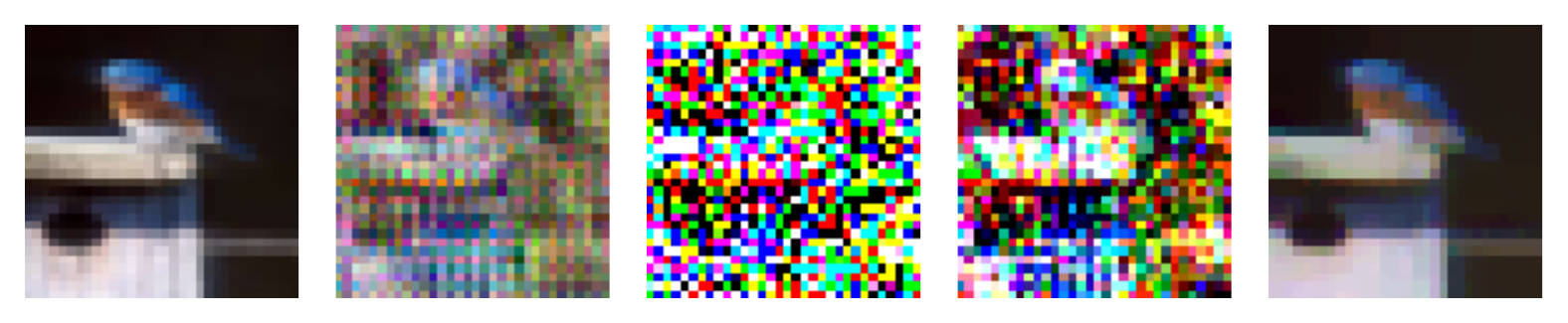}\\
			\includegraphics[width=\linewidth]{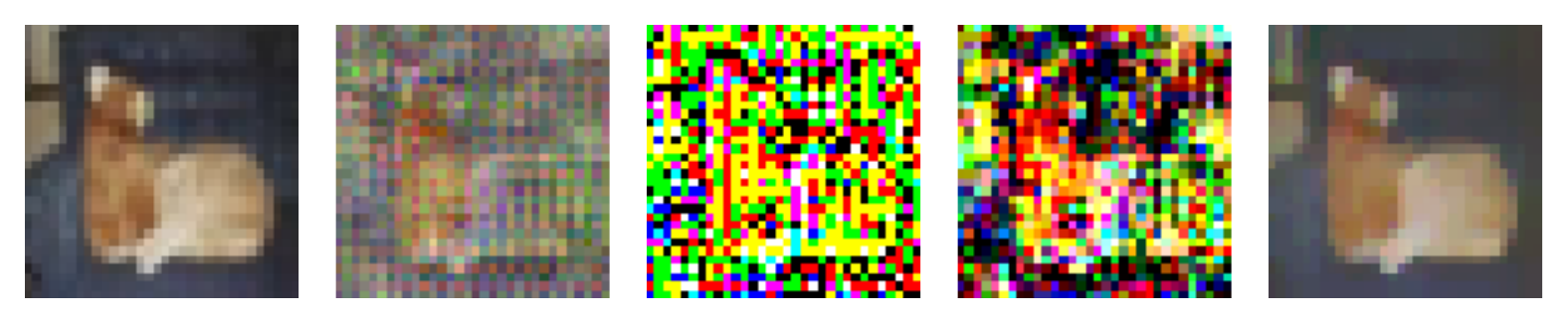}\\
			\includegraphics[width=\linewidth]{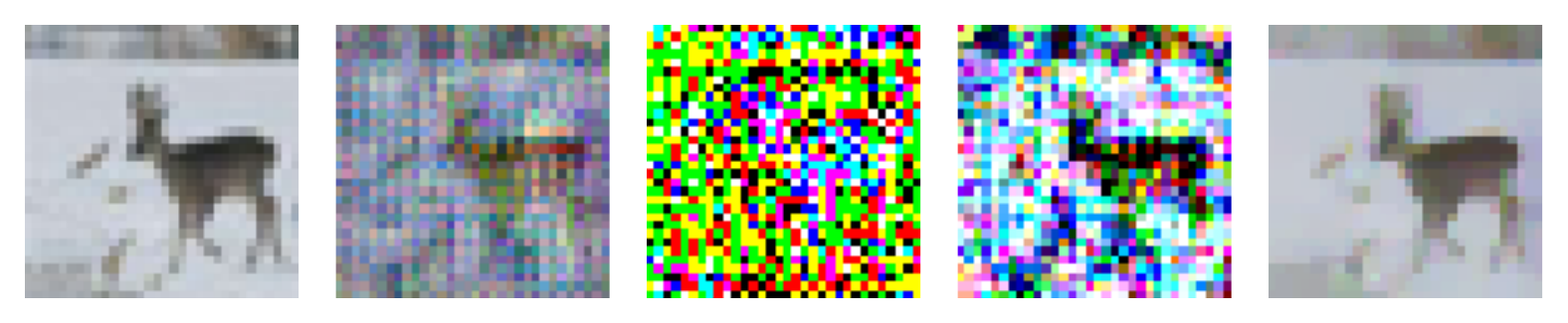}\\
			\includegraphics[width=\linewidth]{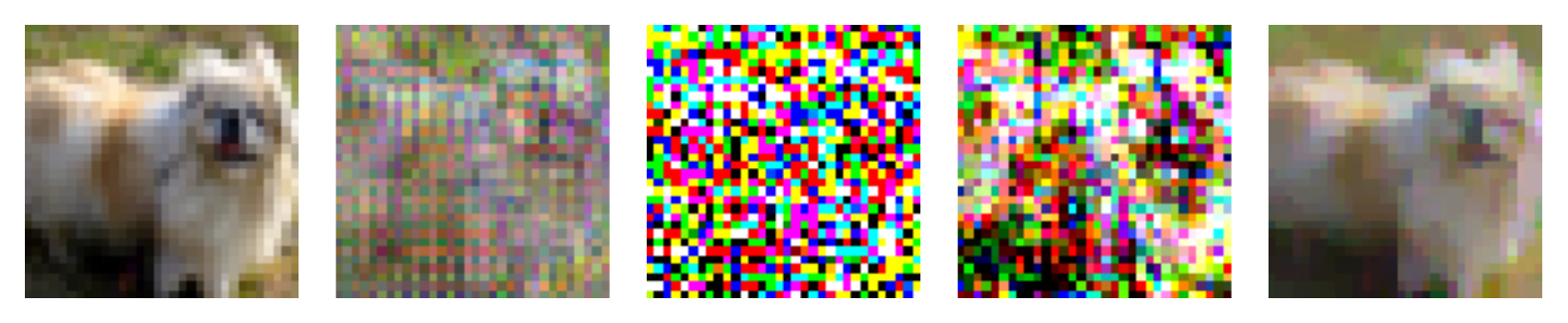}\\
			\includegraphics[width=\linewidth]{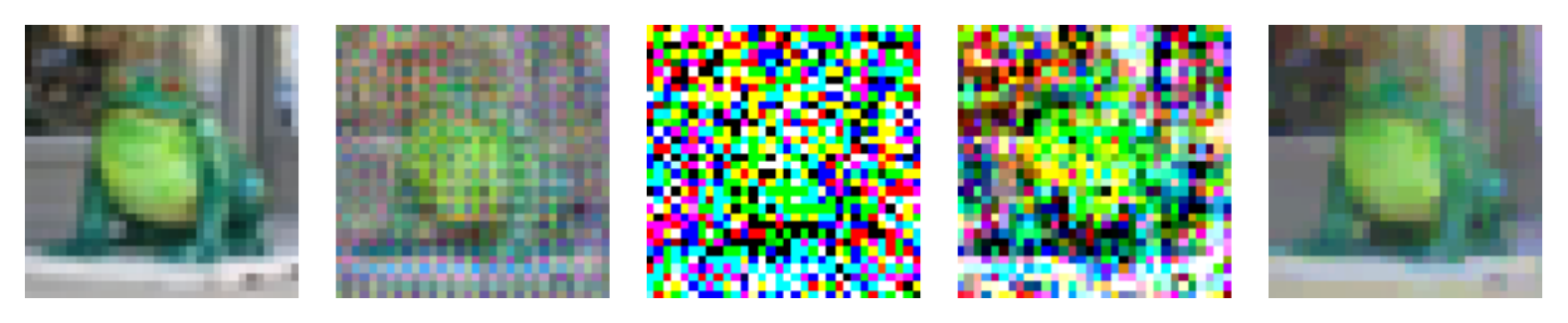}\\
			\includegraphics[width=\linewidth]{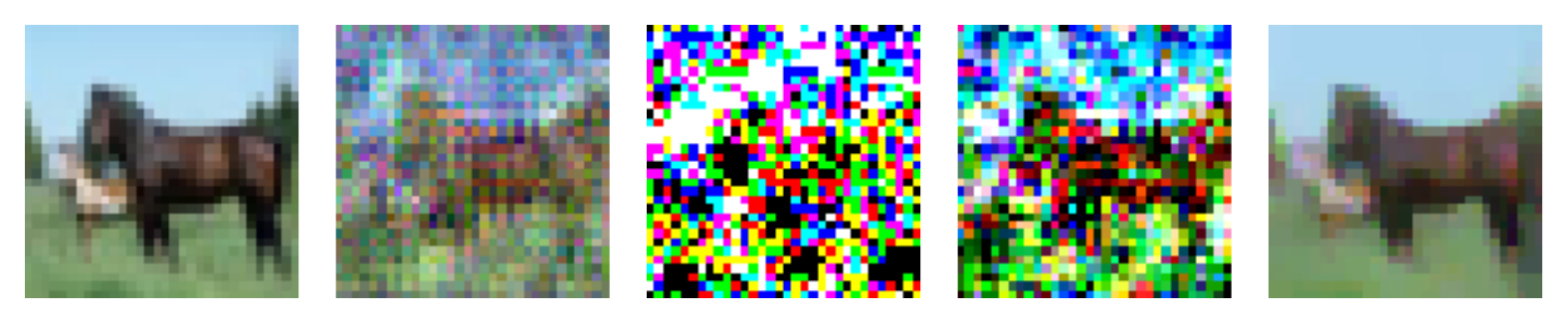}\\
			\includegraphics[width=\linewidth]{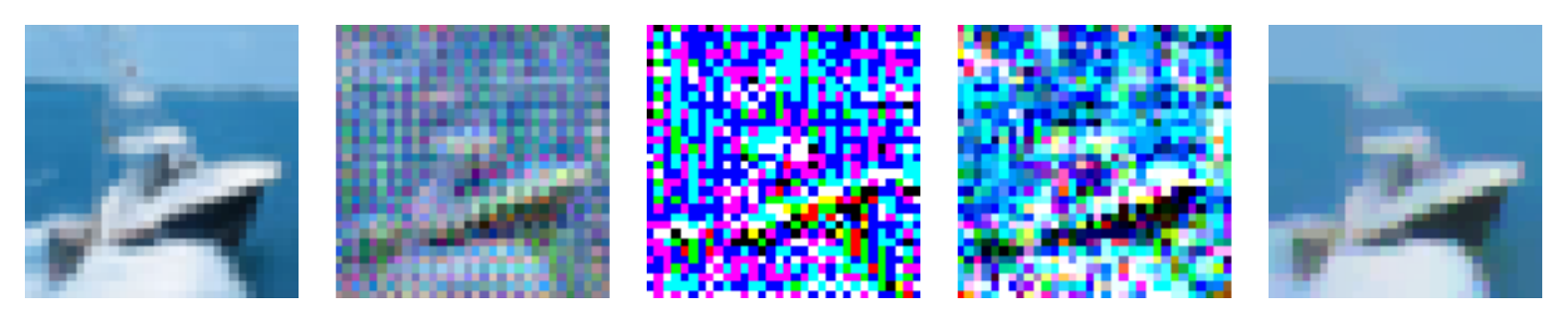}\\
			\includegraphics[width=\linewidth]{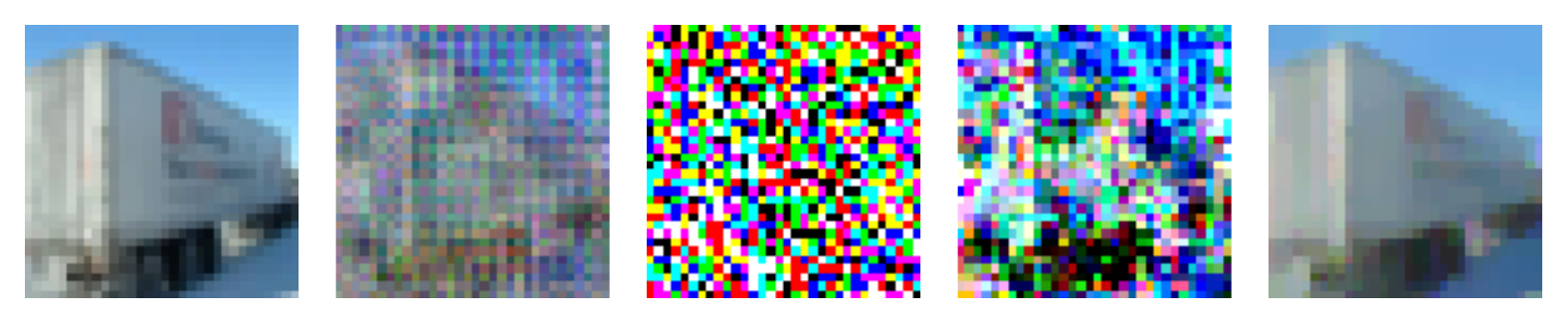}\\
			\caption{CNN2 Variant 2}
		\end{subfigure} \\
		\label{fig:multiimagecnn2v12}
\end{figure}
\begin{figure}[h]
		\centering
		%
		%
		\begin{subfigure}[h]{0.45\textwidth}
			\includegraphics[width=\linewidth]{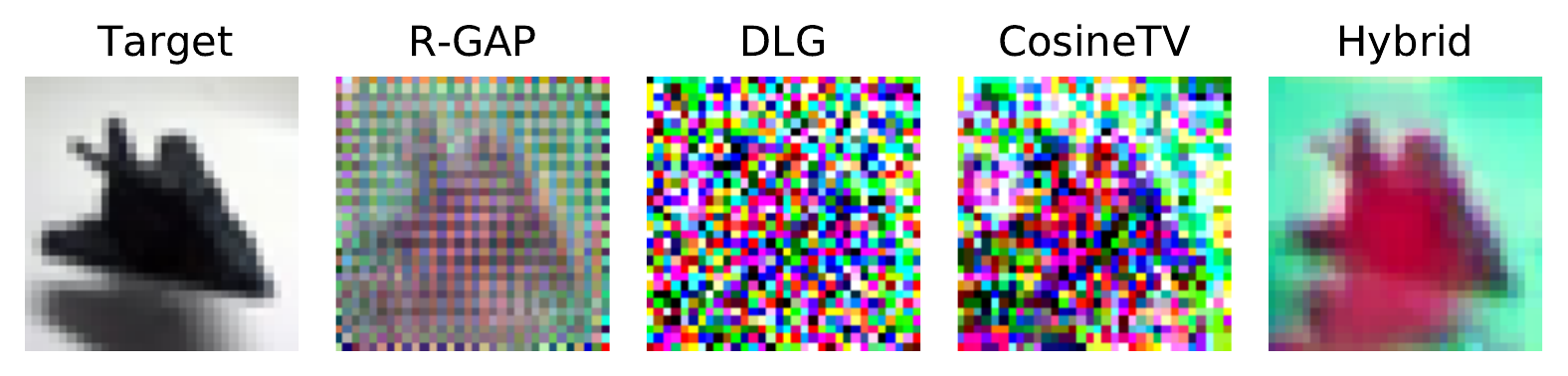}\\
			\includegraphics[width=\linewidth]{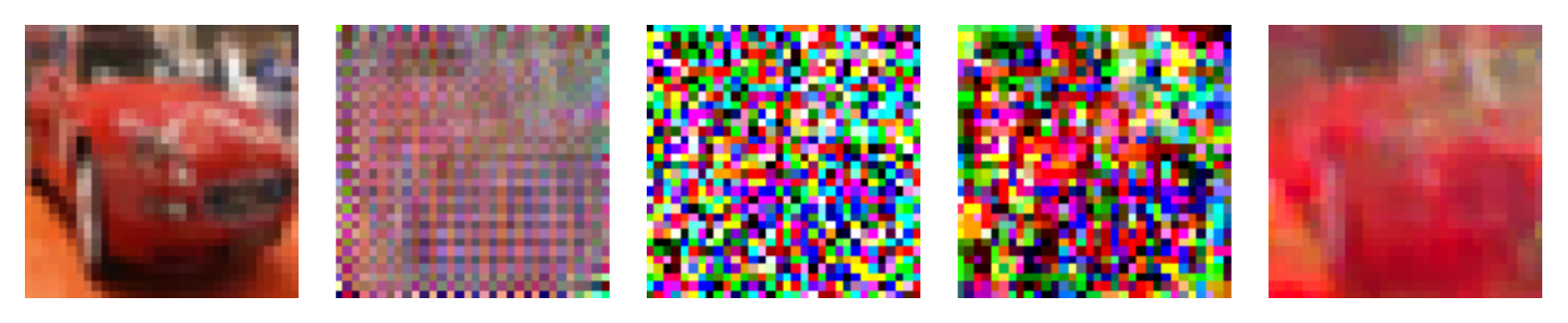}\\
			\includegraphics[width=\linewidth]{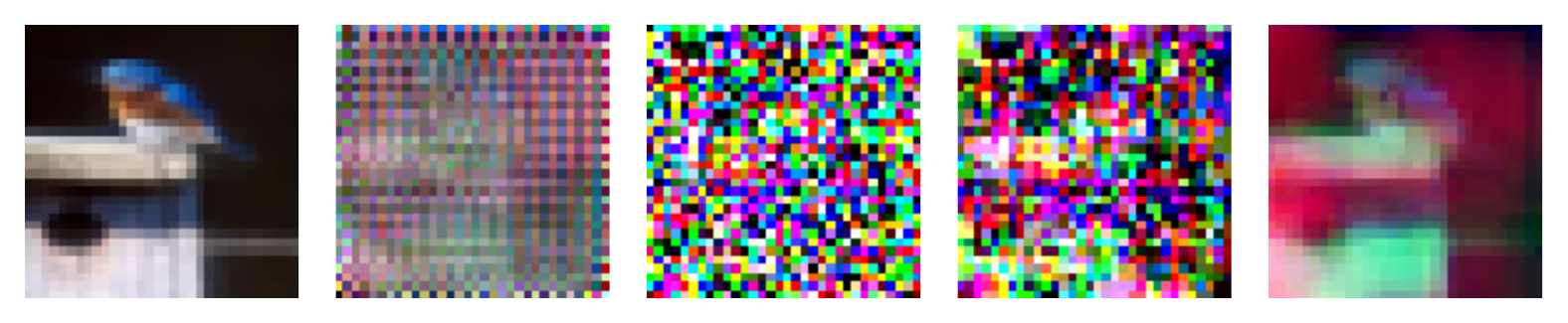}\\
			\includegraphics[width=\linewidth]{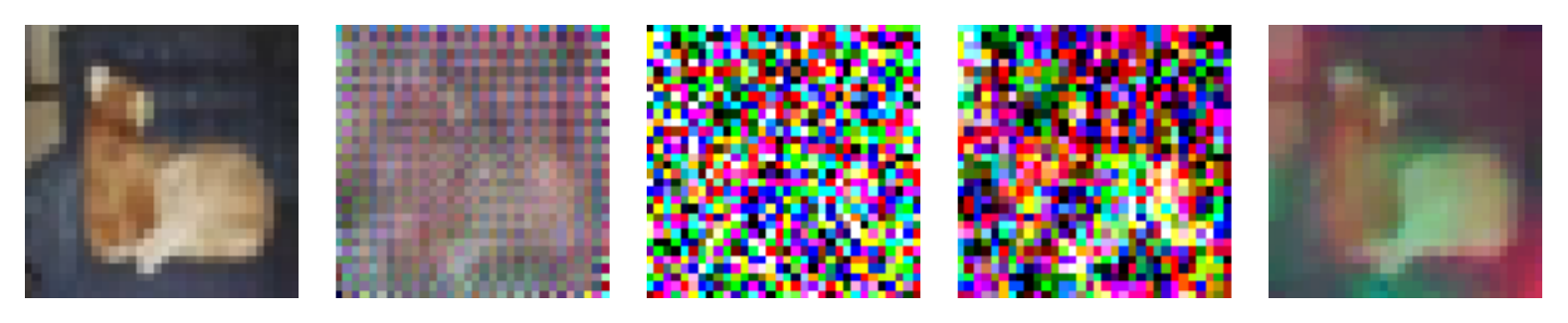}\\
			\includegraphics[width=\linewidth]{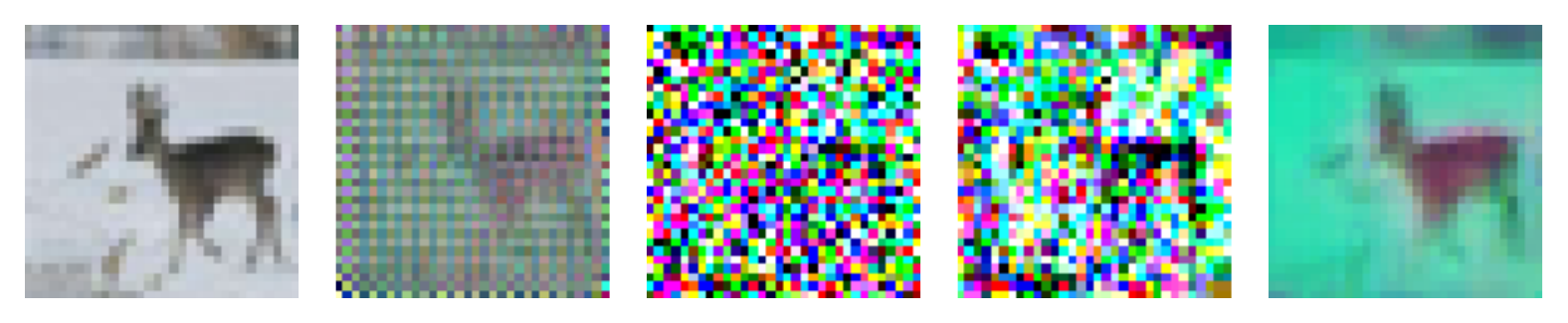}\\
			\includegraphics[width=\linewidth]{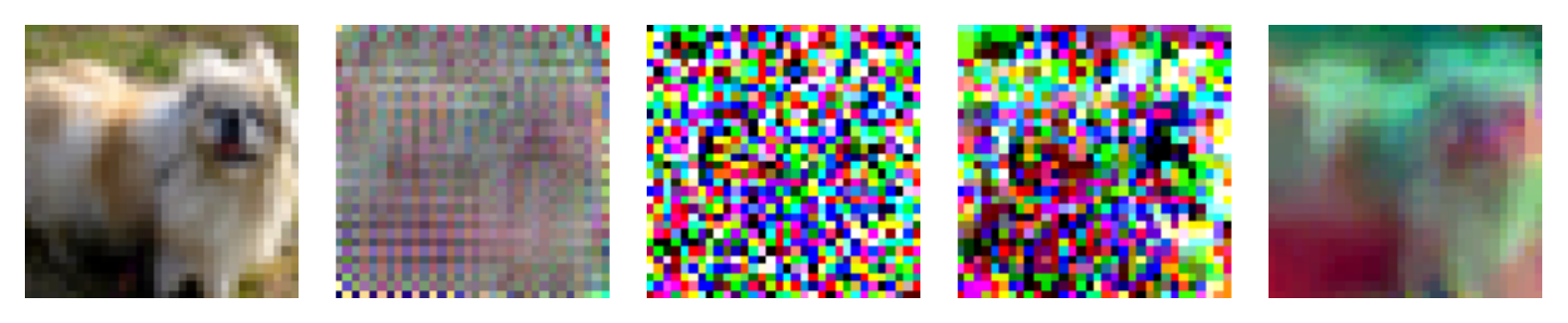}\\
			\includegraphics[width=\linewidth]{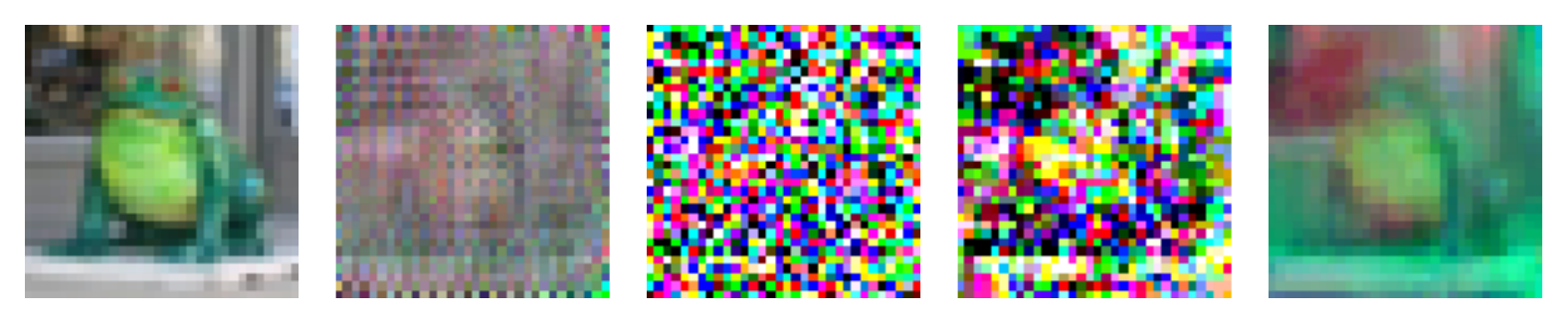}\\
			\includegraphics[width=\linewidth]{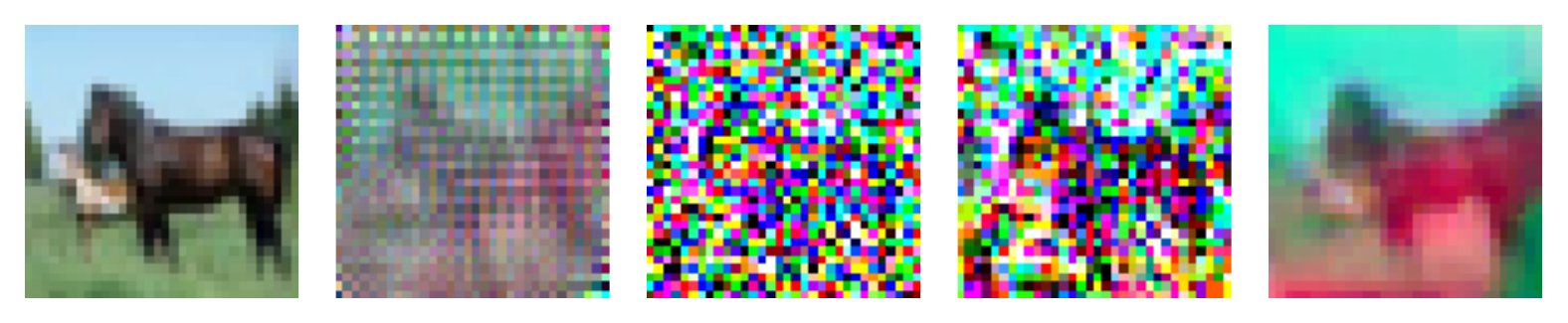}\\
			\includegraphics[width=\linewidth]{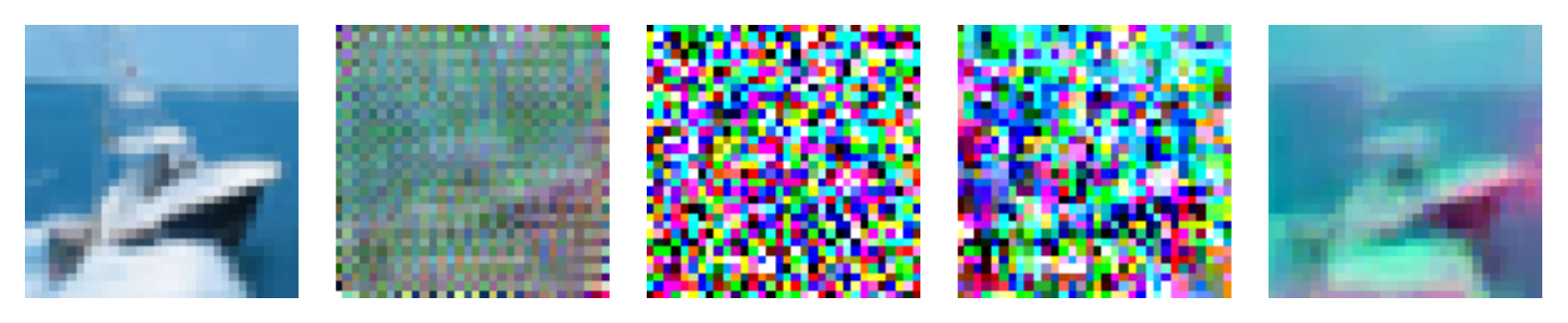}\\
			\includegraphics[width=\linewidth]{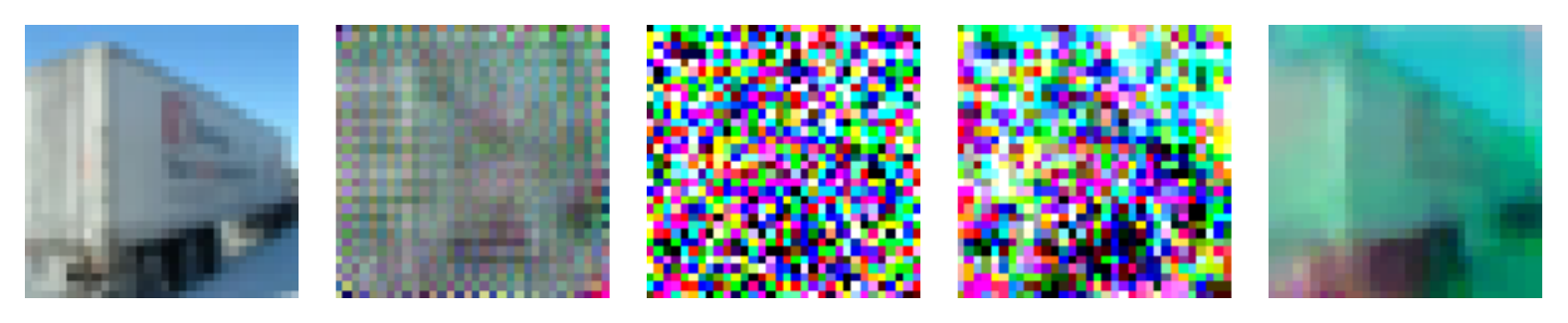}\\
			\caption{CNN3 Variant 1}
		\end{subfigure}\hspace{1cm}
		\begin{subfigure}[h]{0.45\textwidth}
			\includegraphics[width=\linewidth]{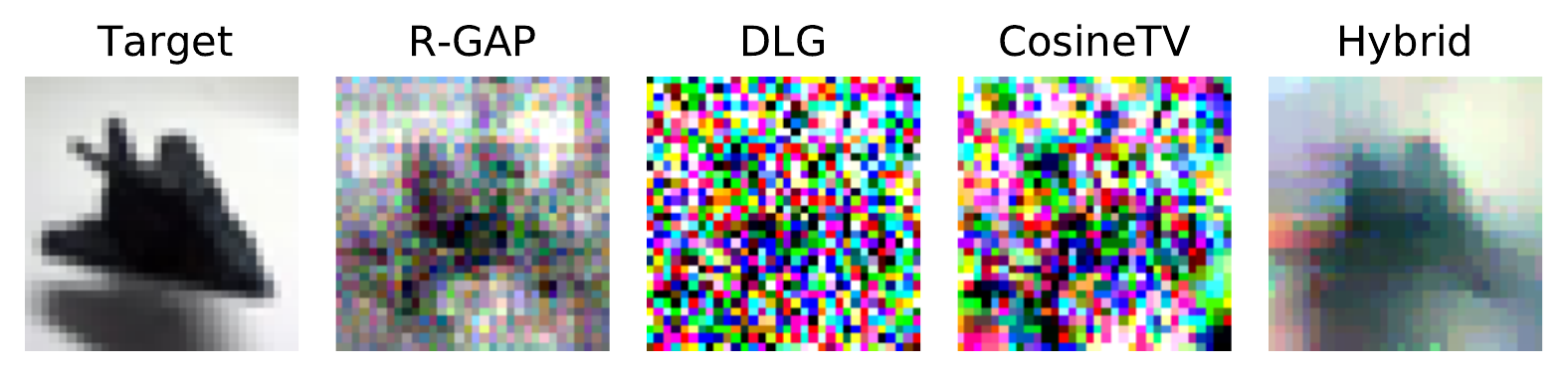}\\
			\includegraphics[width=\linewidth]{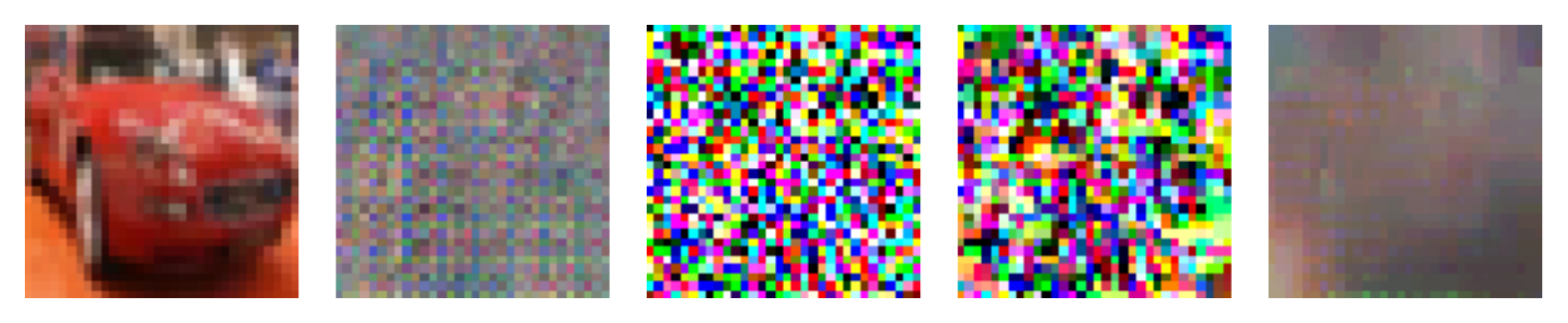}\\
			\includegraphics[width=\linewidth]{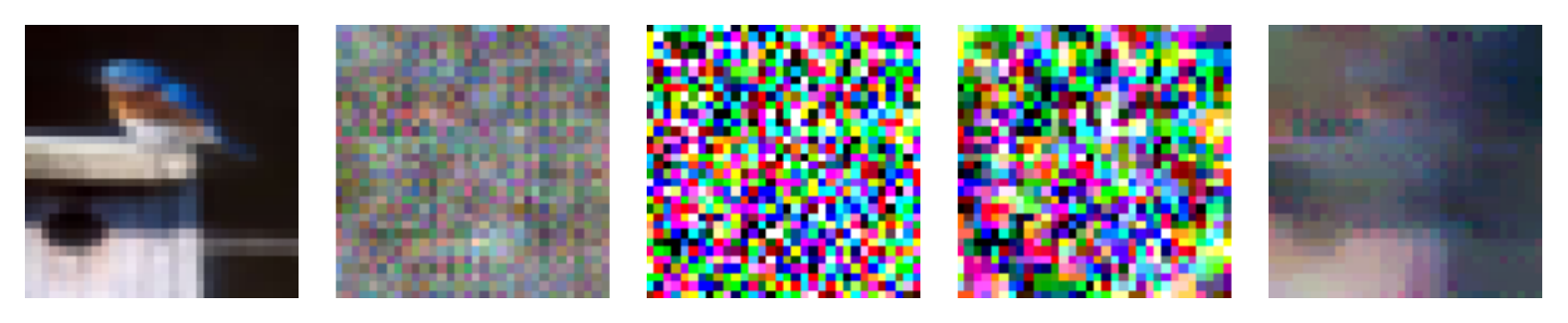}\\
			\includegraphics[width=\linewidth]{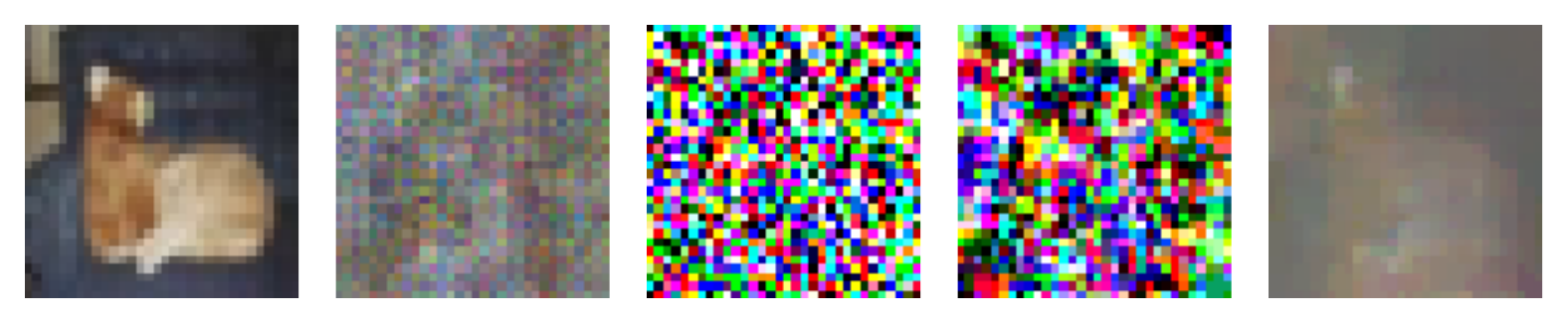}\\
			\includegraphics[width=\linewidth]{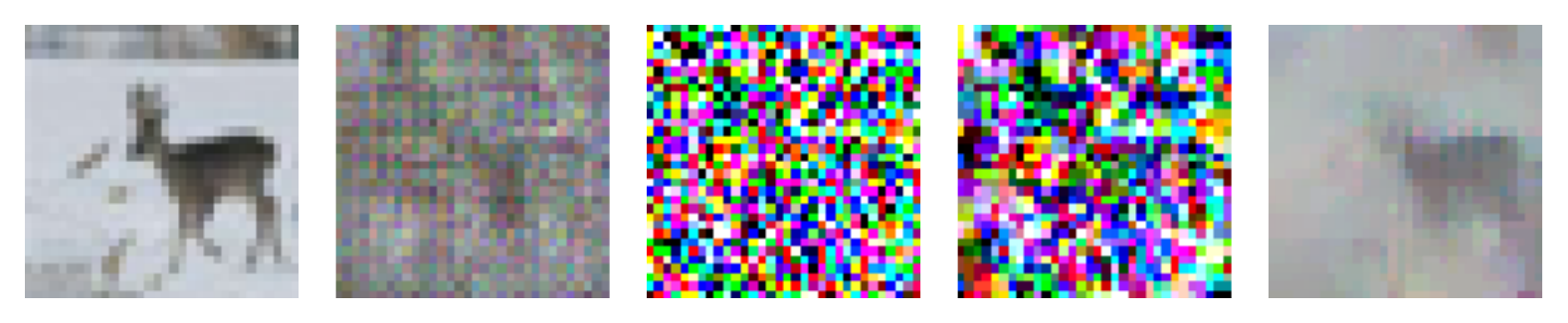}\\
			\includegraphics[width=\linewidth]{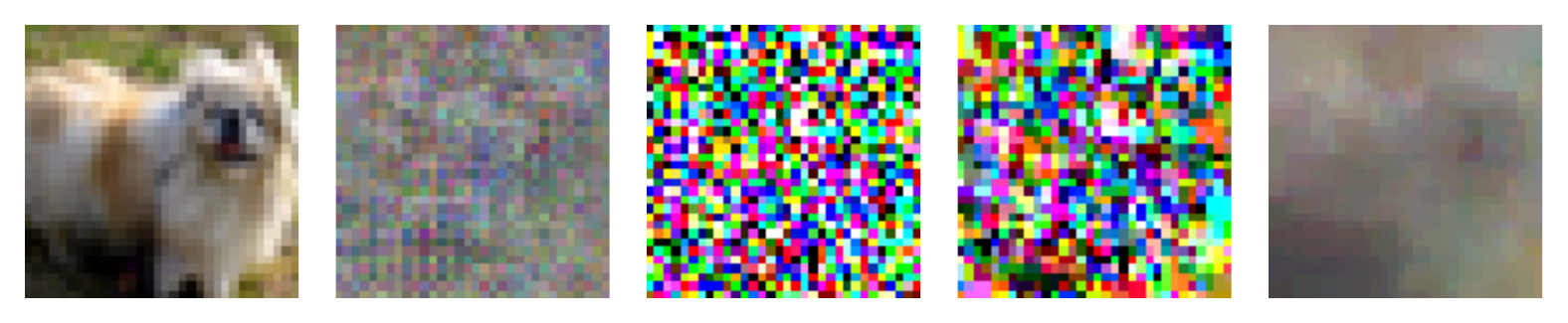}\\
			\includegraphics[width=\linewidth]{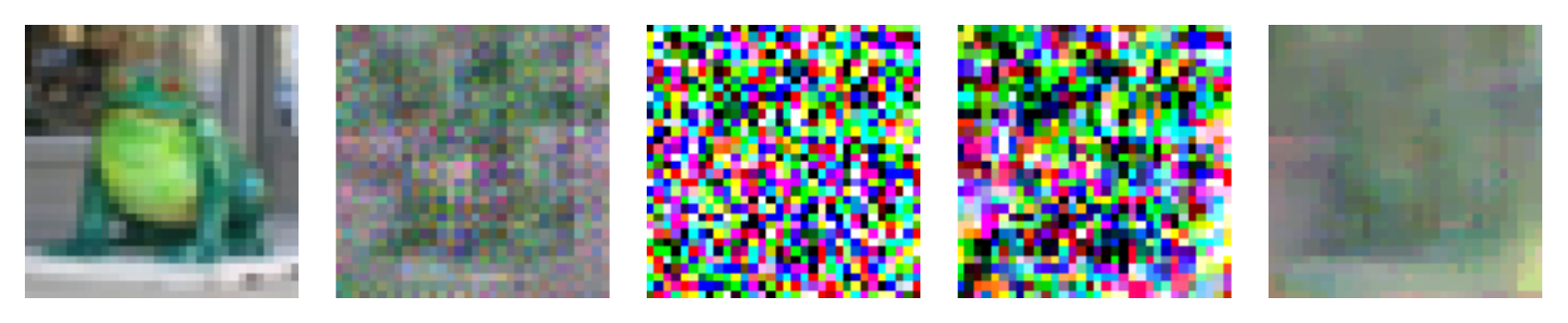}\\
			\includegraphics[width=\linewidth]{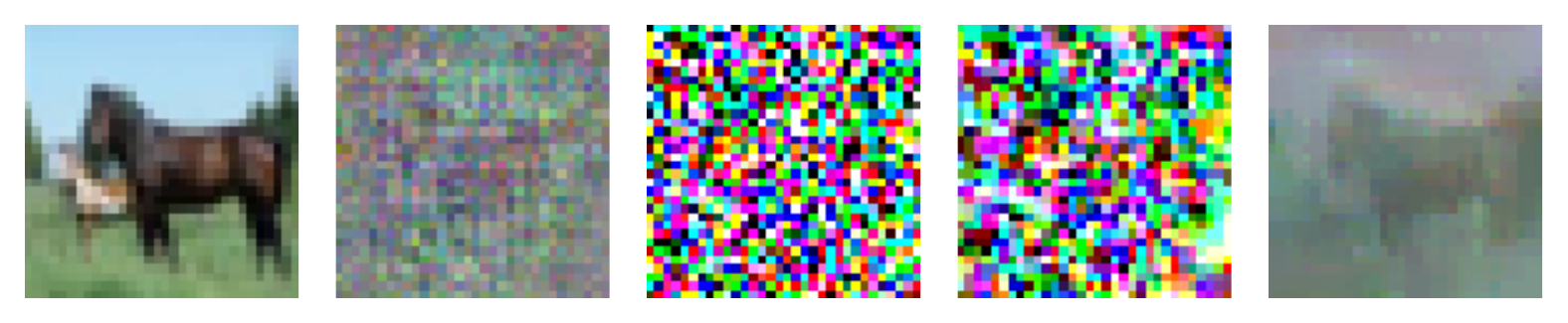}\\
			\includegraphics[width=\linewidth]{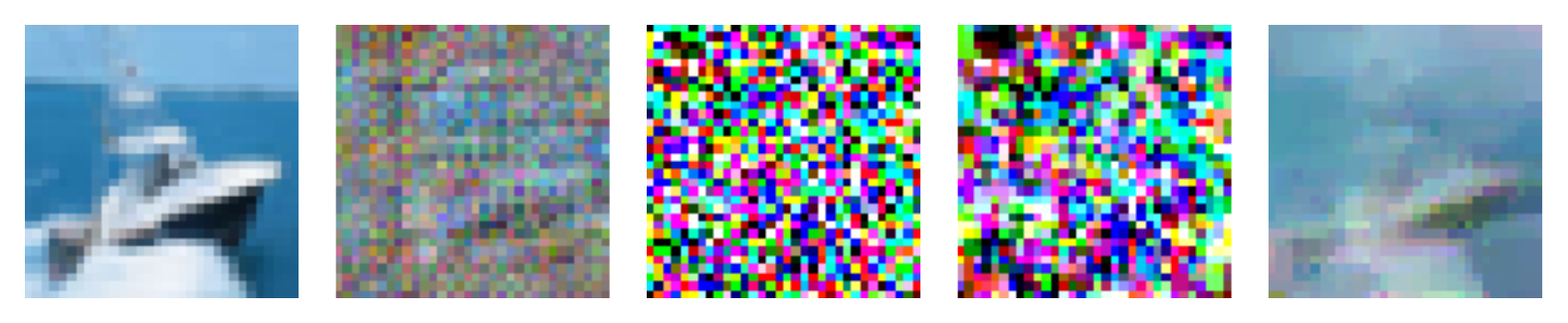}\\
			\includegraphics[width=\linewidth]{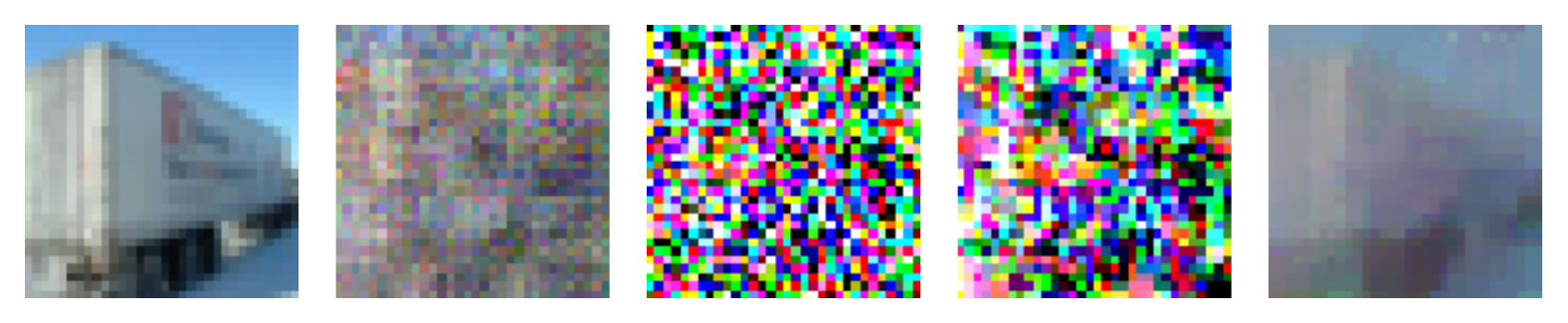}\\
			\caption{CNN3 Variant 2}
		\end{subfigure} \\
		\label{fig:multiimagecnn3v12}
\end{figure}
\begin{figure}[h]
		\centering
		%
		%
		\begin{subfigure}[h]{0.45\textwidth}
			\includegraphics[width=\linewidth]{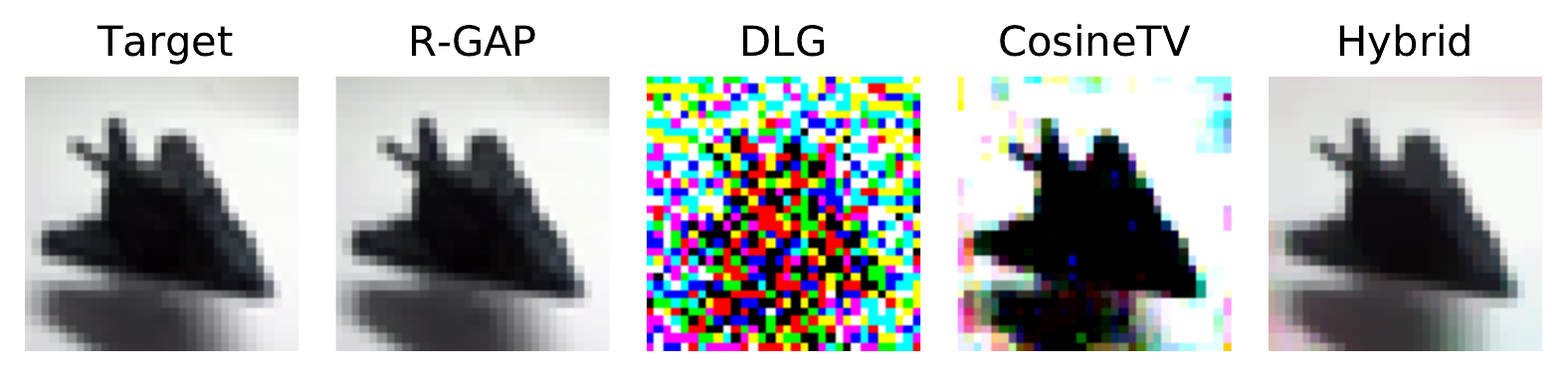}\\
			\includegraphics[width=\linewidth]{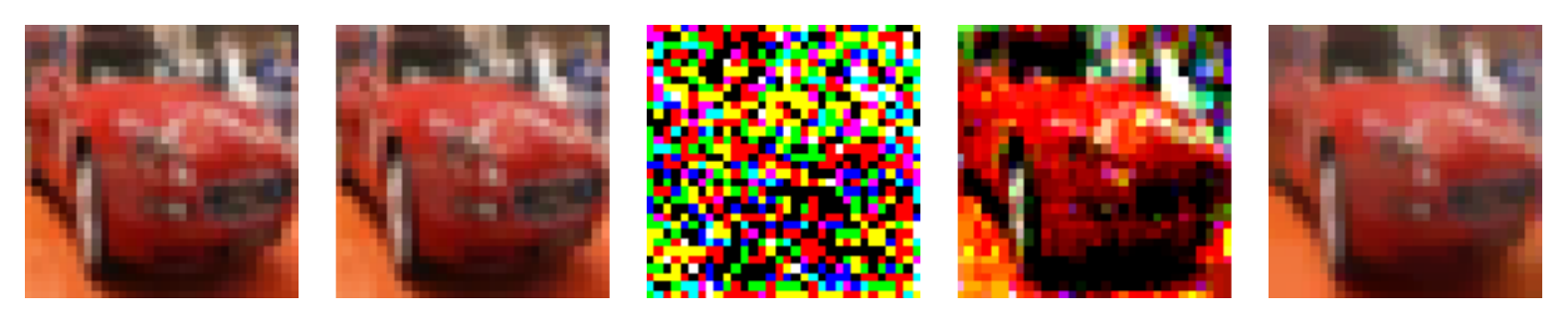}\\
			\includegraphics[width=\linewidth]{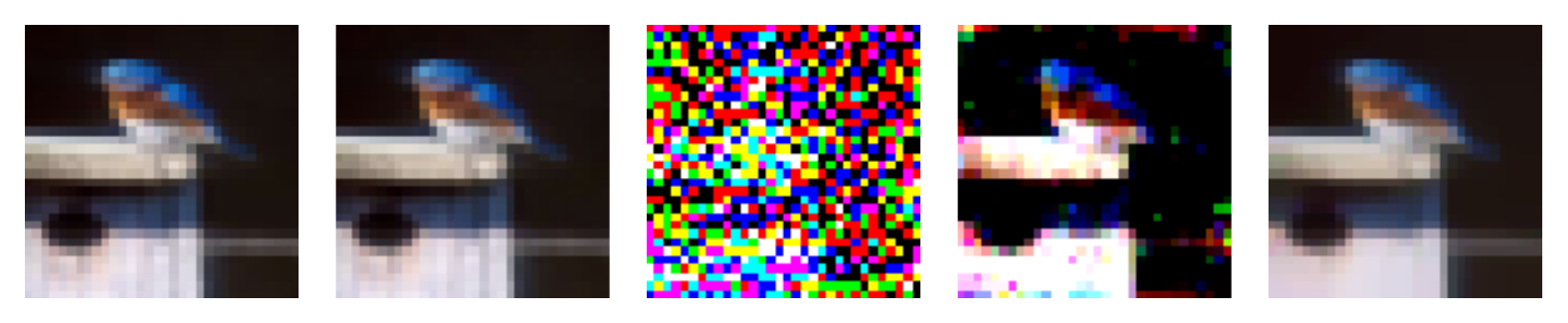}\\
			\includegraphics[width=\linewidth]{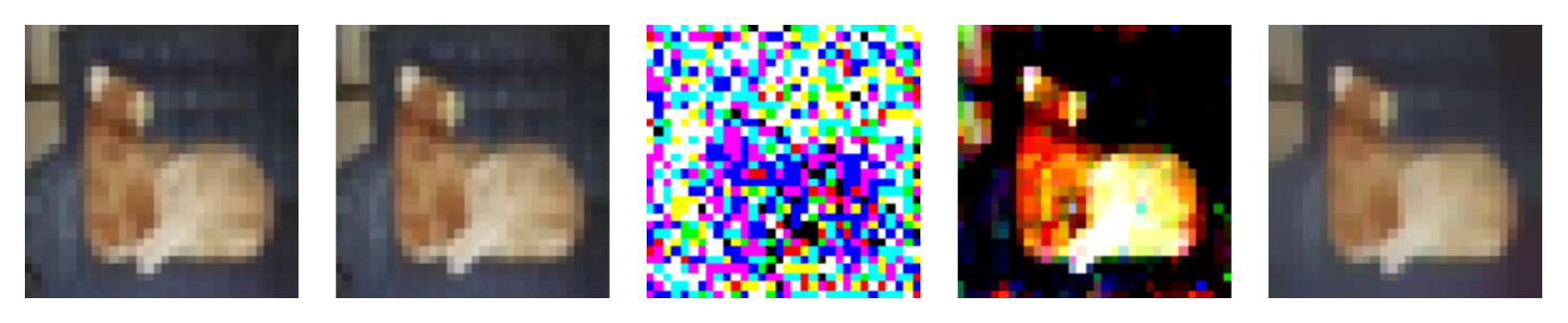}\\
			\includegraphics[width=\linewidth]{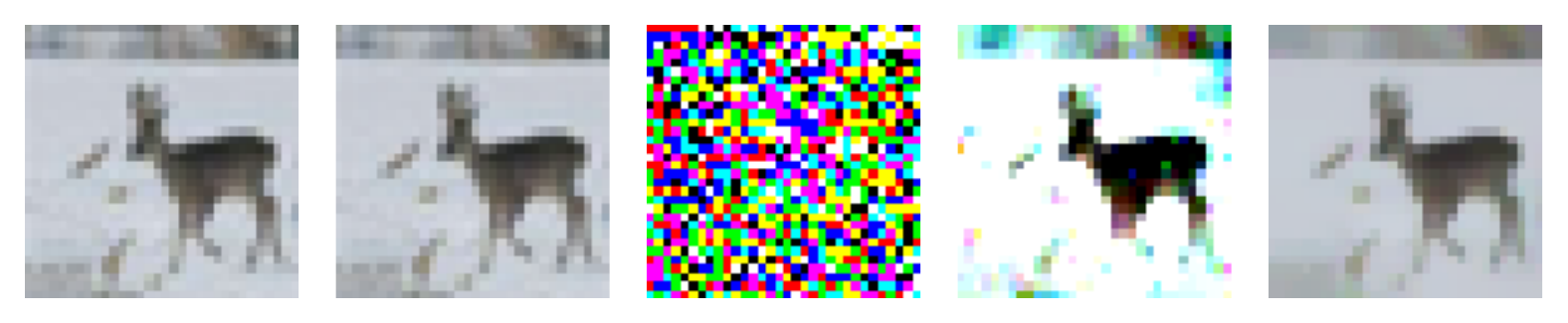}\\
			\includegraphics[width=\linewidth]{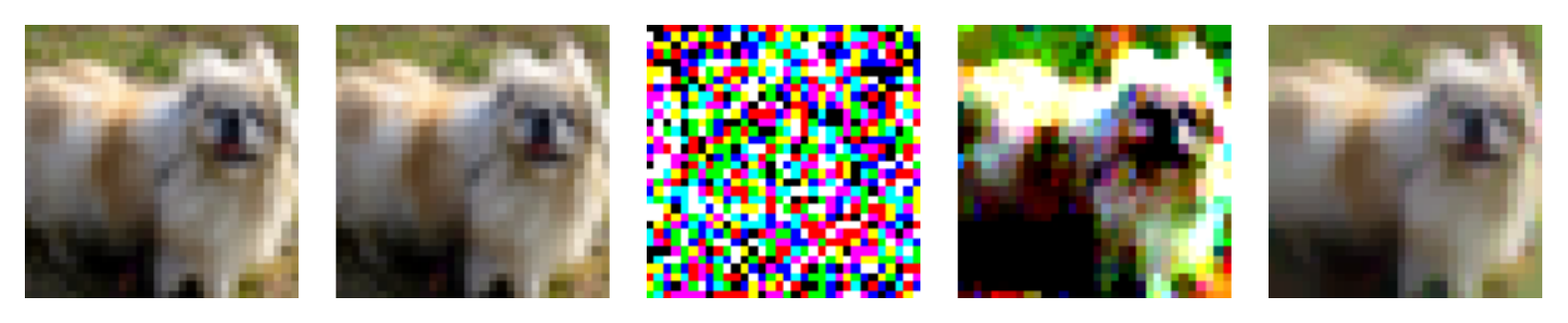}\\
			\includegraphics[width=\linewidth]{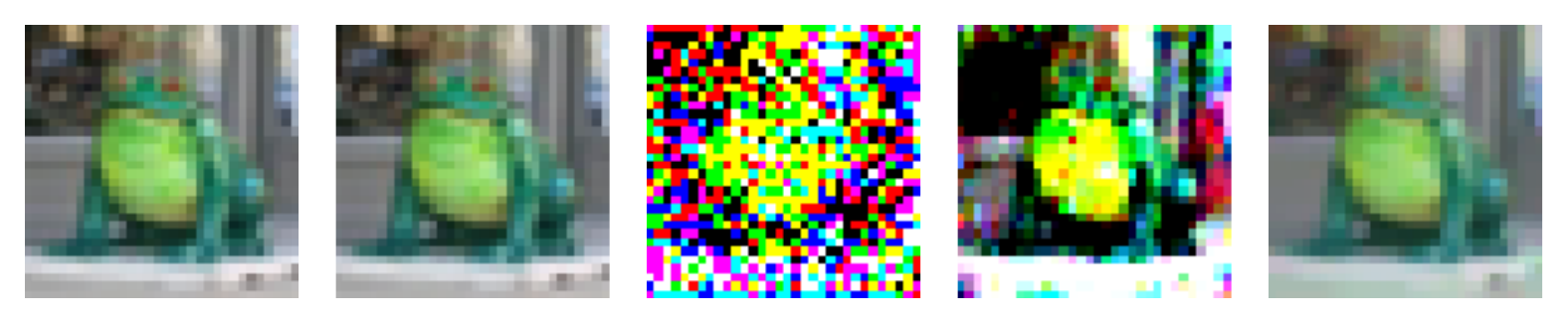}\\
			\includegraphics[width=\linewidth]{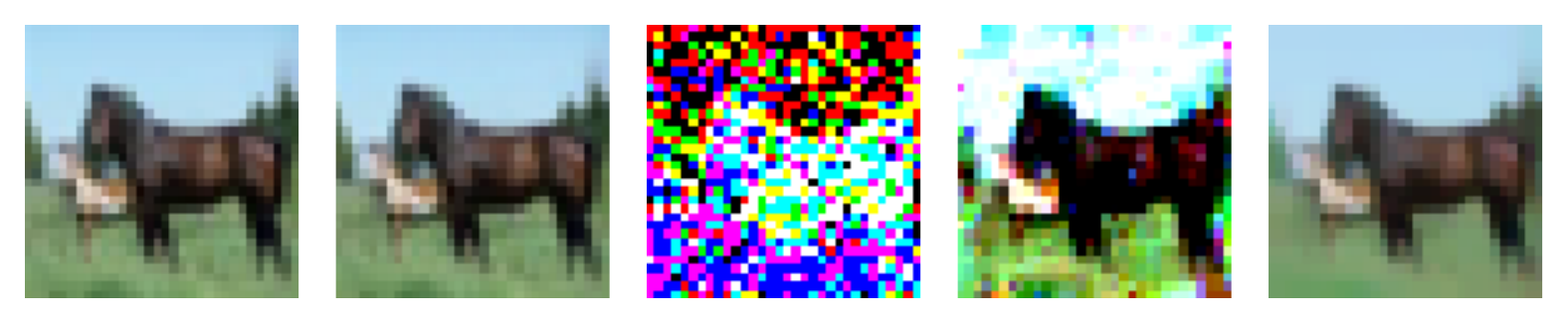}\\
			\includegraphics[width=\linewidth]{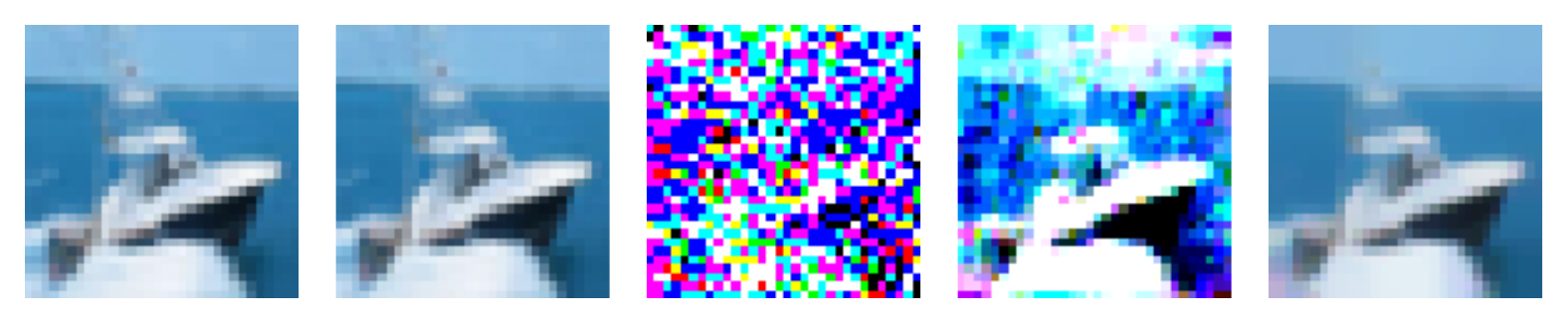}\\
			\includegraphics[width=\linewidth]{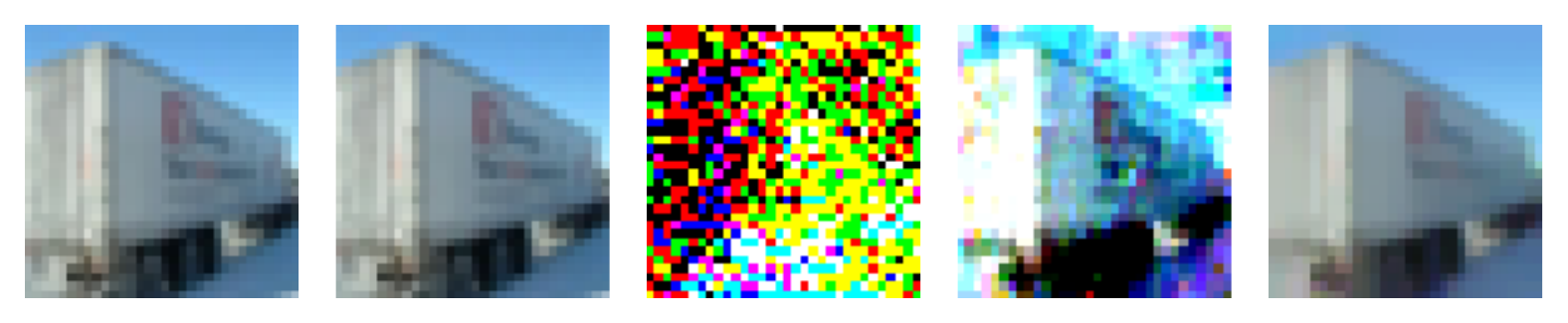}\\
			\caption{CNN3 Variant 3}
		\end{subfigure}\hspace{1cm}
		\begin{subfigure}[h]{0.45\textwidth}
			\includegraphics[width=\linewidth]{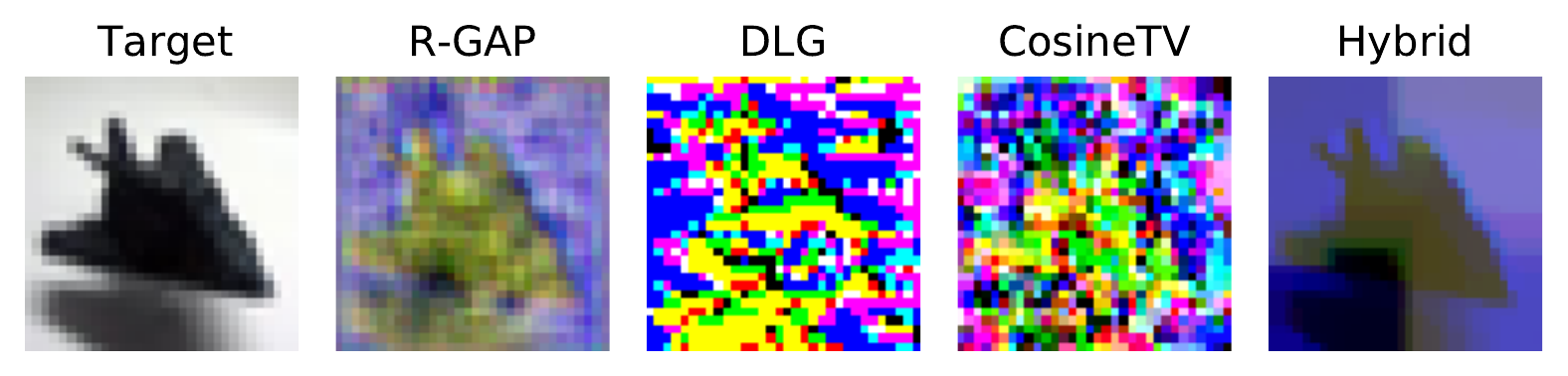}\\
			\includegraphics[width=\linewidth]{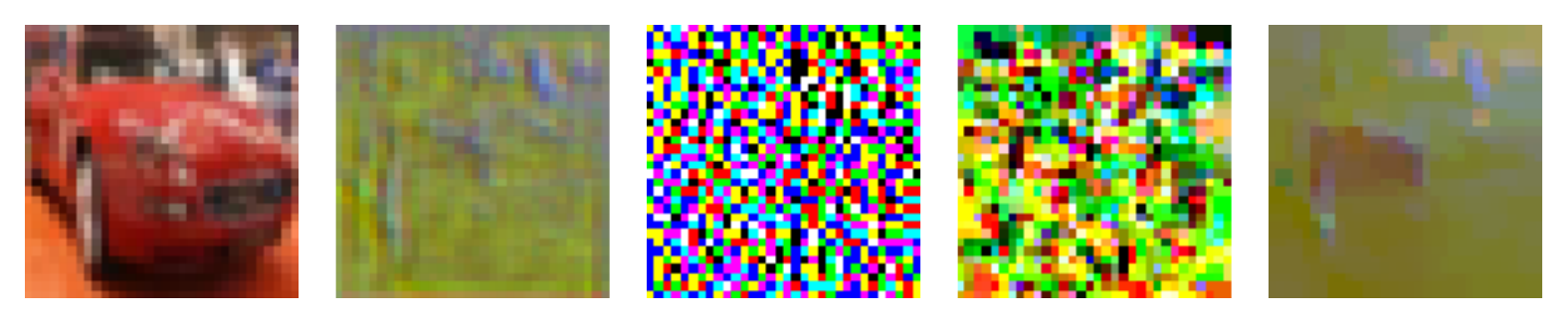}\\
			\includegraphics[width=\linewidth]{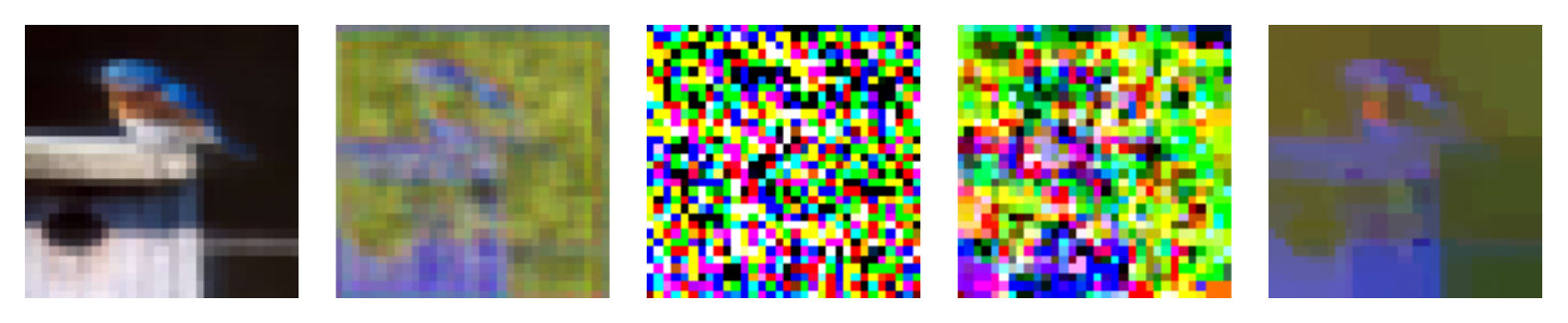}\\
			\includegraphics[width=\linewidth]{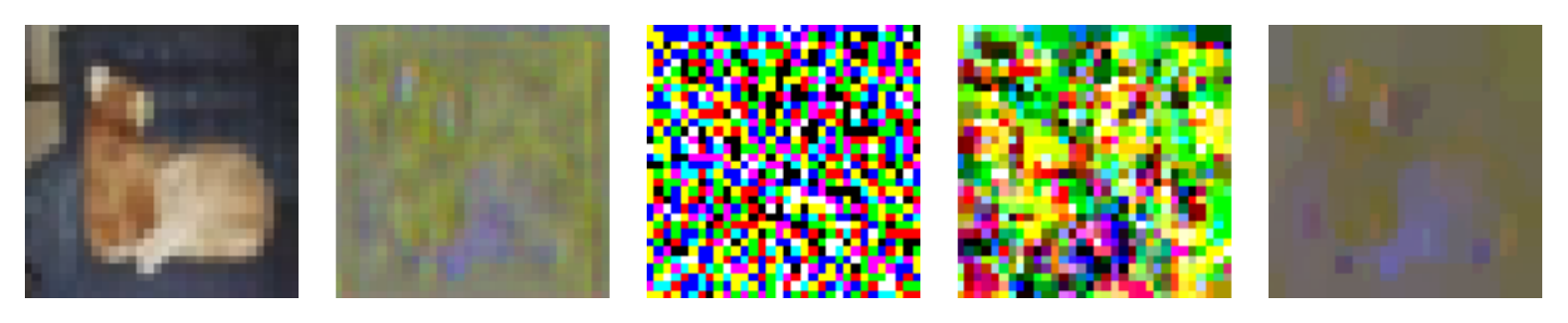}\\
			\includegraphics[width=\linewidth]{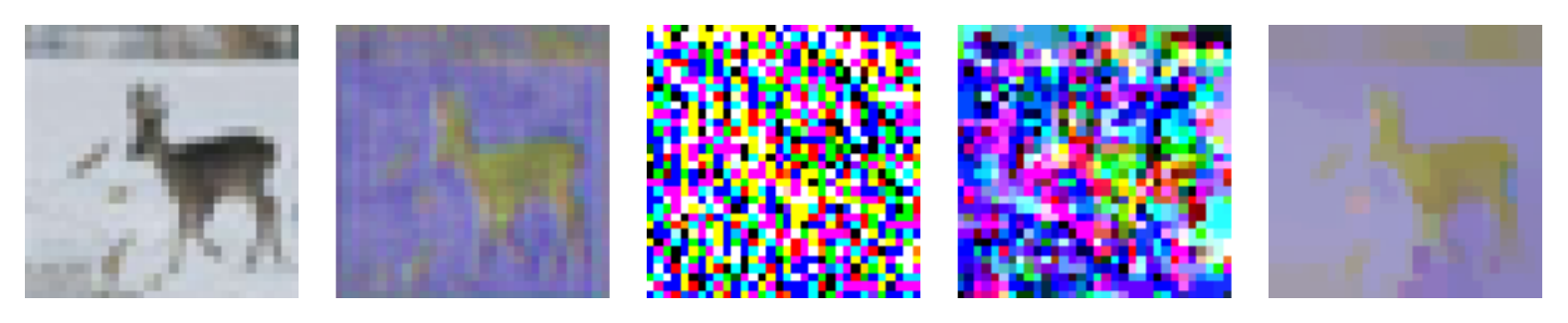}\\
			\includegraphics[width=\linewidth]{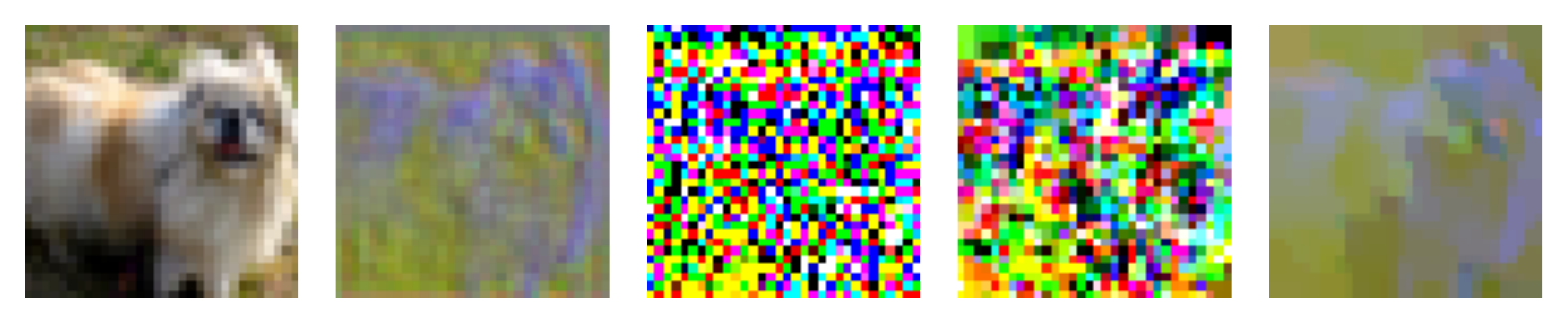}\\
			\includegraphics[width=\linewidth]{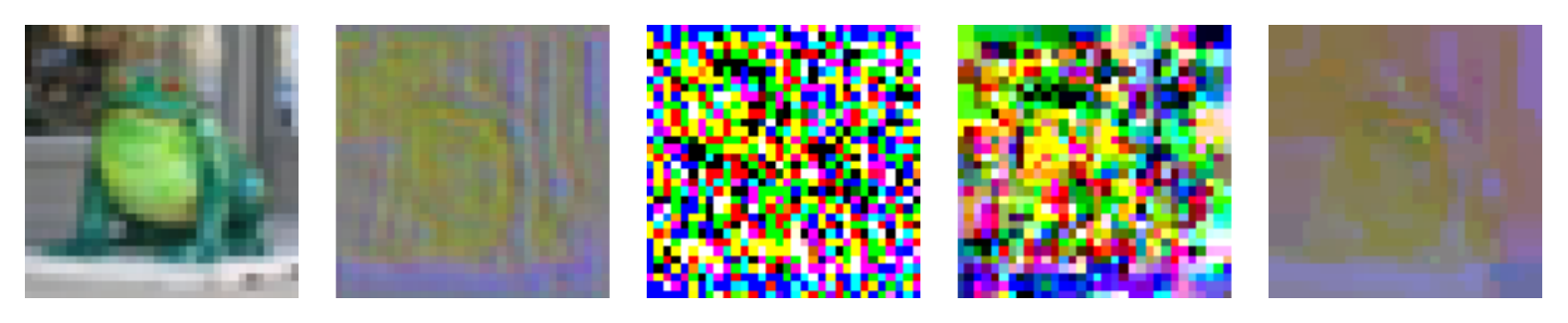}\\
			\includegraphics[width=\linewidth]{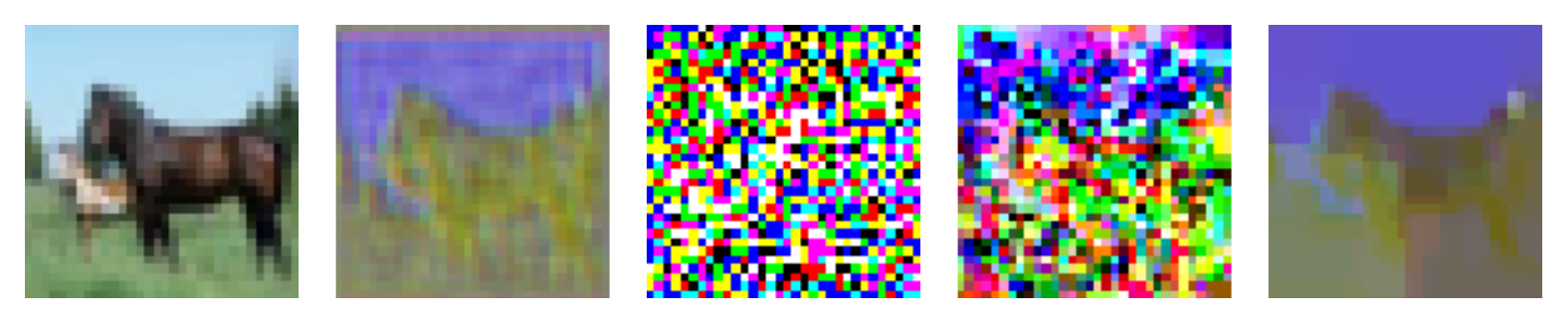}\\
			\includegraphics[width=\linewidth]{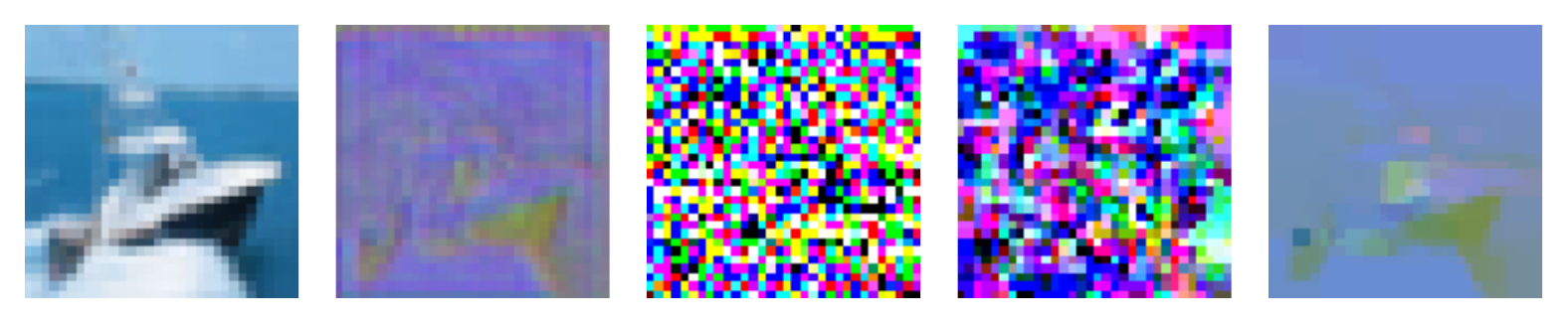}\\
			\includegraphics[width=\linewidth]{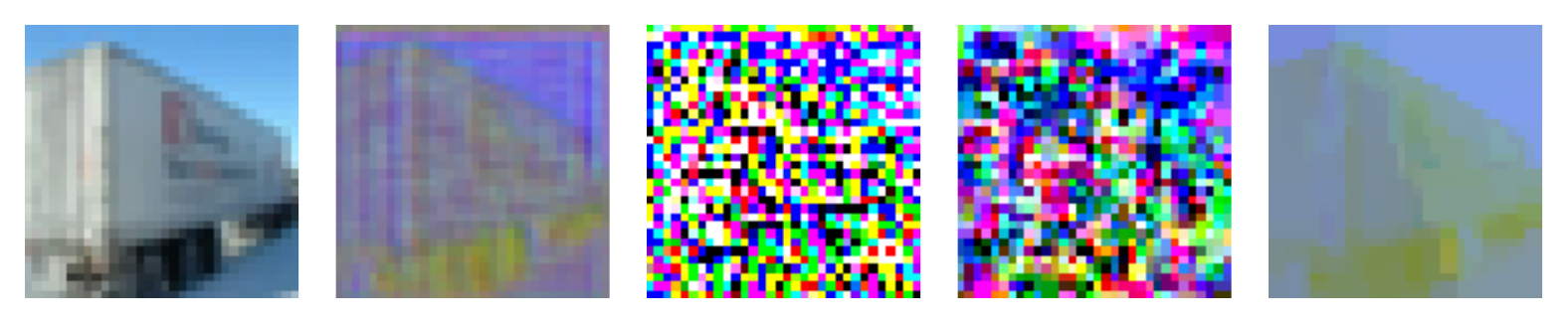}\\
			\caption{CNN3 Variant 4}
		\end{subfigure} \\
		\label{fig:multiimagecnn3v34}
\end{figure}
\begin{figure}[h]
		\centering
		%
		%
		\begin{subfigure}[h]{0.45\textwidth}
			\includegraphics[width=\linewidth]{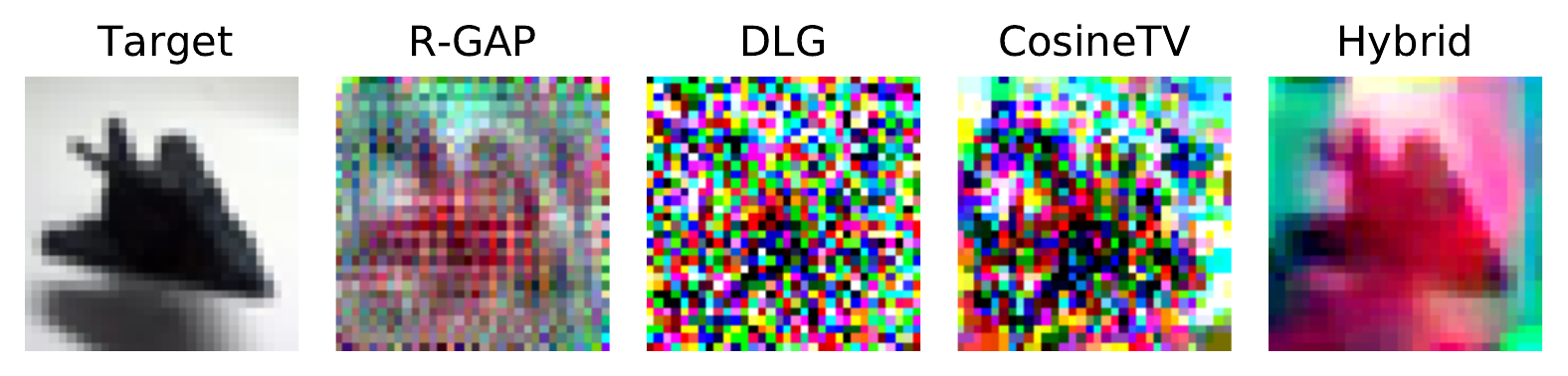}\\
			\includegraphics[width=\linewidth]{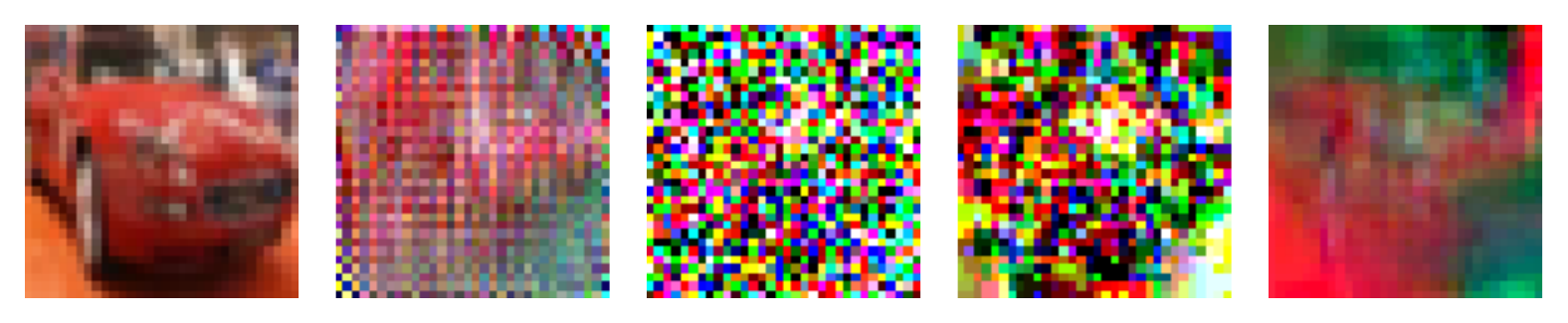}\\
			\includegraphics[width=\linewidth]{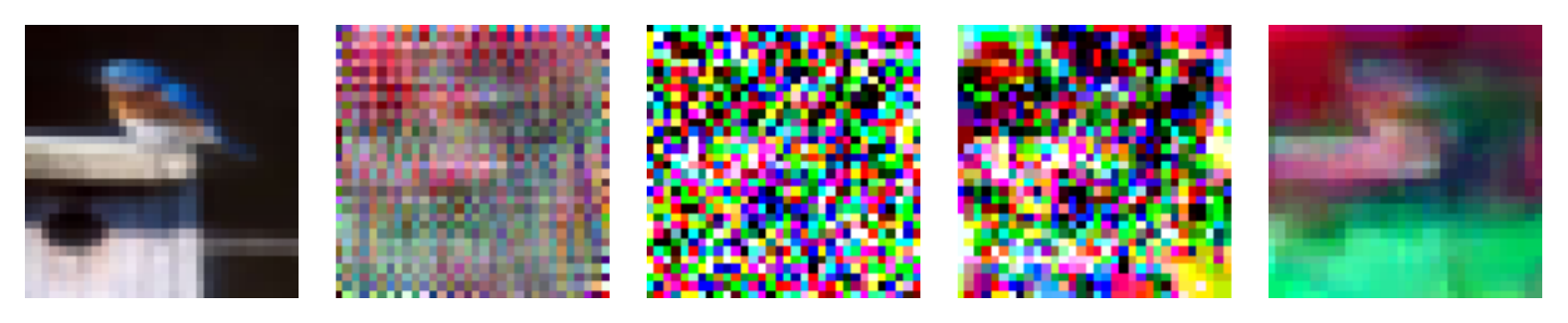}\\
			\includegraphics[width=\linewidth]{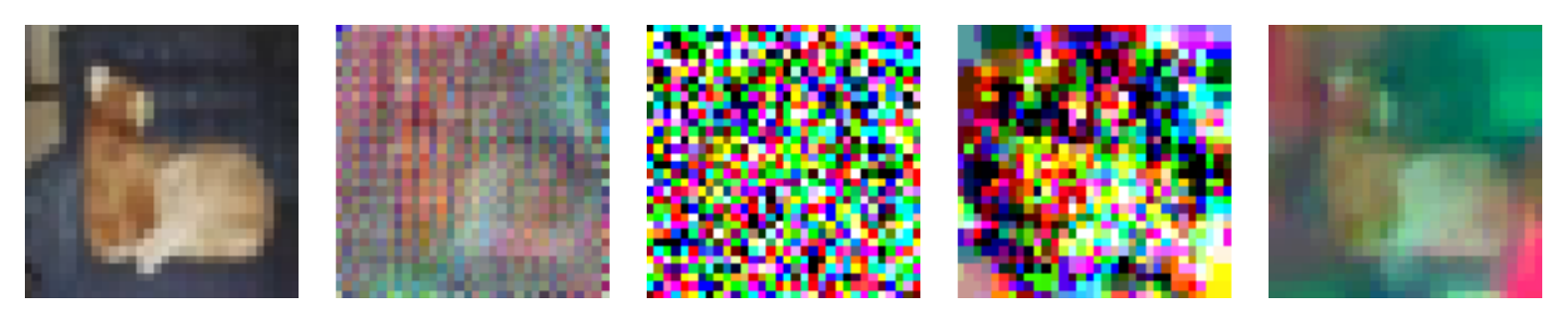}\\
			\includegraphics[width=\linewidth]{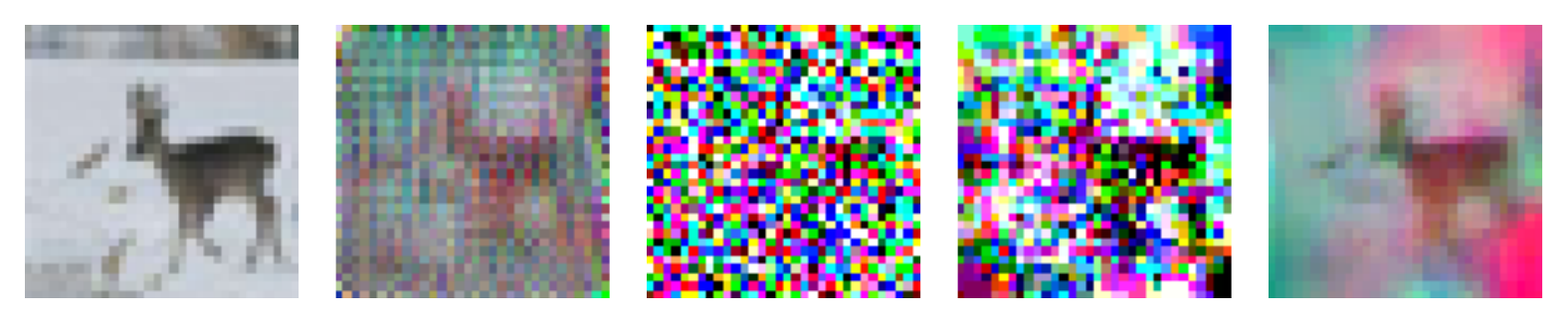}\\
			\includegraphics[width=\linewidth]{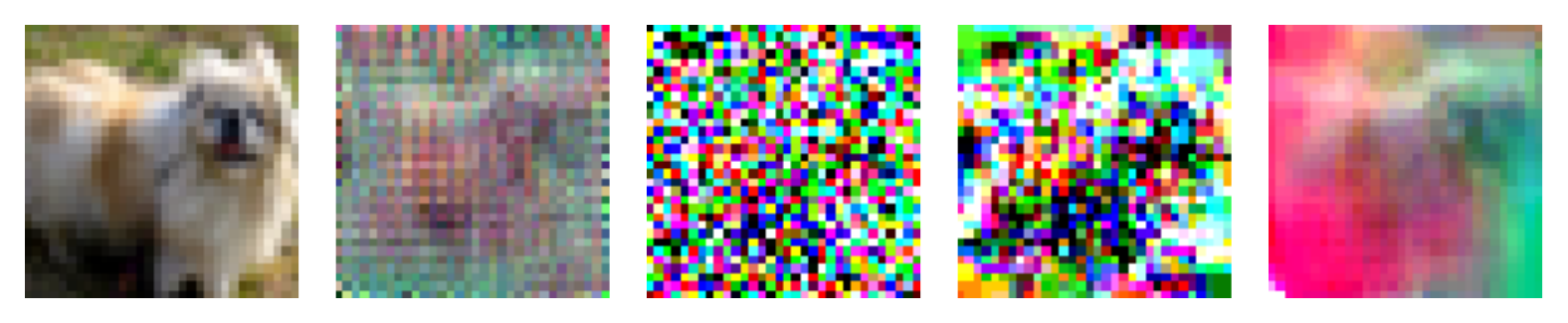}\\
			\includegraphics[width=\linewidth]{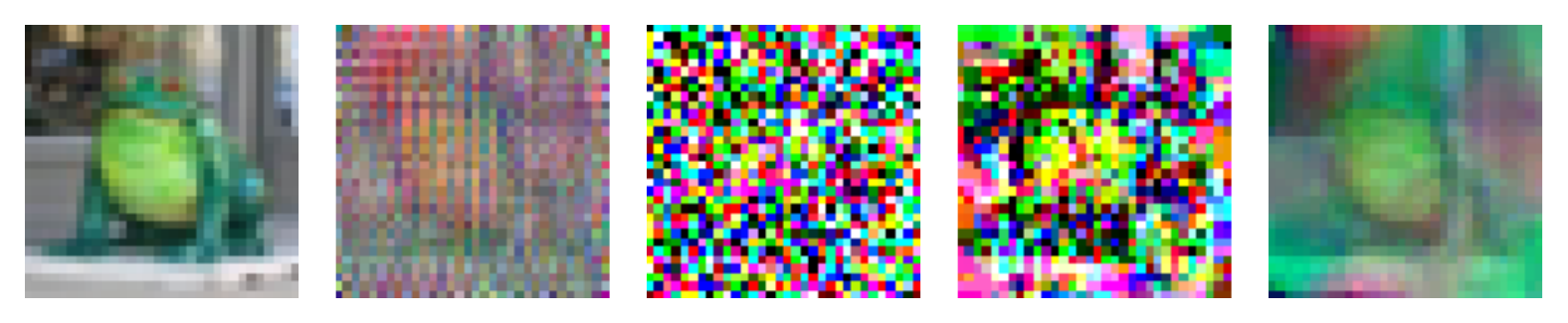}\\
			\includegraphics[width=\linewidth]{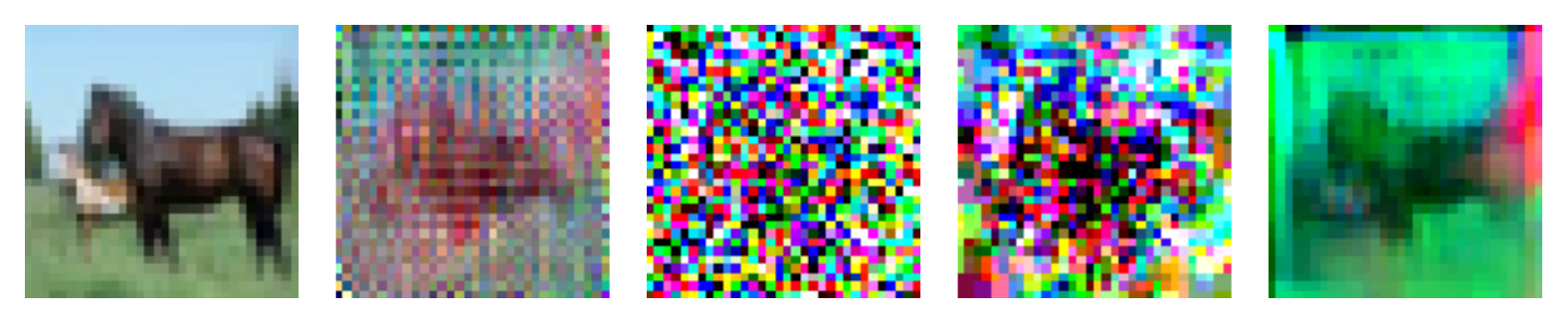}\\
			\includegraphics[width=\linewidth]{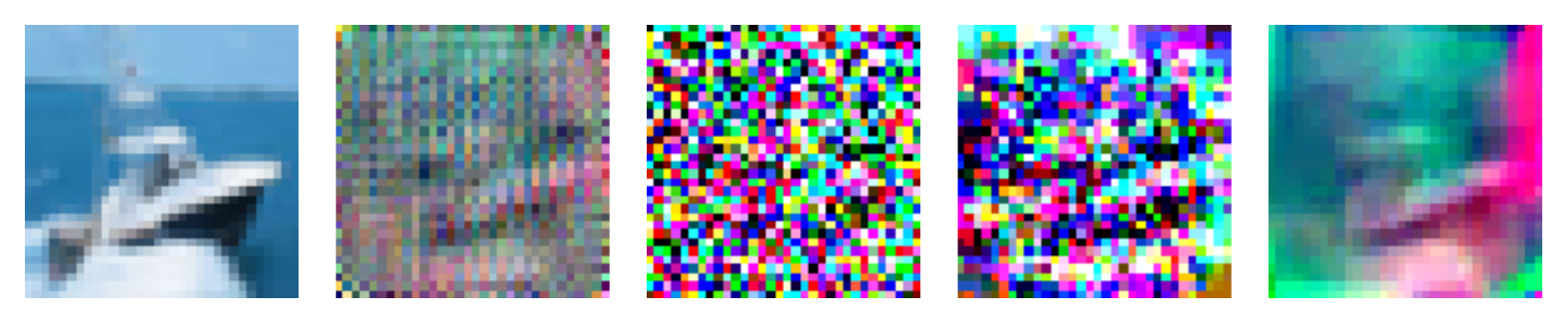}\\
			\includegraphics[width=\linewidth]{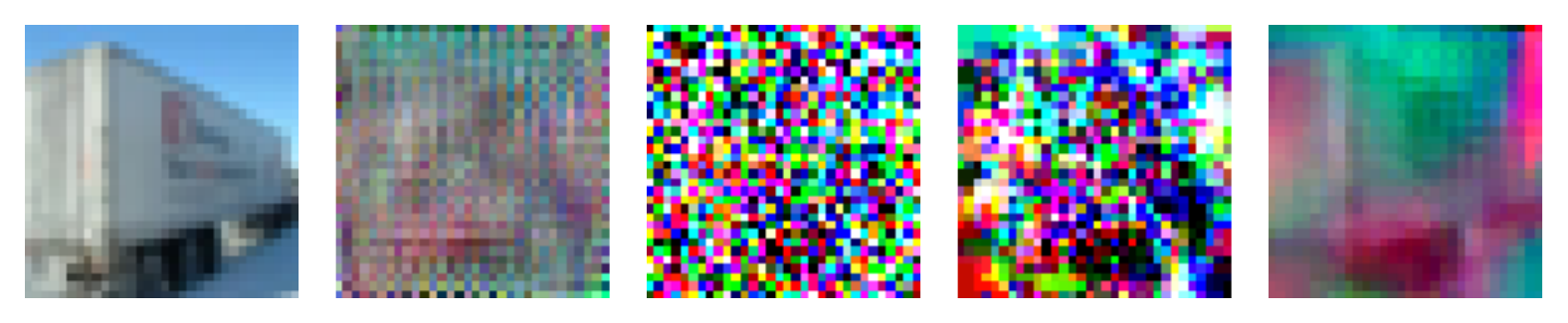}\\
			\caption{CNN4 Variant 1}
		\end{subfigure}\hspace{1cm}
		\begin{subfigure}[h]{0.45\textwidth}
			\includegraphics[width=\linewidth]{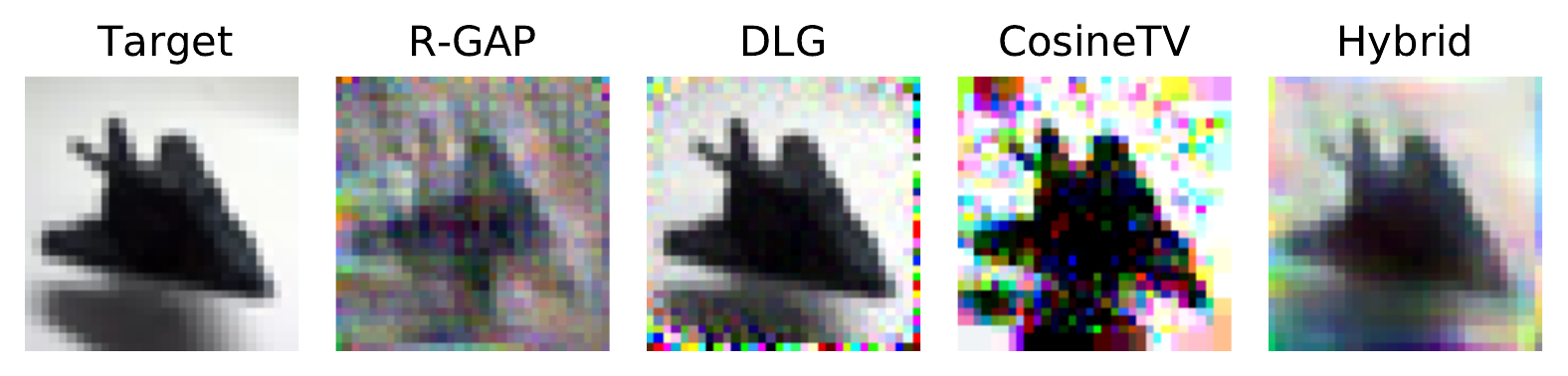}\\
			\includegraphics[width=\linewidth]{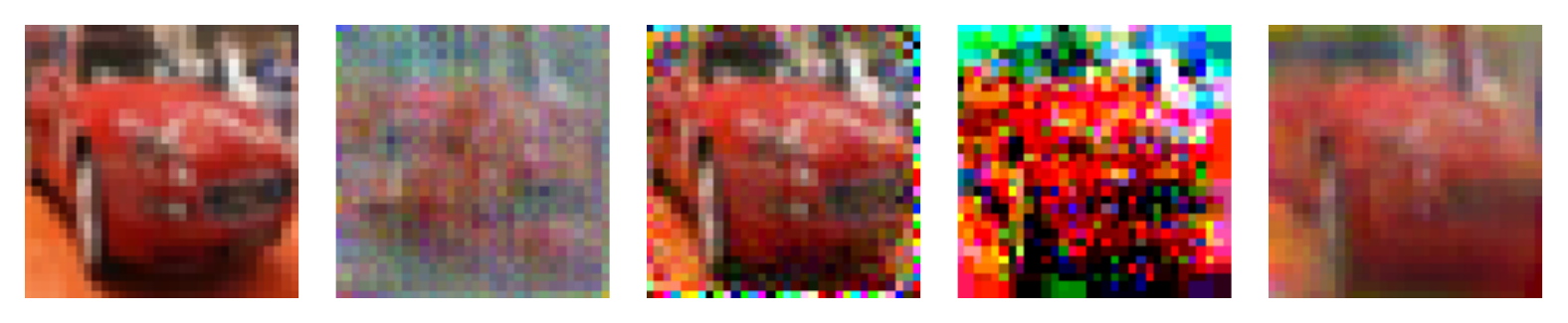}\\
			\includegraphics[width=\linewidth]{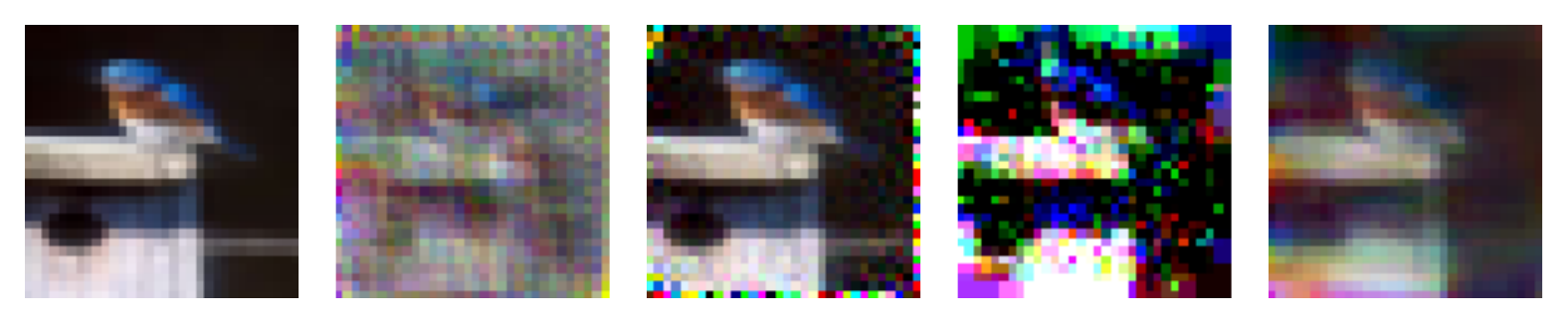}\\
			\includegraphics[width=\linewidth]{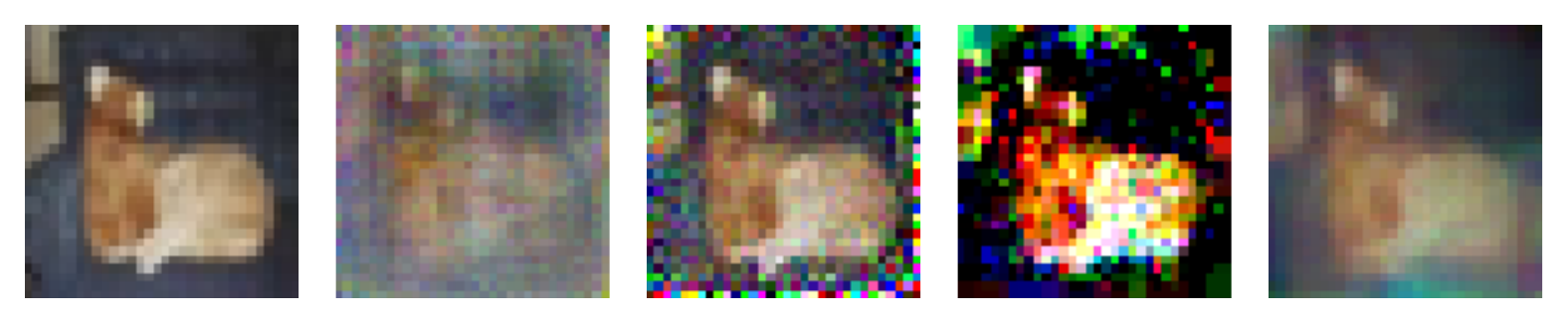}\\
			\includegraphics[width=\linewidth]{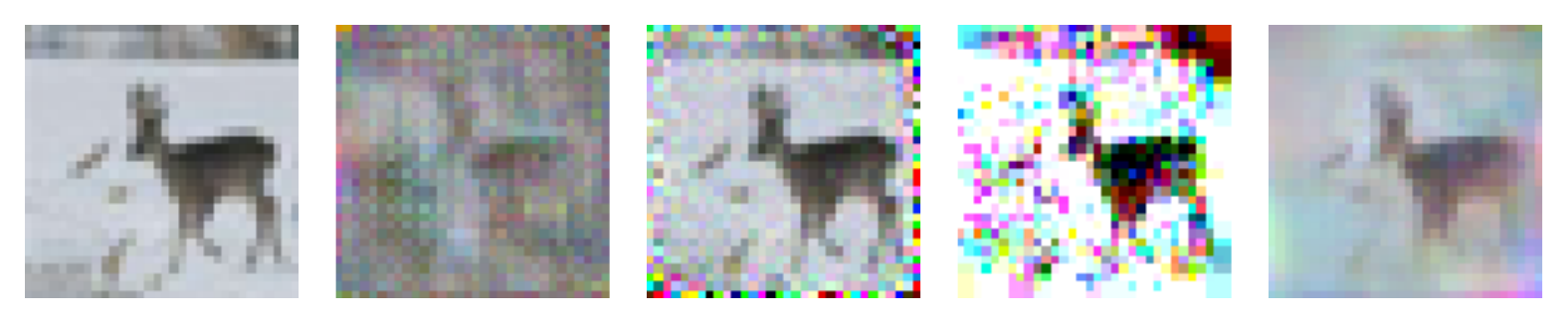}\\
			\includegraphics[width=\linewidth]{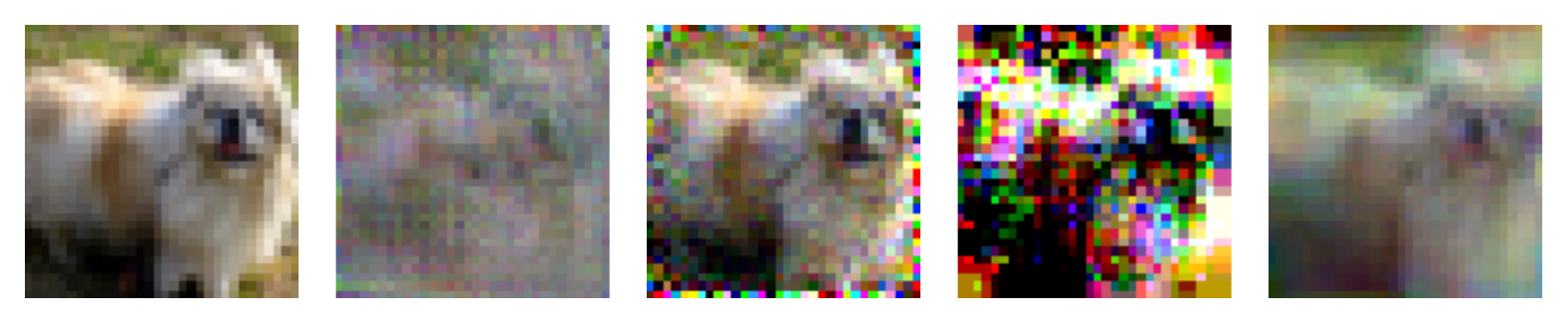}\\
			\includegraphics[width=\linewidth]{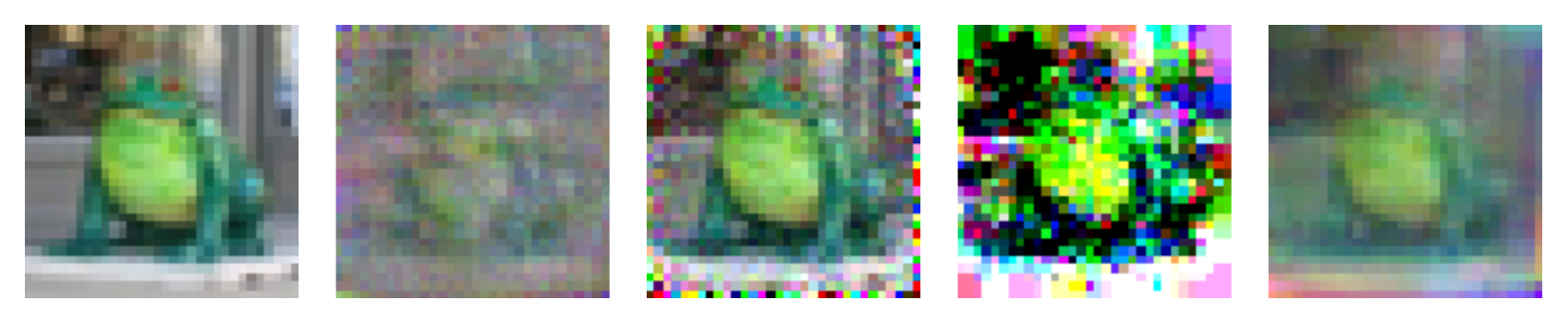}\\
			\includegraphics[width=\linewidth]{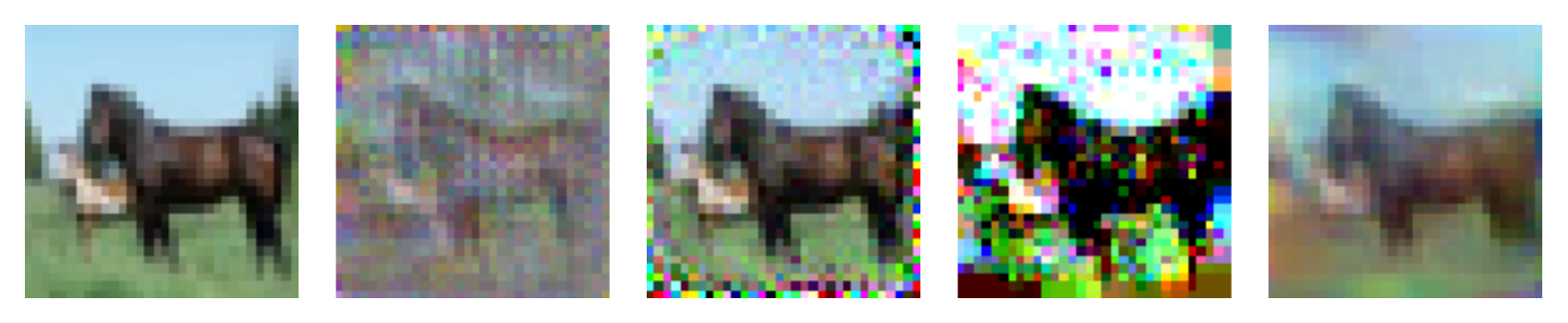}\\
			\includegraphics[width=\linewidth]{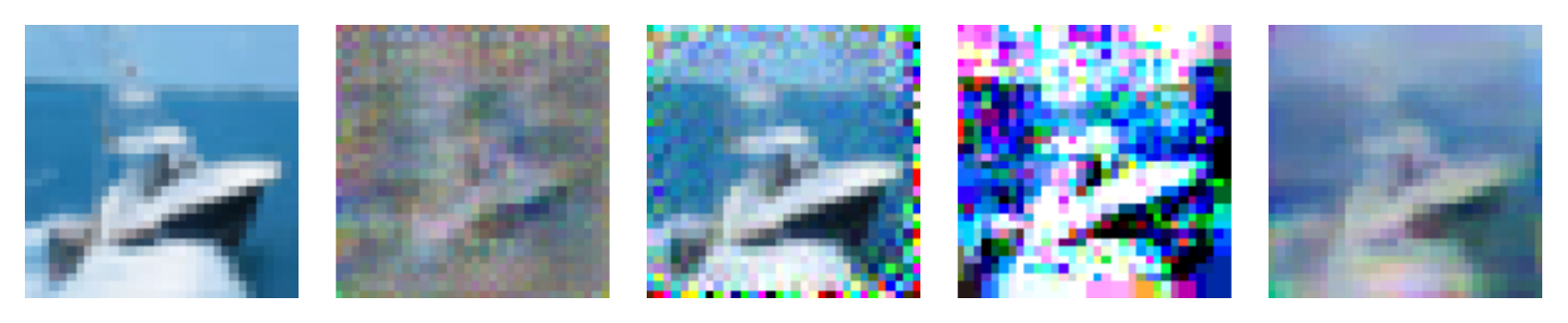}\\
			\includegraphics[width=\linewidth]{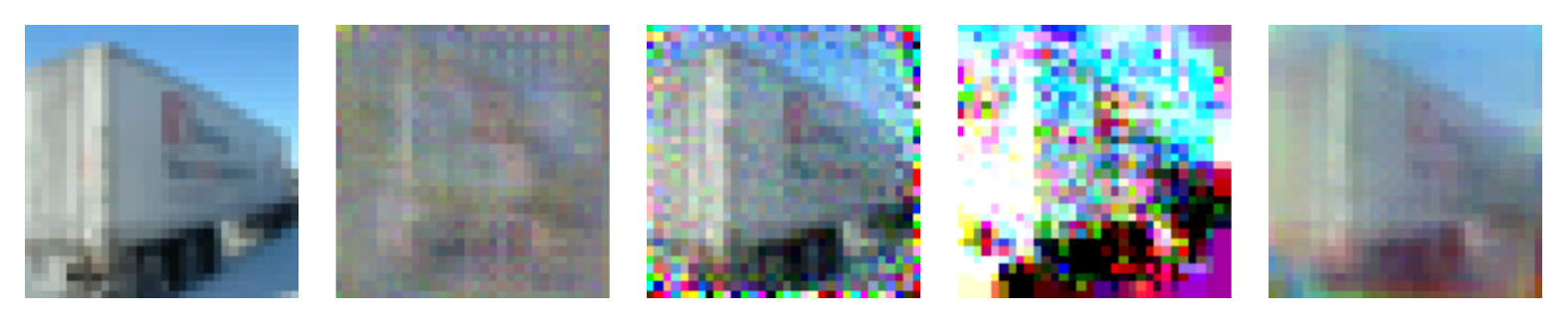}\\
			\caption{CNN4 Variant 2}
		\end{subfigure} \\
		\label{fig:multiimagecnn4v12}
\end{figure}

\end{document}